\documentclass[11pt]{article}

\usepackage{fullpage}
\usepackage{lscape}
\usepackage{setspace}
\usepackage{breakcites}
\usepackage[dvipsnames]{xcolor}
\usepackage[hidelinks]{hyperref}
\hypersetup{
	colorlinks=true,
	linkcolor=blue!70!black,
	citecolor=blue!70!black,
	urlcolor=blue!70!black}

\usepackage{microtype,bm}
\usepackage{xspace}
\usepackage{algorithm}
\usepackage{verbatim}
\usepackage[noend]{algpseudocode}
\usepackage{mathtools}
\usepackage{amsmath}
\usepackage{bbm}
\usepackage{amsfonts}
\usepackage{amssymb}
\usepackage{xpatch}
\usepackage{enumitem}
\usepackage{amsthm}

\usepackage[nointegrals]{wasysym}

\theoremstyle{definition}  %Sets style of 
\newtheorem{fact}{Fact}

\theoremstyle{plain}

\newtheorem{theorem}{Theorem}

\newtheorem{lemma}{Lemma}

\xpatchcmd{\proof}{\itshape}{\normalfont\proofnameformat}{}{}
\newcommand{\proofnameformat}{\bfseries}

\usepackage{prettyref}
\newcommand{\pref}[1]{\prettyref{#1}}

\newcommand{\savehyperref}[2]{\texorpdfstring{\hyperref[#1]{#2}}{#2}}
\newrefformat{eq}{\savehyperref{#1}{\textup{(\ref*{#1})}}}
\newrefformat{eqn}{\savehyperref{#1}{Equation~\ref*{#1}}}
\newrefformat{lem}{\savehyperref{#1}{Lemma~\ref*{#1}}}
\newrefformat{obs}{\savehyperref{#1}{Observation~\ref*{#1}}}
\newrefformat{def}{\savehyperref{#1}{Definition~\ref*{#1}}}
\newrefformat{line}{\savehyperref{#1}{line~\ref*{#1}}}
\newrefformat{thm}{\savehyperref{#1}{Theorem~\ref*{#1}}}
\newrefformat{corr}{\savehyperref{#1}{Corollary~\ref*{#1}}}
\newrefformat{sec}{\savehyperref{#1}{Section~\ref*{#1}}}
\newrefformat{exam}{\savehyperref{#1}{Example~\ref*{#1}}}
\newrefformat{prop}{\savehyperref{#1}{Proposition~\ref*{#1}}}
\newrefformat{table}{\savehyperref{#1}{Table~\ref*{#1}}}
\newrefformat{app}{\savehyperref{#1}{Appendix~\ref*{#1}}}
\newrefformat{ass}{\savehyperref{#1}{Assumption~\ref*{#1}}}
\newrefformat{tbl}{\savehyperref{#1}{Table~\ref*{#1}}}
\newrefformat{ex}{\savehyperref{#1}{Example~\ref*{#1}}}
\newrefformat{fig}{\savehyperref{#1}{Figure~\ref*{#1}}}
\newrefformat{alg}{\savehyperref{#1}{Algorithm~\ref*{#1}}}
\newrefformat{rem}{\savehyperref{#1}{Remark~\ref*{#1}}}
\newrefformat{conj}{\savehyperref{#1}{Conjecture~\ref*{#1}}}
\newrefformat{prop}{\savehyperref{#1}{Proposition~\ref*{#1}}}
\newrefformat{proto}{\savehyperref{#1}{Protocol~\ref*{#1}}}
\newrefformat{prob}{\savehyperref{#1}{Problem~\ref*{#1}}}
\newrefformat{claim}{\savehyperref{#1}{Claim~\ref*{#1}}}

\def\ddefloop#1{\ifx\ddefloop#1\else\ddef{#1}\expandafter\ddefloop\fi}
\def\ddef#1{\expandafter\def\csname 
	bb#1\endcsname{\ensuremath{\mathbb{#1}}}}
\ddefloop ABCDEFGHIJKLMNOPQRSTUVWXYZ\ddefloop
\def\ddef#1{\expandafter\def\csname 
	#1\endcsname{\ensuremath{{\bm{#1}}}}}
\ddefloop ABCDEFGHIJKLMNOPQRSTUVWXYZ\ddefloop
\def\ddef#1{\expandafter\def\csname 
	#1\endcsname{\ensuremath{{\bm{#1}}}}}
\ddefloop abcdefghijklmnopqrstuvwxyz\ddefloop
\def\ddefloop#1{\ifx\ddefloop#1\else\ddef{#1}\expandafter\ddefloop\fi}
\def\ddef#1{\expandafter\def\csname 
	b#1\endcsname{\ensuremath{\mathbf{#1}}}}
\ddefloop ABCDEFGHIJKLMNOPQRSTUVWXYZ\ddefloop
\def\ddef#1{\expandafter\def\csname 
	c#1\endcsname{\ensuremath{\mathcal{#1}}}}
\ddefloop ABCDEFGHIJKLMNOPQRSTUVWXYZ\ddefloop
\def\ddef#1{\expandafter\def\csname 
	f#1\endcsname{\ensuremath{\mathfrak{#1}}}}
\ddefloop ABCDEFGHIJKLMNOPQRSTUVWXYZ\ddefloop
\def\ddef#1{\expandafter\def\csname 
	h#1\endcsname{\ensuremath{\widehat{#1}}}}
\ddefloop ABCDEFGHIJKLMNOPQRSTUVWXYZ\ddefloop
\def\ddef#1{\expandafter\def\csname 
	hc#1\endcsname{\ensuremath{\widehat{\mathcal{#1}}}}}
\ddefloop ABCDEFGHIJKLMNOPQRSTUVWXYZ\ddefloop
\def\ddef#1{\expandafter\def\csname 
	t#1\endcsname{\ensuremath{\widetilde{#1}}}}
\ddefloop ABCDEFGHIJKLMNOPQRSTUVWXYZ\ddefloop
\def\ddef#1{\expandafter\def\csname 
	tc#1\endcsname{\ensuremath{\widetilde{\mathcal{#1}}}}}
\ddefloop ABCDEFGHIJKLMNOPQRSTUVWXYZ\ddefloop

\newcommand{\EE}{\mathbb{E}}

\newcommand{\bzeros}{{\bm{0}}}
\newcommand{\bones}{\bm{1}}

\renewenvironment{proof}{\par\noindent{\bf Proof\ }}{\hfill\BlackBox\\[2mm]}
\newcommand{\BlackBox}{\rule{1.5ex}{1.5ex}}

\newcommand{\trace}{\textrm{trace}}
\newcommand{\overbar}[1]{\mkern 
1.5mu\overline{\mkern-1.5mu#1\mkern-1.5mu}\mkern 1.5mu}

\newcommand{\todoc}[1]{\ignorespaces}

\DeclarePairedDelimiter{\brk}{[}{]}

\DeclarePairedDelimiter{\prn}{(}{)}

\usepackage{nicefrac}

\newsavebox\CBox

\newcommand{\ww}{\w}
\newcommand{\vv}{\mathbf{v}}

\newcommand{\WW}{\W}

\newcommand{\rend}{\hfill $\maltese$}

\newcommand{\dd}{\mbox{$\;|\;$}}

\newcommand{\eps}{\mbox{$\varepsilon$}}
\newcommand{\real}{{\mathbb{R}}}

\newcommand{\defoo}{~\dot{=}~}

\newcommand{\pint}{\mbox{$\mathbb{N}$}}

\newcommand{\arr}{{\rightarrow}}

\newtheorem{prop}[lemma]{Proposition}
\newtheorem{thm}[lemma]{Theorem}

\theoremstyle{definition}

\newtheorem{exam}[lemma]{Example}
\newtheorem{exams}[lemma]{Examples}

\theoremstyle{remark}
\newtheorem{rem}[lemma]{Remark}
\newtheorem{rems}[lemma]{Remarks}

\usepackage{tikz}					
\usepackage{ifthen}

\tikzset{%
	box/.style={
		align=center, minimum size=1cm, inner 
		sep=0pt, font=\tiny\bfseries,
		draw=black, #1
	},
	c1/.style={fill=yellow},
	c2/.style={fill=blue},
	c3/.style={fill=green},
	c3/.style={fill=red},
}

\newcommand{\blockmatrix}[9]{
	\draw[draw=#4,fill=#5] (0,0) rectangle( #1,#2);
	\ifthenelse{\equal{#6}{true}}
	{
		\draw[draw=#7,fill=#8] (0,#2) -- (#9,#2) -- ( #1,#9) -- ( #1,0) -- ( #1 
		- #9,0) -- (0,#2 -#9) -- cycle;
	}
	{}
	\draw ( #1/2, #2/2) node { #3};
}

\newcommand{\mblockmatrix}[4][none]{
	\begin{tikzpicture} 
	\ifthenelse{\equal{#1}{none}}
	{
		\blockmatrix{#2}{#3}{#4}{none}{none}{false}{none}{none}{0.0}
	}
	{
		\definecolor{fillcolor}{rgb}{#1}
		\blockmatrix{#2}{#3}{#4}{none}{fillcolor}{false}{none}{none}{0.0}
	}
	\end{tikzpicture}%this comment is necessary
}

\newcommand{\fblockmatrix}[4][none]{
	\begin{tikzpicture} 
	\ifthenelse{\equal{#1}{none}}
	{
		\blockmatrix{#2}{#3}{#4}{black}{none}{false}{none}{none}{0.0}
	}
	{
		\definecolor{fillcolor}{rgb}{#1}
		\blockmatrix{#2}{#3}{#4}{black}{fillcolor}{false}{none}{none}{0.0}
	}
	\end{tikzpicture}%this comment is necessary
}

\newcommand{\dblockmatrixSD}[6][none]{
	\begin{tikzpicture} 
	\draw[draw=black,fill=none] (0,0) rectangle( #2,#3);
	\definecolor{fillcolor}{rgb}{#1};
	\draw[draw=black,fill=fillcolor] (0,#3) -- (#5,#3) -- ( #2,#5) -- ( #2,0) 
	-- ( #2 - #5,0) -- (0,#3 -#5) -- cycle;
	\draw ( #2/2, #3/2) node { #4};
	\draw ( #2*0.25, #3*0.25) node {#6};
	\draw ( #2*0.75, #3*0.75) node { #6};
	\end{tikzpicture}%this comment is necessary
}

\newcommand{\dblockmatrixSDMO}[7][none]{
	\begin{tikzpicture} 
	\draw[draw=black,fill=none] (0,0) rectangle( #2,#3);
	\definecolor{upperleft}{rgb}{#1};
	\draw[draw=black,fill=upperleft] (0,#3) -- (#5,#3) -- ( #2- #7,#5+#7) -- ( 
	#2 -  #7,#7) -- ( #2 - #5 - #7 ,#7) -- (0,#3 -#5) -- cycle;
	
	\definecolor{lowerleft}{rgb}{1,1,0.8};
	\draw[draw=black,fill=lowerleft] (0,0) rectangle (#2-#7,#7);
	
	\definecolor{upperright}{rgb}{0.8,1,0.8};
	\draw[draw=black,fill=upperright] (#2-#7,#7) rectangle (#2,#3);
	
	\definecolor{lowerright}{rgb}{0.8,0.8,1};
	\draw[draw=black,fill=lowerright] (#2-#7,0) rectangle (#2,#7);
	\draw ( #2*0.5-#7*0.5, #3*0.5+#7*0.5) node { #4};
	\draw ( #2*0.25-#7*0.25, #3*0.25 + #7*0.75) node {#6};
	\draw ( #2*0.75 - #7*0.75, #3*0.75 + #7*0.25) node { #6};
	\draw ( #2*0.5 - #7*0.5, #7*0.5) node {$\gamma$};
	\draw ( #2 - #7*0.5, #3*0.5+ #7*0.5) node {$\delta$};
	\draw ( #2 - #7*0.5, #7*0.5) node {$\epsilon$};
	\end{tikzpicture}%this comment is necessary
}

\newcommand{\dblockmatrixSDMOOO}[7][none]{
	\begin{tikzpicture} 
	\draw[draw=black,fill=none] (0,0) rectangle( #2,#3);
	\definecolor{upperleft}{rgb}{#1};
	\draw[draw=black,fill=upperleft] (0,#3) -- (#5,#3) -- ( #2- #7,#5+#7) -- ( 
	#2 -  #7,#7) -- ( #2 - #5 - #7 ,#7) -- (0,#3 -#5) -- cycle;
	
	\definecolor{lowerleft}{rgb}{1,1,0.8};
	\draw[draw=black,fill=lowerleft] (0,0) rectangle (#2-#7,#7);
	
	\definecolor{upperright}{rgb}{0.8,1,0.8};
	\draw[draw=black,fill=upperright] (#2-#7,#7) rectangle (#2,#3);
	
	\definecolor{lowerright}{rgb}{0.8,0.8,1};
	\draw[draw=black,fill=lowerright] (#2-#7,0) rectangle (#2,#7);
	\draw ( #2*0.5-#7*0.5, #3*0.5+#7*0.5) node { #4};
	\draw ( #2*0.25-#7*0.25, #3*0.25 + #7*0.75) node {#6};
	\draw ( #2*0.75 - #7*0.75, #3*0.75 + #7*0.25) node { #6};
	\draw ( #2*0.5 - #7*0.5, #7*0.5) node {$\gamma$};
	\draw ( #2 - #7*0.5, #3*0.5+ #7*0.5) node {$\delta$};
	\definecolor{lowerdia}{rgb}{0.8,0.8,0.8};
	\draw[draw=black,fill=lowerdia] (#2-#7,#7) -- (#2-#7+#5,#7) -- ( #2,#5) 
	-- ( 
	#2 ,0) -- ( #2 - #5  ,0) -- (#2-#7,#7 -#5) -- cycle;
	\draw ( #2 - #7*0.5, #7*0.5) node {$\epsilon$};
	\draw ( #2-#7*0.25 , #7*0.75) node {$\zeta$};
	\draw ( #2 - #7*0.75 , #7*0.25) node { $\zeta$};
	\end{tikzpicture}%this comment is necessary
	
}

\newcommand{\diagonalblockmatrix}[5][none]{
	\begin{tikzpicture} 
	
	\ifthenelse{\equal{#1}{none}}
	{
		\blockmatrix{#2}{#3}{#4}{black}{none}{true}{black}{none}{#5}
	}
	{
		\definecolor{fillcolor}{rgb}{#1}
		\blockmatrix{#2}{#3}{#4}{black}{none}{true}{black}{fillcolor}{#5}
	}
	
	\end{tikzpicture}%necessary comment
}

\newcommand{\valignbox}[1]{
	\vtop{\null\hbox{#1}}% necessary comment
}

\newenvironment{blockmatrixtabular}
{% necessary comment
	\begin{tabular}{
			@{}c@{}c@{}c@{}c@{}c@{}c@{}c@{}l@{}l@{}l@{}l@{}l@{}l@{}l@{}l@{}l@{}l@{}l@{}l
			@{}c@{}c@{}c@{}l@{}l@{}l@{}l@{}l@{}l@{}l@{}l@{}l@{}l@{}l@{}l@{}l@{}l@{}l@{}l
			@{}l@{}l@{}l@{}l@{}l@{}l@{}l@{}l@{}l@{}l@{}l@{}l@{}l@{}l@{}l@{}l@{}l@{}l@{}l
			@{}
		}
	}
	{
	\end{tabular}%necessary comment
}

\newcommand{\mat}[2]{M\prn*{#1,#2}}
\newcommand{\ploss}{{\cL}} 

\newcommand{\is}[1]{{\mathbf{#1}}}
\newcommand{\VV}{{\is{V}}}
\newcommand{\cs}{\mathfrak{cos}}
\newcommand{\ct}{\mathfrak{cot}}
\newcommand{\sn}{\mathfrak{sin}}
\newcommand{\RRef}[1]{(\ref{#1})}
\newcommand{\vb}{\mbox{{\small 
$\;\veebar\,$}}}

\newcommand{\ibr}[1]{{[#1]}}
\newcommand{\GLG}[1]{{{\text{GL}(#1,\real)}}}
\newcommand{\Od}[1]{{{\text{O}(#1)}}}

\newtheorem{innercustomthm}{Theorem}
\newenvironment{customthm}[1]
{\renewcommand\theinnercustomthm{#1}\innercustomthm}
{\endinnercustomthm}

\usepackage[final]{pdfpages}

\definecolor{darkblue}{rgb}{0, 0, 0.6}

\PassOptionsToPackage{numbers}{natbib}
\usepackage{parskip}
\usepackage[utf8]{inputenc} % allow utf-8 
\def\ddefloop#1{\ifx\ddefloop#1\else\ddef{#1}\expandafter\ddefloop\fi}
\def\ddef#1{\expandafter\def\csname 
	bb#1\endcsname{\ensuremath{\mathbb{#1}}}}
\ddefloop ABCDEFGHIJKLMNOPQRSTUVWXYZ\ddefloop
\def\ddef#1{\expandafter\def\csname 
	#1\endcsname{\ensuremath{{\bm{#1}}}}}
\ddefloop ABCDEFGHIJKLMNOPQRSTUVWXYZ\ddefloop
\def\ddef#1{\expandafter\def\csname 
	#1\endcsname{\ensuremath{{\bm{#1}}}}}
\ddefloop abcdefghijklmnopqrstuvwxyz\ddefloop
\def\ddefloop#1{\ifx\ddefloop#1\else\ddef{#1}\expandafter\ddefloop\fi}
\def\ddef#1{\expandafter\def\csname 
	b#1\endcsname{\ensuremath{\mathbf{#1}}}}
\ddefloop ABCDEFGHIJKLMNOPQRSTUVWXYZ\ddefloop
\def\ddef#1{\expandafter\def\csname 
	c#1\endcsname{\ensuremath{\mathcal{#1}}}}
\ddefloop ABCDEFGHIJKLMNOPQRSTUVWXYZ\ddefloop
\def\ddef#1{\expandafter\def\csname 
	f#1\endcsname{\ensuremath{\mathfrak{#1}}}}
\ddefloop ABCDEFGHIJKLMNOPQRSTUVWXYZ\ddefloop
\def\ddef#1{\expandafter\def\csname 
	h#1\endcsname{\ensuremath{\widehat{#1}}}}
\ddefloop ABCDEFGHIJKLMNOPQRSTUVWXYZ\ddefloop
\def\ddef#1{\expandafter\def\csname 
	hc#1\endcsname{\ensuremath{\widehat{\mathcal{#1}}}}}
\ddefloop ABCDEFGHIJKLMNOPQRSTUVWXYZ\ddefloop
\def\ddef#1{\expandafter\def\csname 
	t#1\endcsname{\ensuremath{\widetilde{#1}}}}
\ddefloop ABCDEFGHIJKLMNOPQRSTUVWXYZ\ddefloop
\def\ddef#1{\expandafter\def\csname 
	tc#1\endcsname{\ensuremath{\widetilde{\mathcal{#1}}}}}
\ddefloop ABCDEFGHIJKLMNOPQRSTUVWXYZ\ddefloop
\usepackage[normalem]{ulem}

\usepackage{hyperref}       % hyperlinks
\usepackage{url}            % simple URL 
\usepackage{booktabs}       % 
\usepackage{amsfonts}       % blackboard 
\usepackage{nicefrac}       % compact 
\usepackage{microtype}      % microtypography

\usepackage{array,multirow}
\title{Analytic Characterization of the 
Hessian\\
	in Shallow 
	ReLU Models:
	A Tale of Symmetry}

	\author{
	Yossi Arjevani \\
	NYU\\
	\texttt{yossi.arjevani@gmail.com} \\
	\and
	Michael Field\\                                 %
	UCSB\\
	\texttt{mikefield@gmail.com}\\
}
\date{}

\begin{document}
\maketitle

\begin{abstract}
We consider the optimization problem associated with fitting 
two-layers ReLU networks with respect to the squared loss, where labels 
are generated by a target network. We leverage the rich 
symmetry structure to analytically characterize the Hessian 
at various families of spurious minima in the natural regime 
where the number of inputs $d$ and the number of hidden neurons 
$k$ is finite. In 
particular, we prove that for $d\ge k$ standard Gaussian inputs: (a) of the 
$dk$ eigenvalues of the Hessian, $dk - O(d)$ concentrate near zero, 
(b) $\Omega(d)$ of the eigenvalues 
grow linearly with $k$. Although this phenomenon of extremely skewed spectrum 
has been observed many times before, to our knowledge, this is 
the first time it has been established {rigorously}. Our analytic 
approach uses techniques, new to the field, from symmetry 
breaking and representation theory, and carries important 
implications for our ability to argue about statistical 
generalization through local~curvature.
\end{abstract}
%	
	% !TEX root = sym_hess.tex

\section{Introduction}
Much of the current effort in understanding the empirical success of 
artificial neural networks is concerned with the geometry of the associated 
nonconvex optimization landscapes. Of particular importance is 
the Hessian spectrum which characterizes the local curvature 
of the loss at different points in the space. This, in turn, 
allows one to closely examine the dynamics of stochastic first 
order methods
\cite{zhu2019anisotropic,ghorbani2019investigation}, design 
potentially better optimization methods
\cite{hochreiter1997flat,chaudhari2019entropy}, and 
argue about various challenging aspects of the network generalization 
capabilities 
\cite{keskar2016large,jastrzkebski2017three,dinh2017sharp}. 
Unfortunately, the 
excessively high cost involved in an exact computation of the Hessian 
spectrum renders this task prohibitive already for 
moderate-sized problems. 

Existing approaches for addressing this computational barrier use 
numerical methods for approximating the Hessian spectrum 
\cite{ghorbani2019investigation,papyan2018full}, 
study the limiting spectral density of shallow models w.r.t. \emph{randomly} 
drawn weights 
\cite{pennington2017nonlinear,pennington2018spectrum,louart2018random},
 or employ various simplified indirect curvature metrics 
\cite{goodfellow2014qualitatively,keskar2016large,dinh2017sharp,li2018visualizing,draxler2018essentially}.
Notably, none of these techniques is able to yield an 
analytic characterization of the Hessian at critical points in 
high-dimensional spaces.  

%None of the current is able to provide analytically study the Spectral density 
%at 
%critical points 
%Given these limitations and the current scattered understanding of the problem 
%geometry, it would appear that we still lack theoretical tools which could 
%facilitate a more systematic analysis of the Hessian.
% of the 
%Hessian spectrum over models with large number of parameters. 

In this paper, we develop a novel approach for studying the Hessian in a class 
of student-teacher (ST) models. Concretely, we focus on 
the squared loss of fitting the ReLU network $\x \mapsto 
\bones^\top_k\phi(\W\x)$, 
\begin{align} \label{def:ploss}
\ploss(\W) \defoo 
\frac{1}{2}\EE_{\x\sim\cN(\bzeros,I_d)}\brk*{ 
	\prn{	 \bones^\top_k\phi(\W \x)- \bones^\top_k\phi(\V\x) 
	}^2  
},\quad \W\in\mat{k}{d}, 
\end{align}
%where 
%$\x \mapsto 
%\bones^\top_k\phi(W\x)$, where $\x \in \RR^k, W\in\mat{k}{k}$, 
where $\phi(z)\defoo \max\{0,z\}$ is the ReLU activation acting 
coordinate-wise, 
$\bones_k$ is the $k$-dimensional vector of all ones, 
$\mat{k}{d}$ denotes the space of all $k\times d$ matrices, 
and $\V\in M(k,d)$ denotes the weight matrix of the target network.
%Given the current limited state of our theoretical understanding, 
The ST framework offers a clean venue for analyzing 
optimization- 
and generalization-related aspects of neural network models, 
and has 
consequently enjoyed 
a surge of interest in recent years, e.g., 
\cite{brutzkus2017globally,du2017gradient,li2017convergence,feizi2017porcupine,
zhang2017electron,ge2017learning,tian2017analytical,safran2017spurious},
 to name a few. 
Perhaps surprisingly, already for this simple model, the rich 
and 
perplexing  geometry of the induced nonconvex optimization 
landscape seems to be out of reach of existing analytic 
methods. 

% by eliminating various 
%expressiveness considerations w.r.t. canonical machine learning 
%datasets---indeed, the trainable 
%weights $\W$ are, by 
%construction, sufficiently expressive to capture the underlying distribution. 
%Our approach for studying the Hessian spectral density is fundamentally 
%different from existing approaches in that it is intimately connected with the 
%invariance properties exhibited by (\ref{def:ploss}). The rich symmetry 
%structure allows a whole new set of powerful analytic and algebraic techniques 
%which we use for deriving various properties of the Hessian spectrum, as we 
%now 
%detail.
%\paragraph{Symmetry-based Framework}
\begin{figure}[H]
	\begin{center}
		\begin{minipage}{0.32\textwidth}
			\includegraphics[scale=0.35]{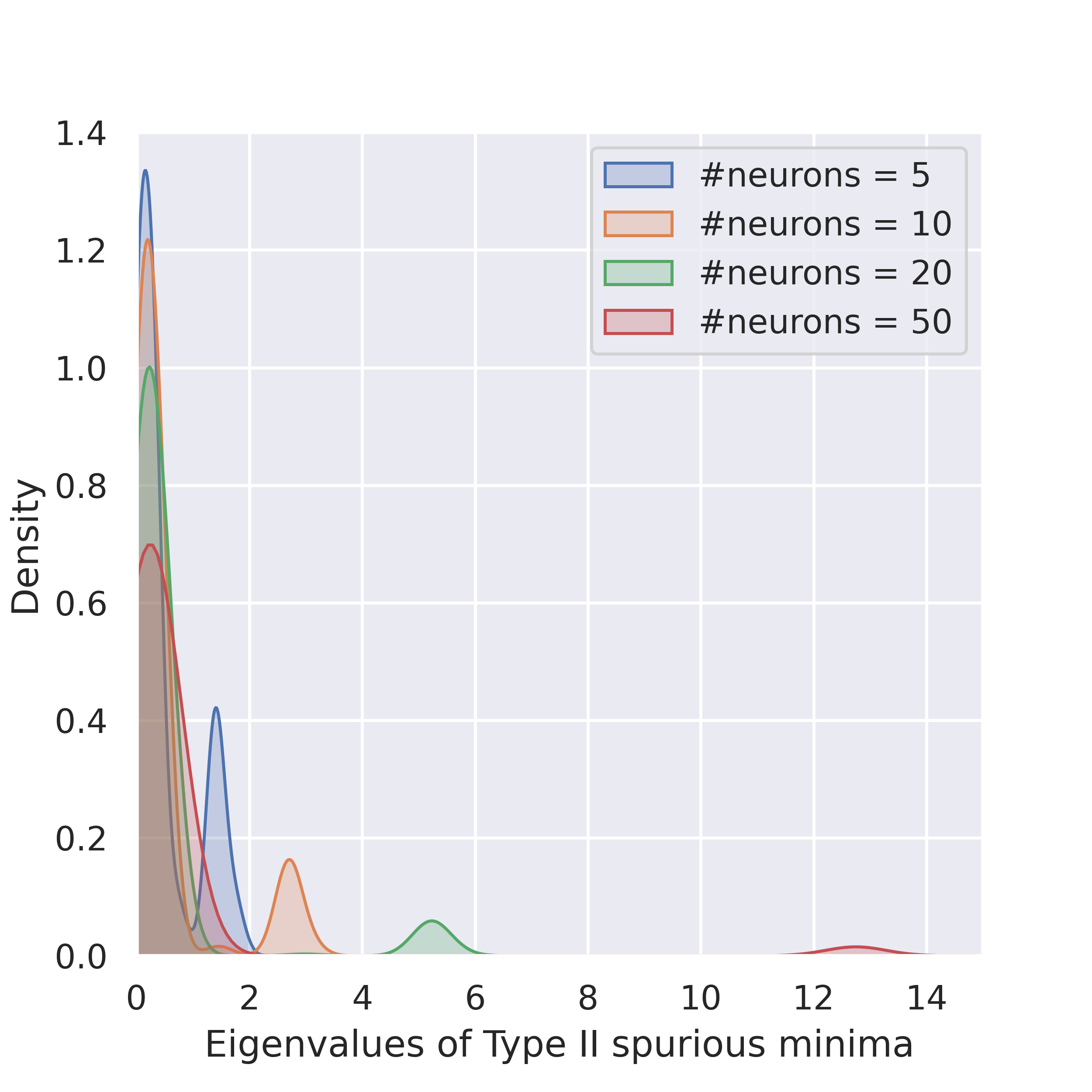}
		\end{minipage}
		\begin{minipage}{0.32\textwidth}
			\includegraphics[scale=0.35]{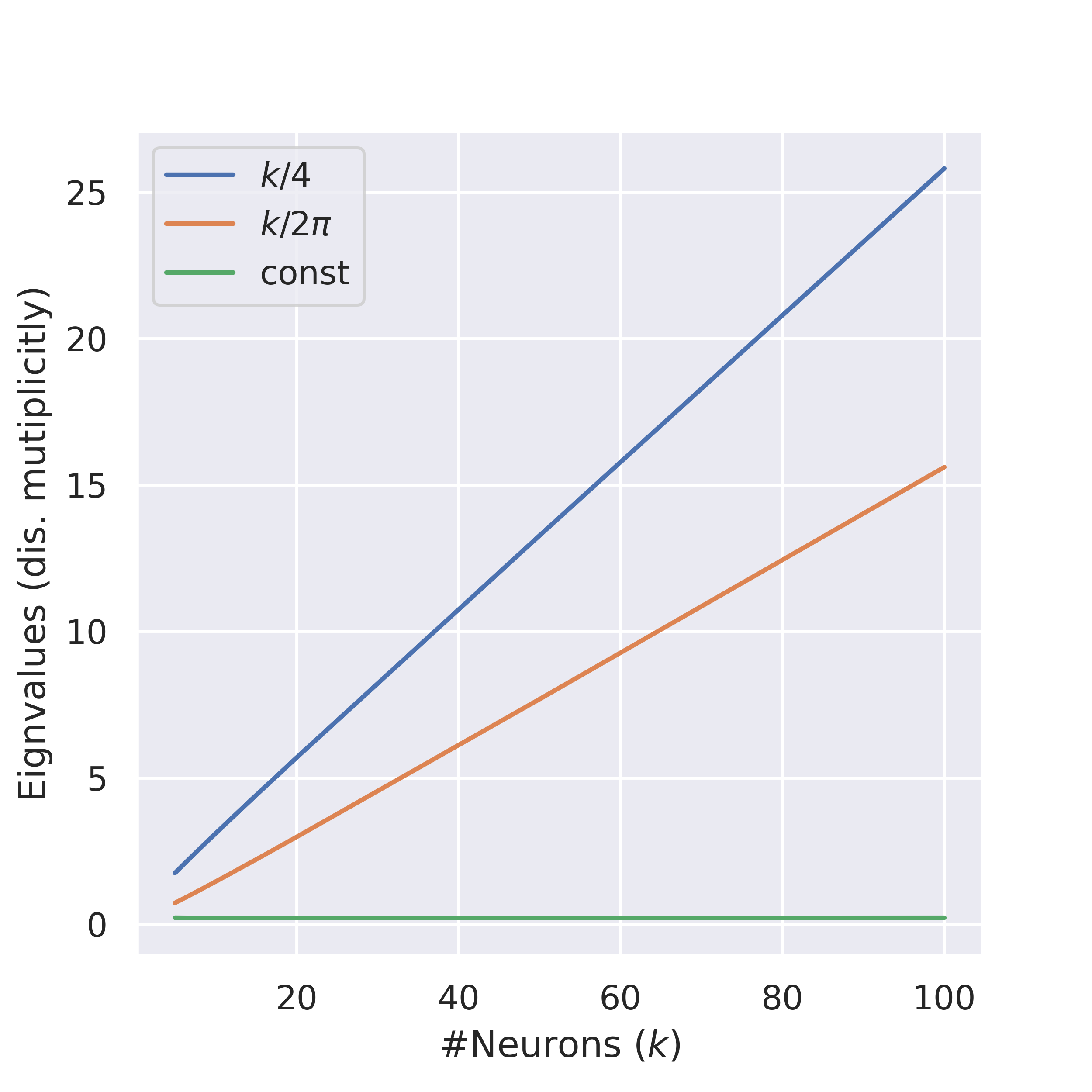}
		\end{minipage}
		\begin{minipage}{0.32\textwidth}
			\includegraphics[scale=0.35]{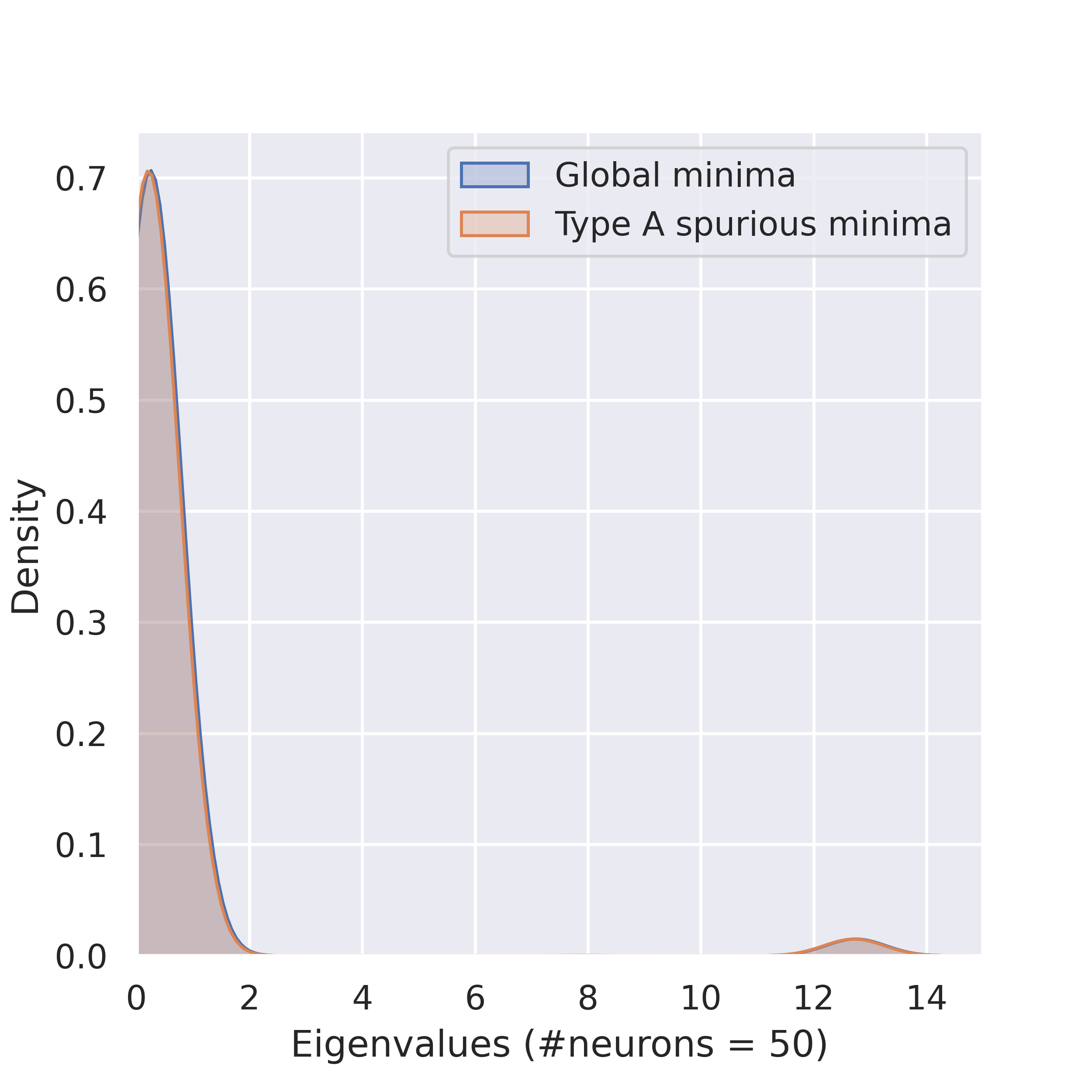}
		\end{minipage}
	\end{center}	
	\caption{(Left) in congruence with \pref{thm: 
			global}, $1-\Theta(1/k)$ 
		fraction  of the spectral density at type II spurious 
		minima 
		concentrates   around $1/4 \pm 1/2\pi$ as the number
		of neurons  $k$ grows simultaneously with the number 
		of inputs. The remaining $\Theta(1/k)$ 
		fraction consists of outliers.  (Middle) examining the 
		spectrum of type II 
		minima (disregarding 
		multiplicity) as $k$ grows confirms the 
		existence of $k+1$ outlier eigenvalues, 
		of which  $k$ grow at a rate 
		of $k/4$ and one at a rate of $k/2\pi$. 
		(Right) the spectra of global 
		minima and type A spurious minima are 
		almost indistinguishable already 
		for $k=50$, thus challenging the  flat minima 
		conjecture.}
	\label{fig:fig1}
\end{figure}
The starting point of our approach is the following simple observation: for any
permutation matrices $\P\in\mat{k}{k},\;\Q\in\mat{d}{d}$, it 
holds that
$\ploss(\P \W \Q^\top)=\ploss(\W)$, for all $\W\in\mat{k}{d}$ \cite[Section 
4.1]{arjevani2019spurious}. It is natural to ask 
how the critical points of $\ploss$ reflect this symmetry.  This question was answered in 
\cite{arjevani2019spurious} where it was shown
that critical points detected by stochastic 
gradient descent (SGD) remain unchanged under transformations of 
the form $\W \mapsto \P\W \Q^\top$ for large groups of pairs of 
permutation matrices $(\P,\Q)$. %This 
%notion of invariance gives rise to an hierarchy among the 
%critical points; 
% the larger the set of $(P,Q)$-pairs is, the more symmetric is 
%the 
%point. In our setting, critical points with maximal symmetry 
%remain fixed under 
%pairs of the form 
%$\Delta (S_p 
%\times S_{q})\defoo \crl*{ (P_p\oplus P_q, P_p^\top\oplus 	
%P_q^\top)~|~P_p\in\mat{p}{p}, P_q \in \mat{q}{q} \text{ are 
%permutation matrices} }$, where $p+q=k$ (modulo 
%conjugation)---we shall make an extensive use of this 
%observation. 
%For example, the point $W=I$, a global minima of $\ploss$, 
%has 
%symmetry $\Delta S_k \defoo \Delta (S_k\times S_0)$, as $I = 
%PIP^\top$ for any 
%permutation matrix $P$. See \pref{fig:max_sym} for more 
%examples.
%These invariance properties of the critical points have 
%far-reaching implications for the structure of their gradients, 
%Hessians, and high-order derivatives.
Using these invariance properties,
%and by combining tools from symmetry breaking, 
%algebraic geometry and analysis,  
families of critical points of $\ploss$ were expressed as power series in 
$1/\sqrt{k}$ leading to, for example, a precise formula for the decay rate of $\ploss$ 
\cite{ArjevaniField2020}. Building on this, we show in this paper how the 
rich symmetry structure can be used to derive an analytic description of the 
Hessian spectral density of $\ploss$, for arbitrarily large, yet finite, 
values 
of~$k$. Having this access to precise high-dimensional 
spectral 
densities, we revisit a number of hypotheses in 
the machine 
learning literature pertaining to curvature, optimization and 
generalization, and establish or refute them \emph{rigorously} 
{for the first time}.

The paper is organized as follows. In \pref{sec: main_results} we state our 
main results and provide discussions aimed at interpreting 
our findings in the light of existing literature. \pref{sec:preliminaries} and 
\pref{sec: submax hessian} are devoted to describing our 
representation 
theory-based approach; all proofs are deferred to the appendix. Lastly,  
detailed empirical corroborations of our analysis are given in 
\pref{sec: 
empirical}.

\section{Main results and related work} \label{sec: main_results}
A formal discussion of our main results requires some familiarity
with group and representation theory. Here, we provide a high-level 
description of our contributions, and defer more detailed statements to later 
sections after the relevant notions have been introduced.

\paragraph{Symmetry-based analysis framework.}
Utilizing the rich symmetry exhibited by neural network models, 
we develop a novel framework for analytically characterizing the 
second-order information of shallow ST ReLU models. 
In its general form, our main result can be stated as 
%%It isn't the most general - that allows for d > k so I've edited a little.
follows.
\begin{theorem}[Informal] \label{thm:info}
Assuming a $k\times k$ orthogonal target matrix $\V$ ($d = k$), the 
spectrum of local minima of $\ploss$ 
consists of a fixed number of distinct eigenvalues (ranging 
between 6 and 22 for high symmetry minima)---independent of 
the number of neurons $k$. Moreover, the spectral distribution is 
massively concentrated in a small number of eigenvalues (ranging 
between 2 and 4 for high symmetry minima) which accounts 
for $k^2-O(k)$ of the spectrum. Similar results hold if $d > k$.
\end{theorem}
The theorem is a consequence of the unique \emph{isotypic}
decomposition of the Hessian that derives from the invariance 
properties of $\ploss$ (see \pref{thm: isot2_body}). Using 
stability arguments, it follows that upon convergence, the spectral density is 
expected to accumulate in clusters whose number does not depend on $k$.
This is confirmed by empirical results which we provide in section \ref{sec: 
emp_pert}.
% 
%of local 
%minima should accumulate in clusters whose number does not 
%depend on $k$
%similar phenomenon to hold
%. That is, the spectral density of local 
%minima should accumulate in clusters whose number does not 
%depend on $k$. We refer to \pref{sec: emp_pert} for an empirical 
%study
%which corroborates our analysis (and see \pref{fig:fig1}). 

Next, we instantiate our framework to the global minima and three
families 
of spurious local minima introduced in \cite{ArjevaniField2020}, 
referred to as types A, I and II (type II corresponds to the 
spurious 
minima described for $6 \le k \le 20$ 
in~\cite{safran2017spurious}).  
A complete description of the minima is provided in \pref{lem: 
asym} 
(type II) and in Section \pref{section: spec} (type A and I).

%---Type II 
%local minima 
%correspond to the spurious minima described for $6 \le k \le 20$ 
%in~\cite{SafranShamir2018}), 

%, we are able to 
%provide \emph{asymptotics} for various eigenvalues-related 
%quantities. 
%Our analysis shows that while 
%$k^2-O(k)$ eigenvalues concentrates around zero, some 
%of the remaining eigenvalues must grow linearly with $k$. 
%\newpage
\begin{theorem}\label{thm: global}
	Assuming a $k\times k$ orthogonal target matrix $\V$, and $k 
	\ge	6$,
	\begin{enumerate}[leftmargin=*]
		\item $\nabla^2 \ploss$ at $\W=\V$ has 6 
		distinct strictly positive eigenvalues:
		\begin{enumerate}
		\item $\frac{1}{4} - 	\frac{1}{2\pi}$ of multiplicity 
		$\frac{k(k-1)}{2}$.
		\item  $\frac{1}{4} + \frac{1}{2\pi}$ of multiplicity 
		$\frac{k(k-3)}{2}$.
		\item 	$\frac{k+1}{4} + O(k^{-1})$ and $ 
		\frac{1}{4} + O(k^{-1})$ of multiplicity $k-1$.
		\item $\approx -0.3471+\frac{k}{2\pi}  + O(k^{-1})$ and 
		$\approx 0.8471+\frac{k}{4} + O(k^{-1}) $ of 
		multiplicity one.
		\item The objective value is $0$.
		\end{enumerate}  
	\item $\nabla^2 \ploss$ at type A spurious local minima 
		has 7 distinct strictly positive eigenvalues:
		\begin{enumerate}
			\item $\frac{1}{4} - \frac{1}{2\pi} - 
			\frac{1}{\pi \sqrt{k}} + 
			O(k^{-1})$ of multiplicity $\frac{(k-1)(k-2)}{2}$.
			\item $\frac{1}{4} + \frac{1}{2\pi} - 
			\frac{1}{\pi \sqrt{k}} + O(k^{-1})$ of multiplicity 
			$\frac{k(k-3)}{2}$.
			\item 3 eigenvalues, $\frac{k+1}{4}+O(k^{-1/2}), 
			\frac{1}{4} 
			+ O(k^{-1/2})$ and $\frac{1}{4}-\frac{1}{2\pi} + 
			O(k^{-1/2})$ of 
			multiplicity $k-1$.
			\item 2 eigenvalues: $c_1+\frac{k}{4}+O(k^{-1/2})$ 
			and 
			$c_2 + \frac{k}{2\pi}+O(k^{-1/2})$ of multiplicity 
			one, 			
			$c_1,c_2>0$.
			\item  The objective value is $(\frac{1}{2}-\frac{1}{\pi}) + 
			O(k^{-1/2})$ \cite{ArjevaniField2020}.
		\end{enumerate}		
		\item $\nabla^2 \ploss$ at type II spurious local minima 
		has 12 distinct strictly positive eigenvalues:
	\begin{enumerate}
		\item $\frac{1}{4} - \frac{1}{2\pi} - 
		\frac{1}{\pi k} + 
		O(k^{-3/2})$ of multiplicity $\frac{(k-2)(k-3)}{2}$.
		\item $\frac{1}{4} + \frac{1}{2\pi} - 
		\frac{1}{\pi k} + O(k^{-3/2})$ of multiplicity 
		$\frac{(k-1)(k-4)}{2}$.
		\item 5 Eigenvalues	of multiplicity $k-2$, of which one grows at a rate 
		of $\frac{k+1}{4}+O(k^{-1})$, and the rest converge to small constants.
		\item 5 Eigenvalues	of multiplicity $1$, of which 2 grow at a rate of 
		$c_3+\frac{k}{4}+O(k^{-1})$, one grows at a rate of $c_4+ 
		\frac{k}{2\pi} + O(k^{-1}),~c_3,c_4>0$,	and the rest converge to 
		small constants.
%		sum is 
%		$\Theta\prn*{\frac{1}{4}k^2+\prn*{\frac{3}{4}-\frac{1}{2\pi}}k}$
%		(in particular, at least one eigenvalue must grow 
%		linearly with $k$),
%		\item 5 Eigenvalues	of 	multiplicity 1 whose 
%		sum is $\Theta\prn*{(\frac{1}{2}+\frac{1}{2\pi})k}$ (in 
%		particular, at least one eigenvalue must grow linearly 
%		with $k$).
		\item The objective value is $(\frac{1}{2}- \frac{2}{\pi^2})k^{-1} +
		O(k^{-3/2})$ 
		\cite{ArjevaniField2020}.
	\end{enumerate}
%	For sufficiently large $k$, each eigenvalue may 
%	be written in the form $ak + b + ek^{-1}$, where $a,b$ are 
%	constants and $e$ is a convergent power series in 
%$1/\sqrt{k}$.\\ 
%	The analysis of type I minima
%	is given in 	\pref{sec: pf_thm_global}.
	If	$d > k$, there will be 2 (resp.~3) additional strictly 
	positive 
	eigenvalues
	for type A (resp.~I or II) minima with total multiplicity 
	$(d-k)k$. The full description, together with that for type I eigenvalues, is given in \pref{sec: pf_thm_global}.
%	One of these eigenvalues grows linearly in $k$ and 
%	contributes 
%	$O(d-k)$ to the spectrum.
%Finally,
%\begin{equation}\label{eq: 1}
%\cal{F}(\mathfrak{c}_k)=(\frac{1}{2} - 
%\frac{2}{\pi^2})k^{-1} + O(k^{-\frac{3}{2}})
%\end{equation}
	\end{enumerate}
\end{theorem}
We note that methods for establishing the existence of spurious 
local 
minima for $\ploss$ are computer-aided and applicable only for 
small-scale problems \cite{safran2017spurious}. Our method establishes the 
existence of spurious local minimum analytically and for 
arbitrarily large $k$ and $d$ (assuming $k\le d$). An additional 
consequence 
of \pref{thm: global} is that \emph{not all 
local minima are alike}. Below, we discuss the implications of the 
similarities and the differences between families of minima of 
$\ploss$.
% in \pref{thm: global}. 

%\begin{table}
%	\caption{Eigenvalues}
%	\label{tbl:delta_evs}
%	\centering
%	\begin{tabular}{lllll}
%		\toprule
%		%		\multicolumn{2}{c}{Part}                   \\
%		%		\cmidrule(r){1-3}
%		& Global minima $W=I$  & Local minima \emph{type A}  & 
%		Local 
%		minima  \emph{type II}     \\ 
%		\midrule
%		Trivial rep.& $\approx 0.16 k$ & & \\
%		& $\approx 0.25 k$ & & \\
%		&  & & \\
%		&  & & \\
%		&  & & \\
%		\midrule
%		Standard rep.
%		& $(k+1)/4$ &&  \\
%		& $1/4$ &&  \\
%		& $1/4-1/2\pi$ &&  \\
%		&  & & \\
%		&  & & \\
%		\midrule
%		$\mathfrak{x}$ rep. & $1/4-1/2\pi$  & & \\
%		\midrule
%		$\mathfrak{y}$ rep. & $1/4+1/2\pi$  & & \\
%		\bottomrule
%	\end{tabular}
%	\caption{Classified into diffr}
%\end{table}
\paragraph{Positively-skewed Hessian spectral density.} 
Although first reported nearly 30 years ago 
\cite{bottou1991stochastic}, 
to the best of our knowledge, this is the first time that this 
phenomenon of extremely skewed spectral density has been 
established \emph{rigorously} for high-dimensional problems (see 
\pref{fig:fig1}). 
Early empirical studies of the Hessian spectrum 
\cite{bottou1991stochastic} revealed 
that local
minima tend to be extremely ill-conditioned. This intriguing 
observation 
was corroborated and further refined in a series of works
\cite{lecun2012efficient,sagun2016eigenvalues,sagun2017empirical}
which studied how the spectrum evolves along the training 
process. It was noticed that, upon convergence, the spectral 
density decomposes into two parts: a bulk of 
eigenvalues concentrated around zero, 
and a small set 
of positive outliers located away from zero. 
%; The 
%former part was claimed to be primarily affected by the network 
%architecture, and the latter by the underlying data 
%distribution. 

Due to the high computational cost of an exact computation of 
the Hessian spectrum ($O(k^3d^3)$ for a $k\times d$ weight matrix), 
this phenomenon of extremely skewed spectral densities has only been confirmed 
for small-scale networks. Other methods for extracting second-order 
information in 
large-scale problems roughly fall into two general categories. The 
first class of 
methods approximate the Hessian spectral density by 
employing various 
numerical estimation techniques, most notably stochastic Lanczos 
method (e.g., \cite{ghorbani2019investigation,papyan2018full}). 
These methods have provided various numerical evidences that
indicate that a similar skewed spectrum phenomenon also occurs in 
full-scale modern neural networks. The second class of 
techniques builds on tools from random matrix theory. This approach yields an 
exact computation of the limiting spectral 
distribution (i.e., the number of neurons is taken to infinity),
assuming the inputs, as well as the model weights are drawn at random 
\cite{pennington2017nonlinear,pennington2018spectrum,louart2018random}.
 In contrast, our method gives an exact description of the 
 spectral density for essentially any (finite) number of neurons, and 
 at critical points rather than randomly drawn weight 
 matrices.		

\paragraph{The flat minima conjecture and implicit bias.} 
It has long been debated whether some notion of local  curvature 
can be 
used to explain the remarkable generalization capabilities of modern 
neural networks 
\cite{hochreiter1997flat, 
keskar2016large,jastrzkebski2017three,wu2017towards,yao2018hessian,chaudhari2019entropy,dinh2017sharp}.
One intriguing hypothesis
%, which has 
%emerged out of the study of the Hessian spectral distributions, 
suggests that minima with wider basins of attraction tend to generalize better. 
%Dating back to 
%\cite{hochreiter1997flat}, the flat-minima phenomenon has been 
%successfully 
%reproduced many times in problems of increasing complexity 
%(\cite{keskar2016large,jastrzkebski2017three,wu2017towards,yao2018hessian,chaudhari2019entropy},
%but see \cite{dinh2017sharp}). 
An intuitive possible explanation is that 
flat minima promote statistical and numerical stability; 
%w.r.t. data perturbations, 
%numerical 
%errors and other sources of noise; 
together with low empirical 
loss, these 
ingredients are widely-used to achieve good generalization, cf. 
\cite{shalev2010learnability}. 

Perhaps surprisingly, our analysis shows that the 
spectra of global minima and the spurious minima considered in 
\pref{thm: global} agree on $k^2-O(k)$ out of $k^2$ eigenvalues to within 
$O(k^{-1/2})$-accuracy ($d=k$). Thus, only the 
remaining $O(k)$ can potentially account for any essential difference in the 
local curvature. However, for type A spurious minima, even the remaining $O(k)$ 
eigenvalues are $O(k^{-1/2})$-far from the spectrum of the 
global minima. Consequently, in our settings, local second-order curvature 
\emph{cannot} be used to separate global minima from spurious minima, thus 
ruling out notions of `flatness' which rely exclusively on the Hessian 
spectrum. Of course, 
other metrics of a `wideness of basins' may well apply. 

Despite being a striking counter-example for a spectral-based notion of 
flatness, we note that, empirically, under Xavier initialization 
\cite{glorot2010understanding}, type A spurious minima are 
rarely detected by SGD \cite{ArjevaniField2020}. This stands in sharp contrast 
to type II minima to 
which SGD converges with a substantial empirical probability. 
Thus, for reasons which are yet to be understood, the bias 
induced by Xavier initialization seems to favor the 
%better-behaved 
class of global and type II minima at which the objective value decays with 
$k$ 
to zero, rather than type~A and type~I minima whose objective value converges 
to strictly positive constants, cf., 
\cite{soudry2018implicit,gunasekar2018implicit}. We leave 
further study of this phenomenon, as well  as other families of 
spurious minima,  to future work.

\paragraph{Proof technique.}  Conceptually, the derivation 
of the eigenvalue estimate in \pref{thm: global} 
is based on ideas originating in symmetry-breaking, equivariant bifurcation 
theory and representation theory. Group invariance properties 
of the loss function (\ref{def:ploss}) imply that the Hessian at 
symmetric points 
(under a proper notion of symmetry) must exhibit a certain 
block structure, and this makes possible an explicit 
computation of the Hessian spectrum. Empirically, and somewhat miraculously, spurious minima of (\ref{def:ploss}) tend to be 
highly symmetric. As a consequence, their Hessian can be simplified
using the same symmetry-based methods.  The reminder of the paper is devoted 
to a 
formal and more detailed  exposition of this approach.

	% !TEX root = sym_hess.tex

\section{The method: a symmetry-based analysis of the 
Hessian}\label{sec:preliminaries}
In order to avoid a long preliminaries section, key ideas and concepts are
introduced and organized so as to illuminate our strategy for 
analyzing the Hessian. We illustrate with reference to the case of 
global minima where $d=k$ and the target weight matrix 
$\V$ is the identity 
$\I_k$.

\subsection{Studying invariance properties via group action}
We first review background material on group actions
and fix notations (see~\cite[Chapters 1, 2]{Field2007} for a 
more complete account). Elementary concepts from group theory 
are assumed known. We start with two examples that are used later.
\begin{exams} \label{exams:groups}
	(1) The \emph{symmetric group} $S_d$, $d\in\pint$, is the group of 
	permutations of $\ibr{d}\defoo \{1,\dots,d\}$.  \\
	(2) Let $\GLG{d}$ denote the space of invertible 
	linear maps on $\real^d$. Under composition, $\GLG{d}$ has the 
	structure of a 
	group. The \emph{orthogonal group} $\text{O}(d)$ is the 
	subgroup 
	of $\GLG{d}$ 
	defined by
	$
	\text{O}(d) =   \{A \in \GLG{d} \dd \|Ax\| = 
	\|x\|,\;\text{for all } x \in \real^d\}.
	$
	Both $\GLG{d}$ and $\text{O}(d)$ can be viewed as groups of 
	invertible $d \times d$ matrices.
\end{exams}
Characteristically, these groups consist of 
\emph{transformations}  
of a set and so we are led to the notion of a 
\emph{$G$-space} $X$ where we have 
an \emph{action} of a group $G$
on a set $X$. Formally, this is a  group homomorphism from $G$ 
to the
group of bijections of $X$.
For example, $S_d$ naturally acts on $[d]$ as permutations and 
both $\GLG{d}$ and $\text{O}(d)$ act on
$\real^d$ as linear transformations (or matrix~multiplication). 

An example, which we use extensively in studying the invariance 
properties of $\ploss$, is given by the action of the
group $S_k \times S_d\subset S_{k\times d},~k,d\in 
\pint$, on
$\ibr{k} \times \ibr{d}$ defined by
\begin{align}\label{eq: Gamma-action}
(\pi,\rho)(i,j) = (\pi^{-1}(i),\rho^{-1}(j)),\; \pi \in 
S_k, 
\rho \in S_d,\; (i,j) \in \ibr{k} \times \ibr{d}.
\end{align}
%For our choice of $\V=\I$, 
%$\cal{L}$ is $S_k \times S_d$-invariant. see \cite{arjevani2019spurious} for 
%a more detailed derivation)
%Indeed, 
%$\mathcal{L}((g,e_d)\WW,(h,e_d)\VV) = \mathcal{L}(\WW,\VV)$,
%for all $g,h \in S_k$ where $e_d \in S_d$ denotes the 
%identity.  Since $S_k$ acts by permuting rows, this is just summing the 
%outputs of the $k$ 
%in different order. Since the 
%underlying distribution is invariant under orthogonal 
%transformations, $\mathcal{L}((e_k,h)\WW,(e_k,h)\VV) 
%= \mathcal{L}(\WW,\VV)$ for all $g\in O(d)$. Now take $d = k$. Since
%$(g,g)\VV = \VV$ for all $g \in S_k$, we see that
%$\cal{F}$ is $S_k$-invariant (diagonal action). That is, 
%$\cF(g \W) = 
%\cF(\W)$, $g \in S_k$ (see \cite{arjevani2019spurious} for 

This action induces an action on the space $M(k,d)$ of $k \times 
d$-matrices $A = [A_{ij}]$ by
$ (\pi,\rho)[A_{ij}] = [A_{\pi^{-1}(i),\rho^{-1}(j)}]$.
The action can be defined in terms of permutation matrices but 
is easier to describe in terms of rows and columns: 
$ (\pi,\rho)A$ permutes rows (resp.~columns) of $A$ according to 
$\pi$ (resp.~$\rho$). As mentioned in the introduction, for 
our choice of $\V=\I_k$, $\ploss$ is $S_k \times S_d$-invariant.
If $d = k$, define the \emph{diagonal subgroup} $\Delta S_{k}$ of $S_k \times 
S_k$ by $\Delta S_{k} = \{(g,g) \dd g \in S_k\}$. Note that 
$\Delta S_{k}\approx S_k$. When we restrict the $S_k \times S_k$-action on 
$M(k,k)$ to $\Delta S_k$, we refer to the diagonal $S_k$-action, or just the $S_k$-action on $M(k,k)$.
This action of $S_{k}$ on $M(k,k)$ maps diagonal matrices to 
diagonal matrices and should not be confused with the actions of $S_k$ on 
$M(k,k)$ defined by either permuting rows or~columns.

\begin{exam}
        Take $p,q \in\pint$, $p+q = k$, and consider the 
        diagonal
action
of $S_{p} \times S_q\subset S_k$ on
$M(k,k)$.  Write $A \in M(k,k)$ in block matrix form as $A =
\left[\begin{matrix} A_{p,p} & A_{p,q}\\
A_{q,p}
&
A_{q,q}
\end{matrix}\right]$.
If $(g,h) \in S_{p} \times S_q\subset S_k$, then $(g,h)A =
\left[\begin{matrix} gA_{p,p} & (g,h)A_{p,q}\\
(h,g)A_{q,p} & hA_{q,q}
\end{matrix}\right]$
where $gA_{p,p}$ (resp.~$hA_{q,q}$) are defined via the 
diagonal
action of $S_p$ (resp.~$S_q$) on $A_{p,p}$ 
(resp.~$A_{q,q}$), and
$(g,h)A_{p,q}$ and $(h,g)A_{q,p}$ are defined
through the natural action of $S_p \times S_q$ on rows and 
columns. Thus,
for $(g,h)A_{p,q}$ (resp.~$(h,g)A_{q,p}$) we permute rows 
(resp.~columns) according to $g$ and
columns (resp.~rows) according to $h$.
In the case when $p=k-1$, $q = 1$,
$S_{k-1}$ will act diagonally on
$A_{k-1,k-1}$, fix
$a_{kk}$, and act by permuting the first $(k-1)$ entries of 
the
last row and column.
\end{exam}
%An irreducible representation $(V,G)$ is \emph{real} if every $G$-map $A: V 
%\arr V$  is a real multiple of the identity. It will important for us that
%every irreducble representation of $S_n$ is 
%real~\cite{James1978,FultonHarris1991}.

Given $\WW \in M(k,k)$, the largest 
subgroup of $S_k \times S_k$ 
fixing $\WW$ is called the \emph{isotropy} subgroup of $\WW$ and is used 
as means of measuring the symmetry of $\WW$. The isotropy subgroup of 
$\VV\in M(k,k)$ is the diagonal subgroup $\Delta S_k$. Our focus 
will be on critical points $\WW$ whose isotropy groups are 
subgroups of the target matrix $\V=\I_k$, that is, $\Delta S_k$ and 
$\Delta S_{k-1}$ (see \pref{fig:max_sym}---we use the notation $\Delta 
S_k$ as the
isotropy is a \emph{subgroup} of $S_k \times S_k$). Other choices of 
target matrices yield different symmetry-breaking of the isotropy of the 
global minima (see \cite{arjevani2019spurious} for more details). In the 
next section, we show how the symmetry of local minima greatly simplifies 
the analysis of their Hessian.
%For our choice of $\V=\I$, 
%$\cal{L}$ is $S_k \times S_d$-invariant. see \cite{arjevani2019spurious} for 
%a more detailed derivation)
%Indeed, 
%$\mathcal{L}((g,e_d)\WW,(h,e_d)\VV) = \mathcal{L}(\WW,\VV)$,
%for all $g,h \in S_k$ where $e_d \in S_d$ denotes the 
%identity.  Since $S_k$ acts by permuting rows, this is just summing the 
%outputs of the $k$ 
%in different order. Since the 
%underlying distribution is invariant under orthogonal 
%transformations, $\mathcal{L}((e_k,h)\WW,(e_k,h)\VV) 
%= \mathcal{L}(\WW,\VV)$ for all $g\in O(d)$. Now take $d = k$. Since
%$(g,g)\VV = \VV$ for all $g \in S_k$, we see that
%$\cal{F}$ is $S_k$-invariant (diagonal action). That is, 
%$\cF(g \W) = 
%\cF(\W)$, $g \in S_k$ (see \cite{arjevani2019spurious} for 
%a more detailed derivation). 
%
%\yacomment{Reduce paragraph above + add here or in last section: Theory 
%	predicts different forms of 
%	critical points 
%	according to LSB. MJF comment. The diagonal subgroup is for quantifying the 
%	symmetry but we
%	need to talk about the $S_k$-action (as opposed to $\Delta S_k$ action) 
%	because we are going to discuss
%	the representation theory of $S_k$ (not $\Delta S_k$)---small point but 
%	otherwise it is confusing.}

\newcommand{\bwidth}{1.2in}
\newcommand{\bheight}{\bwidth}
\newcommand{\dwidth}{0.11in}
\begin{figure}[ht]
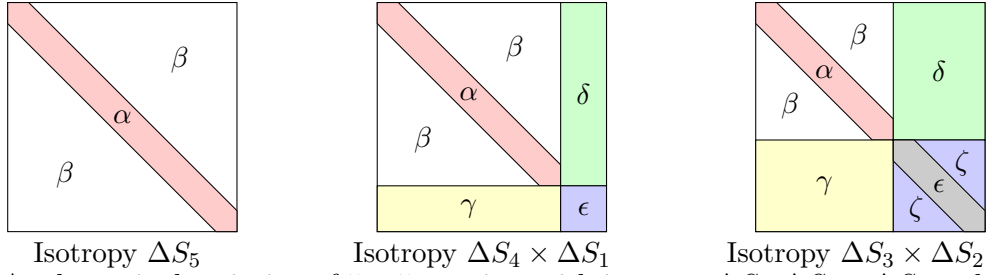
 
	\begin{center}\vskip-0.4cm
		{\setstretch{1.2}
			\begin{blockmatrixtabular}\label{fig:dia_sym}
				
				\valignbox{\dblockmatrixSD[1.0,0.8,0.8]{\bwidth}{\bheight}{$\alpha$}{\dwidth}{$\beta$}}&
				\quad\quad\quad\quad
				
				\valignbox{\dblockmatrixSDMO[1.0,0.8,0.8]{\bwidth}{\bheight}{$\alpha$}{\dwidth}{$\beta$}{0.24in}}&
				\quad\quad\quad\quad
				
				\valignbox{\dblockmatrixSDMOOO[1.0,0.8,0.8]{\bwidth}{\bheight}{$\alpha$}{\dwidth}{$\beta$}{0.48in}}&\\
				Isotropy $\Delta S_5 $& 		 
				\quad\quad\quad\quad 
				Isotropy  $\Delta S_4 \times 
				\Delta 
				S_1$  & 		 \quad\quad\quad\quad Isotropy 
				$\Delta 
				S_3\times \Delta
				S_2$
		\end{blockmatrixtabular}}
	\end{center}
	\vskip-0.4cm		
	\caption{A schematic description of $5\times 5$ matrices 
	with isotropy 	$\Delta S_5, \Delta S_4 \times \Delta S_1$ 
	and $\Delta S_3\times S_2$, from left to right (borrowed 
		from \cite{arjevani2019spurious}). $\alpha, \beta, 
		\gamma,\delta, \epsilon$ and $\zeta$ are assumed to be `sufficiently' 
		different.} 
	\label{fig:max_sym}
\end{figure}

%In future, we refer to this action as the 
%$S_k$-action on $M(k,k)$ (rather than the $\Delta S_k$-action).

\subsection{The spectrum of equivariant linear isomorphisms} 
\label{sec:spec}

If $G$ is a subgroup of $\Od{d}$, the action on $\real^d$ is 
called an \emph{orthogonal} representation of $G$ (we often
drop the qualifier orthogonal). Denote by $(\real^d,G)$ as 
necessary. The \emph{degree} of a representation $(V,G)$ is the 
dimension of $V$ ($V$  will always be a linear subspace of 
some $\real^n$ with the induced Euclidean inner product). The action 
of $S_k \times S_d\subset S_{k\times d}$ on $M(k,d)$  is
{orthogonal} with respect to the standard Euclidean inner 
product on $M(k,d) \approx \real^{k\times d}$ since the action permutes the 
coordinates of $ 
\real^{k\times d}$ (equivalently, components of $k\times d$ matrices). 

Given two representations $(V,G)$ and $(W,G)$, a 
map $A: V \arr W$ is called $G$-equivariant 
if $A(gv) = gA(v)$, for all $g \in G, v \in V$. If $A$ is linear and 
equivariant, we say $A$ 
is a \emph{$G$-map}. Invariant functions naturally provide
examples of equivariant maps. Thus the gradient $\nabla \cF$ is a $S_k \times 
S_d$-equivariant self map 
of $M(k,d)$ and if $\WW$ is a critical point of $\nabla \cF$ with isotropy $G 
\subset S_k \times S_d$, 
then
$\nabla^2 \cF(\W): M(k,d)\arr M(k,d)$ is a $G$-map 
(see~\cite{Field2007,ArjevaniField2020}).
The equivariance of the Hessian is the key ingredient that allows us to 
study the spectral density at \emph{symmetric} local 
minima.

A representation $(\real^n,G)$ is \emph{irreducible} 
if the only linear subspaces of $\real^n$ that are preserved 
(invariant)  by the $G$-action are
$\real^n$ and $\{0\}$. Two orthogonal 
representations $(V,G)$, $(W,G)$ are \emph{isomorphic} (and have the same 
\emph{isomorphism class}) if there 
exists a $G$-map $A: V \arr W$ which is a linear isomorphism. If $(V,G)$, 
$(W,G)$ are irreducible but not isomorphic then
every $G$-map $A: V \arr W$ is zero 
 (as the kernel and the image of a $G$-map are $G$-invariant). If $(V,G)$ is 
 irreducible, then the space $\text{Hom}_G(V,V)$ of $G$-maps (endomorphisms) 
 of $V$ is a real associative division algebra and is isomorphic by a theorem 
 of Frobenius to either $\real, \mathbb{C}$ or $\mathbb{H}$ (the quaternions). 
 The \emph{only} case that will concern us here is when $\text{Hom}_G(V,V) 
 \approx \real$ when we say the representation is \emph{real}.
\begin{exam}\label{ex: iso}
	Let $n > 1$. Take the natural (orthogonal) action of $S_n$ on $\real^n$ 
	defined by permuting coordinates.  The representation is not 
	irreducible since the subspace $T = 
	\{(x,x,\cdots,x)\in\real^n \dd x \in \real\}$ 
	is 	invariant by the action of $S_n$, as is the hyperplane 
	$H_{n-1} 
	= T^\perp =\{(x_1,\cdots,x_n)\dd \sum_{i\in\ibr{n}}x_i =0\}$.
	It is easy to check that $(T,S_n)$, also called the 
	\emph{trivial }representation of 
	$S_n$, and $(H_{n-1},S_n)$, the 
	\emph{standard} representation, are irreducible, real, and 
	not isomorphic.	
%	Their isomorphism classes are denoted by $\mathfrak{t}$ and 
%	$\mathfrak{s}_{k}$, respectively (for all $k \ge 2$, $\mathfrak{t}$ is 
%1-dimensional).
\end{exam}
%An irreducible representation $(V,G)$ is \emph{real} if every $G$-map $A: V 
%\arr V$  is a real multiple of the identity. It will important for us that
%every irreducble representation of $S_n$ is 
%real~\cite{James1978,FultonHarris1991}.
%	
Every representation $(\real^n,G)$ can be written uniquely, up to order, as an orthogonal
direct sum $\oplus_{i\in\ibr{m}} V_i$, where each $(V_i,G)$ is an orthogonal 
direct sum of isomorphic irreducible representations $(V_{ij},G)$, $j \in 
\ibr{p_i}$, and $(V_{ij},G)$ is isomorphic to $(V_{i'j'},G)$
if and only if $i' = i$. The subspaces $V_{ij}$ are \emph{not} uniquely determined if $p_i > 1$. 
If there are $m$ distinct isomorphism classes $\mathfrak{v}_1,\cdots,\mathfrak{v}_m$ of irreducible representations, then $(\real^n,G)$
may be represented by the sum $p_1 \mathfrak{v}_1 + \cdots + p_m \mathfrak{v}_m$, where $p_i\ge 1$ counts the number of representations with isomorphism class $\mathfrak{v}_i$. Up to order, this sum (that is, the $\mathfrak{v}_i$ and their multiplicities) is uniquely determined by $(\real^n,G)$. This is the \emph{isotypic decomposition} of $(\real^n,G)$ (see \cite{thomas2004representations} and 
Section~\ref{sec: B}). The isotypic decomposition is a powerful tool for extracting information about the
spectrum of $G$-maps. 
%Note that a $G$-map $A: V_i \arr V_{i'}$ is zero unless $i = i'$. 

%Using the orthogonality of the action, 
%every representation $(\real^n,G)$ can be written as an orthogonal             
%direct sum of irreducible representations. That is, $\real^n = \oplus_{j\in\ibr{q}} V_j$, where each %$(V_j,G)$ is irreducible. The
%subspaces $V_j$ are generally \emph{not} uniquely determined, but sometimes there are natural choices.
%If there are $m$ distinct isomorphism classes 
%$\mathfrak{v}_1,\cdots,\mathfrak{v}_m$, in $\{(V_i,G)\}_{i\in\ibr{m}}$,
%then $(\real^n,G)$ may be represented by the sum $p_1 \mathfrak{v}_1+ \cdots + p_m\mathfrak{v}_m$, where %$p_i$ counts the number of
%representations $(V_j,G)$ which have isomorphism class $\mathfrak{v}_i$.  Up to order, this sum (that is, %the $\mathfrak{v}_i$ and their multiplicities $p_i$)
%is uniquely determined by $(\real^n,G)$.
%This is the \emph{isotypic decomposition} of $(\real^n,G)$ (see 
%\cite{thomas2004representations}). 

%The isotypic decomposition is a powerful tool for 
%extracting information  about the spectrum of $G$-maps.

If $G = S_k$, then every irreducible representation of $S_k$ is real~\cite[Thm.~4.3]{fulton1991representation}. Suppose, as above, that 
$(\real^n,S_k)=\oplus_{i\in\ibr{m}} V_i$ and $A: \real^n\arr\real^n$ is an $S_k$-map. 
Since the induced maps $A_{ii'}: V_i \arr V_{i'}$ must be zero if $i \ne i'$, $A$ is uniquely determined by
the $S_k$-maps $A_{ii}: V_i \arr V_i$, $i \in\ibr{m}$. Fix $i$ and choose an $S_k$-representation $(W,S_k)$
in the isomorphism class $\mathfrak{v}_i$. Choose $S_k$-isomorphisms $W \arr V_{ij}$, $j \in \ibr{p_i}$.
Then $A_{ii}$ induces $\overline{A}_{ii}: W^{p_i} \arr W^{p_i}$ and so determines a (real) matrix $M_i \in M(p_i,p_i)$ since $\text{Hom}_{S_k}(W,W) \approx \real$. Different choices of $V_{ij}$, or isomorphism $W \arr V_{ij}$, yield a matrix similar to $M_i$. Each eigenvalue of $M_i$ of multiplicity $r$ gives an eigenvalue of $A_{ii}$, and so of $A$, of multiplicity 
$r\,\text{degree}(\mathfrak{v}_i)$.

%Suppose that $(\real^n,G)$ has isotypic 
%decomposition $\sum_{i \in\ibr{m}}p_i \mathfrak{v}_i$ and  $A:  
%\real^n \arr \real^n$ is a $G$-map. Then $A$
%induces a map $\overline{A}:\sum_{i \in\ibr{m}}p_i \mathfrak{v}_i\arr \sum_{i \in\ibr{m}}p_i \mathfrak{v}_i$ 
%that may be written in block diagonal form as $[A_{ii}]_{i \in \ibr{m}}$, where $A_{ii}:   p_i %\mathfrak{v}_i\arr p_i\mathfrak{v}_i$,
%$i \in \ibr{m}$ (every $G$-map between inequivalent irreducible $G$-representations must be zero).
%It is known that \emph{every} irreducible $S_k$-representation is 
%real. That is, if $B: V_i \arr V_i$ is an 
%$S_k$-map, then $B = c I_{V_i}$, where $c \in \real$
%(this is not true for general $G$).
%From this it follows that
%$A_{ii}$ can be represented as a real $p_i \times p_i$-matrix $M_i$. If $p_i > 1$, the matrix $M_i$, but not %the eigenvalues, depends on the
%decomposition $\real^n = \oplus_{j\in\ibr{q}} V_j$. 
\begin{fact}\label{fact: iso} (Notations and assumptions as 
	above.)
	If $A$ is the Hessian, all eigenvalues are real and
	each eigenvalue of $M_{i}$ of multiplicity $r$ will be an eigenvalue of $A$ with 
	multiplicity $r\,\text{degree}(\mathfrak{v}_i)$. In 
	particular, $A$ has most $\sum_{i \in \ibr{m}}p_i$ distinct 
	real eigenvalues---regardless of the dimension of the underlying 
	space.
\end{fact}

Our strategy can be now summarized as follows. Given a local minima 
$\W$, we compute the isotropy 
group $G \subset S_k \times S_d$ of $\WW$. Since the 
Hessian of $\cF$ at $\W$ is a $G$-map, may use the 
isotypic decomposition of the action of $G$ on $M(k,d)$ to extract the 
spectral properties of the Hessian. In our setting, local minima have 
large isotropy groups, typically, as large as $\Delta(S_p 
\times S_{k-p}),~0\le p <k/2$. Studying the Hessian at these 
minima requires the isotopic decomposition corresponding to 
$\Delta(S_p \times S_{k-p}),~0\le p <k/2$, which we detail in 
\pref{thm: isot2_body} below.

\subsection{The isotypic decomposition of $(M(k,k),S_k)$ 
and the spectrum at $\mathbf{W}=\VV$} \label{sec: iso_global}
\newcommand{\ft}{\mathfrak{t}}
\newcommand{\fs}{\mathfrak{s}}
\newcommand{\fx}{\mathfrak{x}}
\newcommand{\fy}{\mathfrak{y}}
Regard $M(k,k)$ as an $S_k$-space (diagonal action). The trivial 
representation, denoted by $\ft_k$, and the standard representation, denoted by 
$\fs_k$, introduced in 
\pref{ex: iso} are examples of the many irreducible 
representations of $S_k$. In the general theory, each irreducible 
representation of $S_k$ is associated to a partition of the set $[k]$. 
The description of the isotypic decomposition of $(M(k,k),S_k)$
is relatively simple  %%(provided in \pref{sec: full_iso_decomp} for $k\ge 4$),  Surely this is done below?
and uses just 4 irreducible representations of $S_k$ for $k \ge 4$.
\begin{itemize}
\item The trivial representation $\mathfrak{t}_k$ of degree 
1.
%, associated to the partition $(k)$
\item The standard representation $\mathfrak{s}_k$ of $S_k$ of degree $k-1$.
%, associated to the partition $(k-1,1)$
\item The exterior square representation 
$\mathfrak{x}_k=\wedge^2 
\mathfrak{s}_k$ of degree $\frac{(k-1)(k-2)}{2}$.
%, associated to the partition $(k-2,1,1)$
\item A representation $\mathfrak{y}_k$ of degree 
$\frac{k(k-3)}{2}$. We describe $\mathfrak{y}_k$ explicitly later in terms of 
symmetric matrices (formally, it is the representation associated to the 
partition $(k-2,2)$).

%, associated to the partition $(k-2,2)$
\end{itemize}
We omit the subscript $k$ when clear from the context. Assume that $k\ge4$. We 
begin with a 
well-known result about the 
representation 
$\mathfrak{s}\otimes\mathfrak{s}$ 
(see, e.g., \cite{fulton1991representation}). If 
$\mathfrak{s}\odot \mathfrak{s}$ denotes the symmetric tensor 
product of $\mathfrak{s}$, then
\begin{equation}\label{eq: ten}
\mathfrak{s}\otimes\mathfrak{s} = \mathfrak{s}\odot \mathfrak{s} 
+ \mathfrak{x} = \mathfrak{t}+  \mathfrak{s} + \mathfrak{y} + 
\mathfrak{x}.
\end{equation}
Since all the irreducible $S_k$-representations are real, they are isomorphic to 
their dual 
representations and so we have the isotypic decomposition
\begin{eqnarray}\label{eq: Miso1}
M(k,k)&\approx&\real^k \otimes \real^k \approx 
(\mathfrak{s}+\mathfrak{t})  \otimes (\mathfrak{s}+\mathfrak{t}) 
%%\nonumber\\
\label{eq: Miso2}
 =  2\mathfrak{t} + 3\mathfrak{s} + \mathfrak{x} + 
\mathfrak{y},
\end{eqnarray}
since $\mathfrak{t}  \otimes \mathfrak{s} = \mathfrak{s}$ and 
$\mathfrak{t}  \otimes\mathfrak{t} =\mathfrak{t} $.

%Using Fact \ref{fact: iso}, information can immediately be deduced from  
%Equation \pref{eq: Miso1}. For any local 
%minima (or critical point) of isotropy $\Delta S_k$, for 
%example, $\W=\V$ and type A, the 
%spectrum of the Hessian contains at most $2+3+1+1=6$ distinct 
%eigenvalues which distribute as follows. $\ft$ contributes 2 
%eigenvalue of multiplicity 1, $\fs$ contributes 2 eigenvalue of 
%multiplicity 1, $\fx$ contributes one eigenvalue of multiplicity 
%$\frac{(k-1)(k-2)}{2}$, and $\fy$ contributes one eigenvalue  of 
%multiplicity $\frac{k(k-3)}{2}$. We stress that the content of 
%the spectrum just described does not depend on $k$ (provided 
%$k\ge4$).

Using Fact \ref{fact: iso}, information can 
immediately be deduced from  Equation \pref{eq: 
	Miso1}. For example, %%WRONG eqn ref.
if $\WW$ is a critical point of isotropy $\Delta S_k$ (a fixed point of the 
$S_k$-action on $M(k,k)$), then
the spectrum of the Hessian contains at most $2+3+1+1=7$ distinct 
eigenvalues which distribute as follows: $\ft$ contributes 2 
eigenvalues of multiplicity 1, $\fs$ contributes $2$ eigenvalues of 
multiplicity $k-1$, $\fx$ contributes one eigenvalue of multiplicity 
$\frac{(k-1)(k-2)}{2}$, and $\fy$ contributes one eigenvalue  of 
multiplicity $\frac{k(k-3)}{2}$. This applies to the global minimum 
$\WW=\VV$ and the spurious minimum of type A. 
%We stress that these statements hold for 
%all $k \ge 4$. 

Next, we would like to compute the actual eigenvalues. We demonstrate the 
method for the single  $\fx$-eigenvalue. Pick a non-zero vector from the 
$\fx$-representation. For example, 
\[
\mathfrak{X}^k = \left[\begin{matrix}
0 & 1 &  \ldots & 1 & - (k-2)\\
-1 & 0  & \ldots & 0 & 1 \\
\cdots  & \cdots& \cdots& \cdots & \cdots\\
-1 & 0   &  \ldots & 0 & 1 \\
(k-2) & -1  &  \ldots & -1 & 0 
\end{matrix}\right],
\]
where rows and columns sum to zero and the only non-zero entries are in rows and columns 1 and $k$. Let $\overbar{\mathfrak{X}^k}\in\real^{k\times k}$
be defined by concatenating the rows of $\mathfrak{X}^k$. 
Since $\fx$ only occurs once in the isotopic decomposition and
$\nabla^2\ploss(\V)$ is $S_k$-equivariant, %%  (the isotropy group of $\V$ is $\Delta S_k$), 
$\overbar{\mathfrak{X}^k}$ must be an eigenvector. In particular,
$(\nabla^2\ploss(\V)\overbar{\mathfrak{X}^k})_i=\lambda_{\mathfrak{x}}\overbar{\mathfrak{X}_i^k}$, all $i \in
\ibr{k^2}$. Choose $i$ so that $\overbar{\mathfrak{X}_i^k} \ne 0$. For 
example, $\overbar{\mathfrak{X}_2^k} = 1$.
Matrix multiplication, yields $\lambda_{\mathfrak{x}} =1/4-1/2\pi$ 
(see \pref{sec: C} for expressions for the Hessian entries). 
% 
%
%
%\[
%\overbar{\mathfrak{X}}^k = \text{vec}\left(\left[\begin{matrix}
%0 & 1 & 1 & \ldots & 1 & - (k-2)\\
%-1 & 0 & 0 & \ldots & 0 & 1 \\
%-1 & 0 & 0  &  \ldots & 0 & 1 \\
%\cdots & \cdots & \cdots& \cdots& \cdots & \cdots\\
%-1 & 0 & 0  &  \ldots & 0 & 1 \\
%(k-2) & -1 & -1  &  \ldots & -1 & 0 
%\end{matrix}\right]\right),
%\]
%
%where $\overbar{\mathfrak{X}}^k$ is the $k^2$-dimensional vector obtained by 
%concatenating the columns of the matrix. Since the factor
%$\fx$ only occurs once in the isotopic decomposition and 
%$\nabla^2\ploss(\V)$ is $S_k$-equivariant, %%  (the isotropy group of $\V$ is $\Delta S_k$), 
%$\overbar{\mathfrak{X}^k}$ must be an eigenvector. That is,   
%$\nabla^2\ploss(\V)\overbar{\mathfrak{X}^k}=\lambda_{\mathfrak{x}}\overbar{\mathfrak{X}^k}$.
%A straightforward multiplication of 
%$\overbar{\mathfrak{X}^k}$ by the Hessian at $\W=\V$ reveals the 
%$\fx$-eigenvalue $\lambda_{\mathfrak{x}} =1/4-1/2\pi$.  

A similar analysis holds for the eigenvalue associated to $\mathfrak{y}$. 
The multiple factors $2\mathfrak{t}$ and $3\mathfrak{s}$ are handled by 
making
judicious choices of orthogonal invariant 
subspaces and representative vectors in 
$M(k,k)$. 
%We 
%give full details later (see \pref{sec: 
%full_iso_decomp}).
A complete derivation of all the 
eigenvalues, including a detailed list of 
the representative vectors and expressions 
for the Hessian of $\ploss$ at $\V$, are 
provided in the appendix.
%A complete derivation of all the 
%eigenvalues, together with the expressions for the 
%Hessian of $\ploss$ at $\V$, are provided in \pref{sec: W equal V}.

%%% I strongly strongly suggest deferring to the details.
%If the factor occurs more than once in the isotopic decomposition (for example, 
%%the factor $3 \mathfrak{s}$),
%then each factor will correspond to a subspace of $M(k,k)$. In this case, say 
%$V_1, V_2$. Pick a 
%non-zero $X_1 \in V_1$, any non-zero $S_k$-map $S: V_1 \arr V_2$, and set $X_2 
%= T(S_1)$. Then if $A: M(k,k)\arr M(k,k)$ is
%an $S_k$-map, $A(X_i) = \sum_{j\in\ibr{2}}a_{ij}X_j$,  $i \in \ibr{2}$ and the 
%eigenvalues of the $2\times 2$ matrix $A=[a_{ij}]$
%give the eigenvalues of $A|2\mathfrak{s}$. In practice, there are natural 
%choice for the representative vectors $X_i$ and we 

	% !TEX root = sym_hess.tex

\section{The Hessian spectrum at spurious minima} 
\label{sec: submax hessian}
Having described the general strategy for analyzing the Hessian spectrum 
for global minima, we now examine the spectrum at various types of 
spurious minima. We need two additional ingredients: a 
specification of the entries of a given family of spurious minima and 
the respective 
isotypic decomposition; we begin with the latter.

As discussed in the introduction, the symmetry-based analysis of the Hessian 
relies on the fact that isotropy groups of spurious minima tend to be (and some 
provably are) maximal subgroups of the target matrix isotropy. 
For $\V=\I$, the relevant maximal isotropy groups are of the form 
$\Delta(S_p\times S_q),~p+q=k$. Below, we provide the corresponding 
isotypic decompositions. Assume $d = k$ and regard $M(k,k)$ as an 
$S_p\times S_q$-space, 
where $S_p \times S_q \subset S_k$ and the (diagonal) action of $S_k$ is 
restricted to the subgroup $S_p\times S_q$. 
\begin{thm}\label{thm: isot2_body}
	The isotypic decomposition of $(M(k,k),S_p\times S_q)$ is given by:
	\begin{enumerate}
		\item If $p= k-1$, $q=1$, and $k \ge 5$,
		\[
		M(k,k) = 5\mathfrak{t} + 5\mathfrak{s}_{k-1} + \mathfrak{x}_{k-1} + \mathfrak{y}_{k-1}.
		\]
		\item If $q \ge 2$, $k-1 > p > p/2$ and $k \ge 4+q$, then
		\[
M(k,k) = 6\mathfrak{t}+6 \mathfrak{s}_p  + a \mathfrak{s}_{q} 
+\mathfrak{x}_p +\mathfrak{y}_p + b\mathfrak{x}_{q} +  
c\mathfrak{y}_{q} + 2\mathfrak{s}_p \boxtimes \mathfrak{s}_{q},
		\]
where if
$q=2$, then $a = 4, b = c = 0$; if  $q=3$, then $a = 5,b = 1, c= 0$; and if $q \ge 4$, then $a = 6, b = c = 1$.
	\end{enumerate}
\end{thm}
\pref{thm: isot2_body} implies that the Hessian 
spectrum of local minima (or critical points) with isotropy $\Delta(S_p\times 
S_q)$ has at most 12 distinct eigenvalues if (1) applies, and if (2) 
holds, at most 19 distinct 
eigenvalues if $q=2$, at most 21 distinct eigenvalues if $q=3$, and 
at most 22 distinct eigenvalues if $q \ge 4$. Moreover, $k^2-O(k)$ of the $k^2$ 
eigenvalues (counting multiplicity) are the $\fx$- and 
$\fy$-eigenvalues. We omit some less interesting cases when $k$ is 
small.

Following the same lines of argument described  in \pref{sec: 
iso_global}, our goal  is to pick a set of non-zero vectors for each 
irreducible representation that will allow us to compute the spectrum. 
While this is simple, estimating the Hessian is not trivial. For this,
we need good estimates on the critical points determining  the 
spurious local minima.

In a recent work~\cite{ArjevaniField2020}, three infinite families of 
critical points were described: type A of isotropy $\Delta S_k$, and 
types I and II of isotropy $\Delta S_{k-1}$.
% These critical points are 
%known to determine spurious minima (for example, if $k = 6$, and by this 
%work, all $k \ge 6$).   
These relatively large isotropy groups made it possible to derive
power series in $1/\sqrt{k}$ for the
critical points and compute the initial terms. Estimates resulting from these series allow us get sharp estimates on the Hessian
which in turn lead to sharp estimates on eigenvalues. 
The derivation is 
lengthy and quite technical and is therefore deferred to the appendix. As an illustration
of the method, 
we sketch the derivation of 
the $\fx$-eigenvalue estimate for the family of type II local minima (case 
1 in \pref{thm: isot2_body}).

Briefly, if $(\mathfrak{c}_k)_{k \ge 3}$ denotes the sequence of type II
critical points of $\cal{F}$, then we may represent 
$\mathfrak{c}_k$ as a point in 
$M(k,k)^{S_{k-1}} = \{\WW \dd g \WW = \WW, g \in 
S_{k-1}\}$---the 5-dimensional fixed point space of the (diagonal) action 
of $ S_{k-1}$ on
$M(k,k)$. If $\mathfrak{c}_k = 
(\xi_1^k,\xi_2^k,\xi_3^k,\xi_4^k,\xi_5^k) \in M(k,k)^{S_{k-1}}$, 
then $\mathfrak{c}_k$ corresponds to
$\WW = [w_{ij}]\in M(k,k)$ where
\[ 
w_{ii} =\begin{cases}
& \xi_1^k,\; i < k\\
& \xi_5^k,\; i=k 
\end{cases},\;\;
w_{ij} = \begin{cases}
&\xi_2^k,\; i,j < k,\; i \ne j\\
&\xi_4^k,\; i < k=j \\
&\xi_3^k,\; j < k = i
\end{cases}.
\]

\begin{lemma}[{\cite[Section 8]{ArjevaniField2020}}]\label{lem: asym}
	(Notation and assumptions as above.)
	For large enough $k$, $\mathfrak{c}_k$ may be written as a 
	convergent power series in $k^{-\frac{1}{2}}$:
	\begin{align*}
	\xi_1^k = 1 + \sum_{\ell =4}^\infty c_\ell& k^{-\ell/2 }
	,\quad \xi_2^k = \sum_{\ell =4}^\infty e_\ell k^{-\ell/2 },\quad 
	\xi_3^k=  \sum_{\ell =2}^\infty f_\ell k^{-\ell/2 },\\
	\xi_4^k &=  \sum_{\ell =2}^\infty g_\ell k^{-\ell/2 }, \quad \xi_5^k= 
	-1+ \sum_{\ell =2}^\infty d_\ell k^{-\ell/2 },
	\end{align*}
	where
	\begin{align*}
	{\arraycolsep=1.4pt\def\arraystretch{2}
	\begin{array}{cccccc}
		c_4 = \frac{8}{\pi},& d_2 = 2 + \frac{8\pi + 
			8}{\pi^2},&
		e_4 = -\frac{4}{\pi},& f_2 = 2,&g_2 = -e_4, \\ 
		c_5 = -\frac{320\pi}{3\pi^4(\pi-2)},& d_3 = 
		\frac{64\pi 
			- 
			768}{3\pi^4(\pi-2)},& e_5 = -\frac{32}{\pi^3},&
			f_3 
		= 0,&
		g_3 = -e_5.
	\end{array}}
%	&c_4 = \frac{8}{\pi},\; d_2 = 2 + \frac{8\pi + 
%	8}{\pi^2},\; 
%	e_4 = -\frac{4}{\pi},\; f_2 = 2,\; g_2 = -e_4, \\ 
%	&c_5 = -\frac{320\pi}{3\pi^4(\pi-2)},\; d_3 = \frac{64\pi 
%	- 
%		768}{3\pi^4(\pi-2)},\; e_5 = -\frac{32}{\pi^3},\; f_3 
%		= 0,\; 
%	g_3 = -e_5 .
	\end{align*}
%
%	\begin{align*}
%	&c_4 = \frac{8}{\pi},\; d_2 = 2 + \frac{8\pi + 
%8}{\pi^2},\; 
%	e_4 = -\frac{4}{\pi},\; f_2 = 2,\; g_2 = -e_4, \\ 
%	&c_5 = -\frac{320\pi}{3\pi^4(\pi-2)},\; d_3 = \frac{64\pi 
%- 
%		768}{3\pi^4(\pi-2)},\; e_5 = -\frac{32}{\pi^3},\; f_3 
%= 0,\; 
%	g_3 = -e_5 .
%	\end{align*}
\end{lemma}

\begin{table*}\centering 
\begin{tabular}{llll}
\toprule
Hessian Entry& Estimate & Hessian Entry&Estimate\\
\midrule 
$H^{11}_{22}$  & $\frac{1}{2}-\frac{1}{\pi k} +  O(k^{-2})$
&$H^{11}_{23}$ &  $O(k^{-2})$\\
$H^{12}_{33}$ & $\frac{1}{4} + O(k^{-2})$
&$H^{12}_{13}$ & $O(k^{-2})$\\
$H^{12}_{12}$  & $\frac{1}{2\pi}+O(k^{-2})$\\
\bottomrule
\end{tabular}
\caption{Estimates of the Hessian entries for type II 
critical points based on the formula provided in \pref{sec: C} and 
\pref{lem: asym} below. 
$H^{pq}$ denotes the $(p,q)$'th $k\times k$ block of the 
$k^2\times k^2$ matrix $\nabla^2 
\ploss(\mathfrak{c}_k)$.}
\label{tab: estimates}
\end{table*}
Proceeding with the lines of argument described  in \pref{sec: 
	iso_global}, we use these power series for $\xi_1,\xi_2, 
	\xi_3,\xi_4,\xi_5$ to derive estimate
for the Hessian entries (see Table \ref{tab: estimates}), which in 
turn  give:
\begin{align*}
\prn*{\nabla^2\ploss(\mathfrak{c}_k)\overbar{\mathfrak{X}^{k-1}}}_2&=
(H^{11}_{22} - H^{11}_{23} - H^{12}_{33} + 2H^{12}_{13} - 
H^{12}_{12})\overbar{\mathfrak{X}^{k-1}}\\
&=\prn*{\frac{1}{4} 
-\frac{1}{2\pi} - \frac{1}{\pi k} + 
O(k^{-2})}\overbar{\mathfrak{X}^{k-1}}_2,
\end{align*}
showing that $\frac{1}{4} -\frac{1}{2\pi} - \frac{1}{\pi k} + O(k^{-2})$ 
is an eigenvalue of $\nabla^2\ploss(\mathfrak{c}_k)$ of multiplicity 
$\frac{(k-2)(k-3)}{2}$ (note that the computation implicitly relies on 
the symmetry of the entries of $\nabla^2\ploss(\mathfrak{c}_k)$). The 
complete derivation of the eigenvalue estimates stated in \pref{thm: 
global}
%for 
%type 
%A, I and II minima 
is provided in Sections A-E.

\section{Conclusion}
We exploit the presence of rich symmetry in ST two-layers ReLU models
to derive an analytic characterization of the Hessian spectrum in the 
natural regime where the number of inputs and hidden neurons is 
finite. This allow us, for the 
first time,
to rigorously confirm (and refute) various hypotheses regarding 
the mysterious generalization 
abilities of neural networks. The methods described in the paper 
apply more broadly 
\cite{arjevani2019spurious}, and yield different spectral 
properties for the Hessian that vary by the choice of the 
underlying distributions, activation functions and 
architectures.
%Adding bias  to the activation or 
%addressing a non-fixed second-layer increases the technical 
%complexity but is quite tractable. 
%The extension of the methods 
%and results to multi-layers is very much a topic of our current 
%research. 
The approach we wish to put forward follow 
in the 
tradition of mathematics and physics in that we start with a 
symmetric model, for which we can prove detailed analytic results, 
and subsequently break symmetry to get insight into the general 
theory (since critical points are non-degenerate, the results we obtain are robust under symmetry breaking 
perturbations of $\VV$~\cite[9.2]{Field2007}; see also \cite[Cor.~1]{safran2017spurious}).

Some of the results derived in this work seem to challenge several 
research directions. Although much effort has been invested in establishing 
conditions under which no spurious minima exist 
\cite{DBLP:journals/corr/abs-2001-00098,DBLP:journals/corr/SoudryC16,DBLP:conf/nips/GeLM16},
we prove the existence of \emph{infinite} families of spurious 
minima for a simple shallow ReLU model. The hope 
for nonconvex optimization landscapes with no spurious minima requires 
therefore further refinement, at least for certain parameter regimes. Secondly, 
as 
demonstrated by type A and type II minima, not all 
local minima are alike. In particular, the hidden mechanism 
under which such spurious minima are alleviated may be somewhat 
different. Lastly, it is the authors' belief that a deep 
understanding of basic models, such as ST models, is a 
prerequisite for any general theory aimed at explaining 
the success of deep learning.

%can be used to study other 
%important phenomena exhibited by modern networks, in a more principled 
%manner.

%Our analysis shows that, although purely 
%synthetic, the ST model shares some important phenomena which have been observed 
%many times for full scale neural networks, such as skewed Hessian spectrum, and 
%
%
%Even more, the analytic tools developed in this work allow us 
%to test various conjectures which have been raised over the years in trying to 
%explain the mysterious generalization abilities of neural networks, some of 
%which already require further refinement in light of our findings. 
%Theoretical physics often proceeds in terms of solvable synthetic models, 
%describe the related line of work on solvable models of simple feed-forward 
%neural networks. highlight a path forward to capture the subtle interplay 
%between the structure of the data, the architecture of the network, and the 
%learning algorithm.
%Lastly, although we consider here a specific case, we believe that 
%the quantitative power of the representation-theory approach in describing the 
%local curvature can be applied more broadly to other highly nonconvex 
%optimization landscapes.

\subsection*{Acknowledgements}
Part of this work was completed while YA was visiting the 
Simons Institute for the Foundations of Deep Learning program. 
We thank Amir Ofer, Itai Safran, Ohad Shamir, Michal Shavit 
and Daniel Soudry for valuable discussions. 
Thanks also to Bob Howlett, University of Sydney, for help with the 
representation theory of $S_n$.

\subsection*{Broader Impact}
To the best of our knowledge, there are no ethical aspects or 
future societal consequences directly involved in our work.

	\bibliography{bib}
	\bibliographystyle{plain}
	\newpage	
	\appendix

\paragraph{A hitchhiker's guide to the appendix.}
The appendix is organized as follows. In \pref{sec: 
	pf_thm_global}, we provide a description of the Hessian spectrum of type I 
spurious minima, as well as the additional eigenvalues which correspond 
to the $d>k$ case. This completes the statement of \pref{thm: global} given in 
the 
main paper. Next, we devote \pref{sec: B} to representation-theoretic 
preliminaries for the group action under consideration. Concretely, we compute 
the relevant isotypic decompositions and list our choice of representative 
vectors. In \pref{sec: C}, we use the symmetry of the Hessian (w.r.t. the group 
action) to simplify and specialize the generic expressions of the Hessian 
entries to the families of spurious minima considered in this paper. Once  
the $\Delta S_k$-case is completed (see \pref{sec: isosk}), we show how 
to fully analyze the Hessian spectrum of global minima in a relatively simple 
way using  symmetry (see \pref{exam: global}). In 
\pref{sec: D}, the long groundwork laid in previous sections is put to use for  
deriving the Hessian spectrum of types A, I and II minima for $k=d$. The 
derivation of the additional $d>k$ case eigenvalues is presented in \pref{sec: 
	E}. In \pref{sec: empirical}, we demonstrate the eigenvalue bulks 
	phenomenon for perturbed minima, as discussed in the follow-up discussion 
	of \pref{thm:info}. We conclude with numerical estimates for the Hessian 
	spectrum which we obtain through \text{LinAlg}, a linear algebra package of 
	Python. The numerical results confirm our analytic characterization of the 
	Hessian spectra.

\section{Type I Hessian spectrum and proof of 
\pref{thm: global}} 
\label{sec: pf_thm_global}

Below, we provide a description of the Hessian spectrum for type I 
spurious minima. This completes the statement of \pref{thm: global} given 
in the main paper. 
\begin{customthm}{2}[Cont.]
Assuming a $k\times k$ orthogonal target matrix $\V$, and $k 
	\ge	6$, $\nabla^2 \ploss$ at type I spurious local minima 
	has 12 distinct strictly positive eigenvalues:
\begin{enumerate}
	\item $\frac{1}{4} - \frac{1}{2\pi} - 
	\frac{1}{\pi \sqrt{k}} + 
	O(k^{-1})$ of multiplicity $\frac{(k-2)(k-3)}{2}$.
	\item $\frac{1}{4} + \frac{1}{2\pi} - 
	\frac{1}{\pi \sqrt{k}} + O(k^{-1})$ of multiplicity 
	$\frac{(k-1)(k-4)}{2}$.
	\item 5 Eigenvalues	of multiplicity $k-2$, of which one grows at a rate 
	of $\frac{k+1}{4}+O(k^{-1})$, and the rest converge to small constants.
	\item 5 Eigenvalues	of multiplicity $1$, of which 2 grow at a rate of 
	$c_3+\frac{k}{4}+o(1)$, one grows at a rate of $c_4+ 
	\frac{k}{2\pi} + o(1),~c_3,c_4>0$,	and the rest converge to 
	small constants.
	%		sum is 
	%		$\Theta\prn*{\frac{1}{4}k^2+\prn*{\frac{3}{4}-\frac{1}{2\pi}}k}$
	%		(in particular, at least one eigenvalue must grow 
	%		linearly with $k$),
	%		\item 5 Eigenvalues	of 	multiplicity 1 whose 
	%		sum is $\Theta\prn*{(\frac{1}{2}+\frac{1}{2\pi})k}$ (in 
	%		particular, at least one eigenvalue must grow linearly 
	%		with $k$).
%	\item The objective value is $\approx 0.3k^{-1} - 
%	O(k^{-2})$ 
%	\cite{ArjevaniField2020}.
\end{enumerate}
\end{customthm}
\renewcommand{\cal}[1]{\mathcal{#1}}
We extend \pref{thm: global} to allow for $d >k$.
Recall that if $d > k$, we append $d-k$ zeros to the end of each 
row of $\VV$ to define $\VV\in M(k,d)$.
We denote the resulting objective function by $\cal{F}_n$, where 
$n = d-k$ and so $\mathcal{F}_0=\cal{F}$, the objective 
function of Theorem~\ref{thm: global}.
%/ and Proposition~\ref{prop: global}.
\begin{thm}\label{thm: spurious2}
	Assume the conditions of Theorem~\ref{thm: global} and let 
	$d > k\ge 6$. Set $n = d-k$.
	The sequence of spurious minima described in \pref{lem: asym}
	uniquely determines a sequence $(\mathfrak{c}^n_k)$ of critical points 
	defining 
	spurious minima for $\cal{F}_n$ which have isotropy $\Delta 
	S_{k-1} \times S_n$ ($S_n$ permutes columns).
	In particular, $\cal{F}_n$ is real analytic at  
	$\mathfrak{c}^n_k$, $k \ge 6$, and
	the spectrum of the Hessian of $\cal{F}_n$ will be the union 
	of the spectrum of the Hessian of $\cal{F}_0$,
	together with 3 strictly positive eigenvalues $\lambda_1, 
	\lambda_2,\lambda_3$ satisfying
	\begin{enumerate}
		\item $\lambda_1$ has multiplicity $n(k-2)$ and 
		$\lambda_1 = \frac{1}{4} + O(k^{-1})$.
		\item $\lambda_2$ has multiplicity $n$ and $\lambda_2 = 
		\frac{1}{4} + O(k^{-\frac{1}{2}})$.
		\item $\lambda_3$ has multiplicity $n$ and $\lambda_3 = 
		\frac{k+1}{4} + O(k^{-1})$.
	\end{enumerate}
%	Conversely, if $(\mathfrak{c}^n_k)$ is 
%a sequence of 
%	critical points defining spurious 
%minima for $\cal{F}_n$ 
%	which have isotropy $\Delta S_{k-1} 
%\times S_n$, 
%	then it arises from the sequence of 
%spurious minima of 
%	isotropy type $S_{k-1}$ described in 
%Theorem~\ref{thm: 
%		spurious}
	For the spurious minima of \pref{lem: asym},
	$\cal{F}_n(\mathfrak{c}^n_k)=(\frac{1}{2} - 
	\frac{2}{\pi^2})k^{-1} + O(k^{-\frac{3}{2}})$. 
\end{thm}
The proof is given in \pref{sec: E}.

%The proof uses a Hessian computation for $\cal{F}_n$~\cite[\S 
%4.2]{safran2017spurious}, together with a simple
%argument based on the isotypic decomposition of the 
%$S_{k-1}$-representation on $\real^k$.  For more details, see 
%the appendix. 
%\begin{rem}
	%(2) Corollary~\ref{cor: stab} extends to Theorem~\ref{thm: 
	%spurious2}: since the new eigenvalues are bounded above 
	%zero, we may perturb $\VV$ within
	%$M(k,d)$.
%\end{rem}

	% !TEX root = sym_hess.tex

\section{The isotypic decomposition of $(M(k,k),G)$}\label{sec: B} 

In this section our aim is give, with minimal prerequisites, the results needed from the representation theory of the symmetric group.
A little background in character theory would be helpful for checking a few statements (for example, showing specific representations
of $S_k$ are irreducible or real)---for this the introductory text~\cite{Thomas2004} would suffice.
The first three or four lectures in~\cite{fulton1991representation} give a 
good, but terse, introduction to the representation theory of
$S_k$. There are many texts covering the general theory, for example~\cite{James1978}, but a lot of work is often required to extract the information
needed here. Moreover, the representation theory of $S_k$ is special because the ground field can be taken to be $\real$ (or the rationals). 
Many introductory texts on representation 
theory work over 
the complex field:  the proofs are often much 
easier but it 
is often awkward to translate to results 
over the real 
field. 

\subsection{The isotypic decomposition}
We begin with a precise version of the orthogonal decomposition described in 
\pref{sec:spec}.
Suppose $V \subset \real^m$  is a linear subspace, with  Euclidean inner product induced from $\real^m$, and
$(V,G)$ is an orthogonal $G$-representation.
\begin{lemma}\label{lem: isow}
The representation $(V,G)$ may be written as an orthogonal direct sum $\bigoplus_{i=1}^m (\oplus_{j=1}^{p_i}V_{ij})$ where
$V_{ij} \subset V$, $(V_{ij},G)$ is irreducible, and $(V_{ij},G)$ is isomorphic to $(V_{\ell k},G)$ iff $i = \ell$, and $j,k \in \ibr{p_i}$.
The subspaces $\oplus_{j=1}^{p_i}V_{ij}$ are unique, $i\in\ibr{m}$.
\end{lemma}
\proof Induction on $n = \text{dim}(V)$. Trivial for $n=1$. Assume proved for all representations of degree less than $n$. 
If $(V,G)$ is of degree $n$ either it is irreducible, and there is nothing to prove, or 
not. If not, there exists a proper $G$-invariant linear subspace $V_1$ of $V$. By the orthogonality of the action,
$V_2 = V_1^\perp$ is $G$-invariant and so $(V,G)$ is the orthogonal direct sum of representations $(V_1,G)$ and $(V_2,G)$. 
Apply the inductive hypothesis to $(V_1,G)$ and $(V_2,G)$. The proof of uniqueness is straightforward and we omit the details. \qed

If $p_i = 1$, for all $i \in m$, the orthogonal decomposition given by the lemma is unique, up to order; otherwise the decomposition is not unique. 
For this reason, Theorem~\ref{thm: isot2_body} was formulated in terms of isomorphism \emph{classes} rather than in terms of specific subspaces.

In spite of the lack of uniqueness of Lemma~\ref{lem: isow}, in some cases there may be \emph{natural} choices of
invariant subspace for the irreducible components. This is exactly the situation for the isotypic decomposition of $(M(k,k),G)$, $G = S_p \times S_{k-p}$,
given in Theorem~\ref{thm: isot2_body}. This naturality allows us to give natural constructions of the matrices $M_i$, $i \in \ibr{m}$, used for
determining the spectrum of $G$-maps $A: M(k,k)\arr M(k,k)$. 
\begin{exam}\label{ex: diags}
The isotypic decomposition for $(M(k,k),S_k)$ is $2 \mathfrak{t}+3 \mathfrak{s} + \mathfrak{x}+\mathfrak{y}$, $k \ge 4$.
The subspace of $M(k,k)$ determined by $2 \mathfrak{t}$ is the set of all $k\times k$ matrices $\mathcal{T}= \{T_{a,b}\dd a,b\in\real\}$
where the diagonal entries of $T_{a,b}$ all equal $a$ and the off-diagonal entries all equal $b$. There are many ways to write $\mathcal{T}$
as an orthogonal direct sum. For example, $\mathcal{T} = T_{1,1}\real \oplus T_{\frac{2}{k},-\frac{1}{k(k-1)}}\real$. However, there
is only one natural way: $\mathcal{T} = T_{1,0}\real  \oplus T_{0,1}\real$. 
Define $\mathfrak{D}^k_1 =  T_{1,0}$, $\mathfrak{D}^k_2 = T_{0,1}$. 
If we take the \emph{standard} realization of $(\mathfrak{t},S_k)$ to be $(\real,S_k)$,
where $S_k$ acts trivially on $\real$, then we have natural $S_k$-maps $\alpha_1,\alpha_2: \real \arr M(k,k)$ defined by $\alpha_i(t) = t\mathfrak{D}^k_i$, $i = 1,2$.
If $A:M(k,k)\arr M(k,k)$ is an $S_k$-map, then $A$ restricts to the $S_k$-map $A_\mathfrak{t}: \mathcal{T} \arr\mathcal{T}$ and $A_\mathfrak{t}$ uniquely determines a 
$2 \times 2$-matrix $[a_{ij}]$ by
$A_\mathfrak{t}(\mathfrak{D}^k_i) = a_{i1}\mathfrak{D}^k_1 + a_{i2}\mathfrak{D}^k_2$, $i = 1,2$. The eigenvalues (and multiplicities in this case)
of $A_\mathfrak{t}:\mathcal{T}\arr\mathcal{T}$ are the same as the eigenvalues of $[a_{ij}]$.
If we choose a different orthogonal decomposition of $\mathcal{T}$, we get a different $2\times 2$-matrix that is similar to $[a_{ij}]$ and so has the same eigenvalues. 
\end{exam}

In the isotypic decompositions of $M(k,k)$ we consider in detail here, only $\mathfrak{t}$ and $\mathfrak{s}$ occur with multiplicity greater than 1 (later we address
the exterior tensor product
representation $2\mathfrak{s}_p \boxtimes \mathfrak{s}_{q}$---but methods are the same).   Before describing how we 
handle the factors $\mathfrak{s}$, we 
need a more explicit description of the representation $(M(k,k),S_k)$.

\subsection{Decomposition of $(M(k,k),S_k)$ into spaces of matrices.}
Assume $k \ge 4 $ in what follows (results are easily obtained if $k \le 3$ but are not interesting for our applications).

Let $\mathbb{D}_k$ denote the space of diagonal $k \times k$-matrices, $\mathbb{A}_k$ the space of skew-symmetric $k \times k$-matrices, and 
$\mathbb{S}_k$ the space of symmetric $k \times k$-matrices with diagonal entries zero. 
We have the orthogonal direct sum decomposition 
\[
M(k,k) = \mathbb{D}_k \oplus \mathbb{A}_k \oplus \mathbb{S}_k
\]
Since $S_k$ acts diagonally on $M(k,k)$, this direct sum is $S_k$-invariant.

Recall that $H_{k-1}\subset \real^k$ is the hyperplane $\sum_{i\in\ibr{k}} x_i = 0$.
In Example~\ref{ex: iso},  we defined $(H_{k-1},S_k)$ and $(T,S_k)$ to be the standard and trivial representations 
of $S_k$. We write here $(\real,S_k)$, rather than $(T,S_k)$, but caution that there is always at least one non-trivial representation of $S_k$ on $\real$. However, these representations
do not not occur here. 
View $(H_{k-1},S_k)$ and $(\real,S_k)$ as standard \emph{models} or \emph{realizations} of the isomorphism classes $\mathfrak{s}_k$ and $\mathfrak{t}$. 
\begin{lemma}\label{lem: isotD}
$\mathbb{D}_{k}$ is the orthogonal $S_k$-invariant direct sum $\mathbb{D}_{k,1} \oplus  \mathbb{D}_{k,2}$,
where 
\begin{enumerate}
\item $\mathbb{D}_{k,1} $ is the space of diagonal matrices with all entries equal and is naturally isomorphic to $(T,S_k)$. 
\item $\mathbb{D}_{k,2}$ is the $(k-1)$-dimensional space of diagonal matrices with diagonal entries summing to zero and is naturally isomorphic to $(H_{k-1},S_k)$.
\end{enumerate}
In particular, the isotypic decomposition of $(\mathbb{D},S_k)$ is $\mathfrak{t}+\mathfrak{s}_k$.
\end{lemma}
\proof For (1), define the $S_k$ map $\real\arr\mathbb{D}_{k,1}$ by $t \mapsto t \mathfrak{D}_1^k$ and for
(2), map $(x_1,\cdots,x_k) \in H_{k-1}$ to the diagonal matrix $D$ with entries $d_{ii} = x_i$, $i \in \ibr{k}$. \qed 

The lemma gives a simple instance of natural choices of subspace in the isotopic decomposition as well as a 
natural choice of matrix $\mathfrak{D}_1^k \in \mathbb{D}_{k,1}$ corresponding to $1 \in T$ (we give a choice of matrix for $\mathbb{D}_{k,2}$ shortly).

Next we extend the previous lemma to $\mathbb{A}_k$ and $\mathbb{S}_k$ and give 
and define \emph{explicit} matrices in the isotypic components.
\begin{lemma}\label{lem: isotA}
$\mathbb{A}_{k}$ is the orthogonal $S_k$-invariant direct sum $\mathbb{A}_{k,1} \oplus  \mathbb{A}_{k,2}$, where
\begin{enumerate}
\item $\mathbb{A}_{k,1}$ is the $(k-1)$-dimensional space of matrices $[a_{ij}]$ for which there exists $(x_1,\cdots,x_k) \in H_{k-1}$ such that
for all $i,j \in \ibr{k}$,  $a_{ij} = x_i - x_j$, 
\item  $\mathbb{A}_{k,2}$ consists of all skew-symmetric matrices with row sums zero.
\end{enumerate}
As representations, $(\mathbb{A}_{k,1},S_k)$ is isomorphic to $(H_{k-1},S_k)$ and $(\mathbb{A}_{k,2},S_k)$ is isomorphic to $(\wedge^2 H_{k-1},S_k)$.
In particular, the isotypic decomposition of $(\mathbb{A}_k,S_k)$ is $\mathfrak{s}_k + \mathfrak{x}_k$.  
\end{lemma}
\proof The isotypic decomposition of $(\mathbb{A}_k,S_k)$ and irreducibility of the 
exterior square representation may be found 
in~\cite{James1978,fulton1991representation}.  Alternatively, use the explicit 
description and character theory to verify 
irreducibility.
\qed
\begin{lemma}\label{lem: isotS}
$\mathbb{S}_k$ is the orthogonal $S_k$-invariant direct sum $\mathbb{S}_{1,k} \oplus  \mathbb{S}_{2,k}\oplus \mathbb{S}_{3,k}$, where
\begin{enumerate}
\item $\mathbb{S}_{1,k}$ is the $1$-dimensional space of symmetric matrices with diagonal entries zero and all off diagonal entries equal.
\item $\mathbb{S}_{2,k}$ is the $(k-1)$-dimensional space of matrices $[a_{ij}]\in \mathbb{S}_k$ for which there exists $(x_1,\cdots,x_k) \in H_{k-1}$ such that
for all $i,j \in \ibr{k}$, $i \ne j$,  $a_{ij} = x_i + x_j$.
\item $\mathbb{S}_{3,k}$ consists of all symmetric matrices in $\mathbb{S}_k$ with all row (equivalently, column) sums zero.
\item $\text{dim}(\mathbb{S}_{3,k}) = \frac{k(k-3)}{2}$.
\end{enumerate}
The representations $(\mathbb{S}_{k,i},S_k)$ are irreducible, $i \in \ibr{3}$:  $(\mathbb{S}_{k,1},S_k)$ is isomorphic to the trivial representation,
$(\mathbb{S}_{k,2})$ is isomorphic to the standard representation and
$(\mathbb{S}_{k,3},S_k)$ is isomorphic to the $S_k$-representation associated to the partition $(k-2,2)$ (isomorphism type $\mathfrak{y}_k$).
\end{lemma}
\proof It is straightforward to check the orthogonality, (1--4) and the $S_k$-invariance of the decomposition.
The isotypic decomposition of $\mathbb{S}_{1,k}\oplus\mathbb{S}_{2,k}$ is $\mathfrak{t}+\mathfrak{s}_k$.
It is known that the isotypic decomposition of $(\mathbb{S}_k,S_k)$ is 
$\mathfrak{t}+\mathfrak{s}_k+\mathfrak{y}_k$~\cite{James1978,fulton1991representation}.
Since we have already identified the factors  $\mathfrak{t},\mathfrak{s}_k$, $(\mathbb{S}_{k,3},S_k)$ has isomorphism type $\mathfrak{y}_k$. 
Alternatively, use the explicit description of $(\mathbb{S}_{k,3},S_k)$ and character theory to verify irreducibility---which is all we need. \qed

\subsection{The general method}
We have now identified three sub-representations in $(M(k,k),S_k)$ that are isomorphic to the standard representation $(H_{k-1},S_k)$.
Moreover lemmas~\ref{lem: isotD}, \ref{lem: isotA}, and \ref{lem: isotS} give explicit parametrizations of the representations in terms of 
the standard representation $(H_{k-1},S_k)$.
Choose a non-zero vector in $H_{k-1}$, for example $(1,-1,0,\cdots,0)$. Denote the corresponding elements in $\mathbb{D}_k$, $\mathbb{A}_{k}$ and $\mathbb{S}_k$ by
$\mathfrak{S}_1^k$, $\mathfrak{S}_2^k$ and $\mathfrak{S}_3^k$ respectively. Then 
$$\mathfrak{S}^k_1 = \left[\begin{matrix}
1 & 0 & \ldots & 0 & 0 \\
0 & -1 & \ldots & 0 & 0 \\
\ldots &\ldots & \ldots & \ldots& \ldots\\
0 & 0   &  \ldots & 0& 0  \\
0 & 0   &  \ldots & 0 &0
\end{matrix}\right],$$ 
\[
\mathfrak{S}^k_2 = \left[\begin{matrix}
0 & 2 & 1 & \ldots & 1 & 1 \\
-2 & 0 & -1&\ldots & -1 & -1 \\
-1 & 1 & 0&\ldots & 0 & 0 \\
\ldots & \ldots &\ldots & \ldots & \ldots& \ldots\\
-1 & 1 & 0   &  \ldots & 0& 0  \\
-1 & 1 & 0  &  \ldots & 0 &0
\end{matrix}\right],\quad
\mathfrak{S}^k_3 = \left[\begin{matrix}
0 & 0 & 1 & \ldots & 1 & 1 \\
0 & 0 & -1&\ldots & -1 & -1 \\
1 & -1 & 0&\ldots & 0 & 0 \\
\ldots & \ldots &\ldots & \ldots & \ldots& \ldots\\
1 & -1 & 0   &  \ldots & 0& 0  \\
1 & -1 & 0  &  \ldots & 0 &0
\end{matrix}\right].
\]
Suppose $A: M(k,k)\arr M(k,k)$ is an $S_k$-map. Set $\mathbb{V} = \mathbb{D}_{k,2}\oplus \mathbb{A}_{k,1}\oplus \mathbb{S}_{k,2}$ so that
$\mathbb{V}$ has isotypic decomposition $3\mathfrak{s}_k$. Setting
$A_\mathfrak{s} = A|\mathbb{V}$, we have $A_\mathfrak{s}:\mathbb{V}\arr\mathbb{V}$. 
Since $\mathfrak{s}_k$ is a real representation, 
\[
A(\mathfrak{S}_i^k) = \sum_{j\in\ibr{3}} a_{ij}\mathfrak{S}_j^k,\; i \in \ibr{3},
\]
where $[a_{ij}]$ is a real $3\times 3$-matrix.
The eigenvalues of the matrix $[a_{ij}]$ give the eigenvalues of $A_\mathfrak{s}:\mathbb{V}\arr\mathbb{V}$ (with multiplicities multiplied by $(k-1)$).

We have shown how to deal with multiple factors of $\mathfrak{t}$ and $\mathfrak{s}$. For the representations $\mathfrak{x}_k$ and $\mathfrak{y}_k$,
we have $A | \mathfrak{x}_k = \lambda_\mathfrak{x}I$, $A | \mathfrak{y}_k = \lambda_\mathfrak{y}I$. It is enough to compute
$A(M)_i$ where $M$ is a non-zero matrix in $\mathbb{A}_{k,2}$ (resp. $\mathbb{S}_{k,3}$) with $M_i \ne 0$ ($M$ in vectorized form so $i\in\ibr{k^2}$). To simplify 
computations, we choose matrices with many zeros
and take 
\[
\mathfrak{X}^k = \left[\begin{matrix}
0 & 1 & 1 & \ldots & 1 & - (k-2)\\
-1 & 0 & 0 & \ldots & 0 & 1 \\
-1 & 0 & 0  &  \ldots & 0 & 1 \\
\cdots & \cdots & \cdots& \cdots& \cdots & \cdots\\
-1 & 0 & 0  &  \ldots & 0 & 1 \\
(k-2) & -1 & -1  &  \ldots & -1 & 0 
\end{matrix}\right] \in \mathbb{A}_{k,2}
\]
and 
\[
\mathfrak{Y}^k = \left[\begin{matrix}
0 & k-3 & 3-k & \ldots & 0 & 0 & 0 \\
k-3 & 0 & 0 & \ldots & -1 & -1&-1\\
3-k & 0 & 0  &  \ldots & 1& 1 & 1 \\
0 & -1 & 1  &  \ldots & 0& 0 & 0 \\
\cdots & \cdots & \cdots & \cdots& \cdots& \cdots\\
0 & -1 & 1  &  \ldots &0 & 0 & 0 \\
0& -1 & 1  &  \ldots& 0& 0 & 0 
\end{matrix}\right] \in \mathbb{S}_{k,3}
\]
\subsection{Isotypic decomposition of 
$(M(k,k),S_p \times S_q)$} \label{sec: 
iso_decomp_Mkk}
Assume $p+q = k$, regard $S_p \times S_q$ as a subgroup of $S_k$ and restrict the diagonal action of $S_k$ on $M(k,k)$ to $S_p \times S_q$ to define
$M(k,k)$ as an $S_p \times S_q$-space.
We assume $k >p > k/2$ so that $S_p \times S_q$ will be a maximal intransitive subgroup of $S_k$~\cite{arjevani2019spurious,ArjevaniField2020}.
Clearly, $M(k,k)$ decomposes as an orthogonal $S_p \times S_q$-invariant direct sum
\[
M(k,k) = M(p,p) \oplus M(p,q) \oplus M(q,p) \oplus M(q,q),  
\]
where $M(p,p)$ is an $S_p$-space and $M(q,q)$ is an $S_q$ space (diagonal actions). We regard $M(p,q)$  and $M(q,p)$ as $S_p \times S_q$-spaces.
Thus, $S_p$ acts on $M(p,q)$ (resp.~$M(q,p)$) by permuting rows (resp.~columns) and $S_q$ acts on $M(p,q)$ (resp.~$M(q,p)$) by permuting columns (resp.~rows).
At first sight this convention may seem confusing but
observe that the map $M(p,q) \arr M(q,p); A \mapsto A^T$, is a linear isomorphism and an $S_p \times S_q$-map. Hence the representations
$(M(p,q),S_p \times S_q)$ and $(M(q,p),S_p \times S_q)$ are isomorphic. 

If $A \in   M(k,k)$, write $A$ in block form as
$
A = \left[ \begin{matrix}

A_{p,p} & A_{p,q} \\
A_{q,p} & A_{q,q}
\end{matrix}\right],
$
where $A_{r,s} \in M(r,s)$, $(r,s) \in \{p,q\}$. 
Certain special block matrices will be needed for the analysis of the eigenvalue structure. 
We make use of the matrices defined in the previous section.
\subsubsection{Block matrix decompositions related to $\mathfrak{x}$}
Define
\[
\mathfrak{X}^{p,p} = \left[\begin{matrix}
\mathfrak{X}^{p} & \is{0}_{p,q}\\
\is{0}_{q,p} &\is{0}_{q,q}  
\end{matrix}\right], \quad
\mathfrak{X}^{q,q} = \left[\begin{matrix}
\is{0}_{q,q} & \is{0}_{p,q}\\
\is{0}_{q,p} &\mathfrak{X}^{q}  
\end{matrix}\right], \quad
\]
where for the definition of $\mathfrak{X}^{q,q}$ it is assumed that $q \ge 3$.
\subsubsection{Block matrix decompositions related to $\mathfrak{y}$}
Define
\[
\mathfrak{y}^{p,p} = \left[\begin{matrix}
\mathfrak{y}^p & \is{0}_{p,q}\\
\is{0}_{q,p} &\is{0}_{q,q}  
\end{matrix}\right], \quad
\mathfrak{y}^{q,q} = \left[\begin{matrix}
\is{0}_{q,q} & \is{0}_{p,q}\\
\is{0}_{q,p} &\mathfrak{y}^q  
\end{matrix}\right], \quad
\]
where for the definition of $\mathfrak{Y}^{q,q}$ it is assumed that $q > 3$.
\subsubsection{Block matrix decompositions related to $\mathfrak{t}$}
Let $\mathcal{I}_{r,s}$ denote the $r \times s$-matrix with all entries equal to $1$. Define
\begin{align*}
&\mathfrak{D}^{p,p}_1 = \left[\begin{matrix}
\mathfrak{D}_1^{p} & \is{0}_{p,q}\\
\is{0}_{q,p} & \is{0}_{q,q} 
\end{matrix}\right],\quad
&\mathfrak{D}^{p,p}_2 = \left[\begin{matrix}
\mathfrak{D}_2^{p} &  \is{0}_{p,q}\\
\is{0}_{q,p} & \is{0}_{q,q} 
\end{matrix}\right], \quad \\
& \mathfrak{D}_1^{q,q} = \left[\begin{matrix}
\is{0}_{p,p}&   \is{0}_{p,q}\\
\is{0}_{q,p} & \mathfrak{D}_1^q 
\end{matrix}\right],\quad 
&\mathfrak{D}^{q,q}_2 = \left[\begin{matrix}
\is{0}_{p,p} &  \is{0}_{p,q}\\
\is{0}_{q,p} & \mathfrak{D}_2^{q} 
\end{matrix}\right],\quad \\
&\mathfrak{D}_3^{p,q} = \left[\begin{matrix}
\is{0}_{p,p}& \mathcal{I}_{p,q}\\
\is{0}_{q,p} & \is{0}_{q,q} 
\end{matrix}\right],\quad
&\mathfrak{D}_3^{q,p} = \left[\begin{matrix}
\is{0}_{p,p}&   \is{0}_{p,q}\\
\mathcal{I}_{q,p} & \is{0}_{q,q} 
\end{matrix}\right],\quad
\end{align*}
where for the definition of $\mathfrak{D}^{q,q}_2$, it is assumed that $q \ge 2$.
\subsubsection{Block matrix decompositions related to $\mathfrak{s}$}
Define $\mathfrak{S}_r^{p,q},\mathfrak{S}_c^{p,q}\in M(p,q)$ by
\[
\mathfrak{S}_r^{p,q}= \left[\begin{matrix}
1 & -1 & 0&\ldots&0  \\ 
1 & -1 & 0&\ldots&0  \\ 
\cdots & \cdots & \cdots &\cdots & \cdots \\
1 & -1 & 0&\ldots&0 \end{matrix}\right],
\mathfrak{S}_c^{p,q}= \left[\begin{matrix}
1 & 1 & 1&\ldots&1  \\ 
-1 & -1 & -1&\ldots&-1  \\ 
\cdots & \cdots & \cdots &\cdots & \cdots \\
0 & 0 & 0&\ldots&0 \end{matrix}\right],
\]
and $\mathfrak{S}_r^{q,p},\mathfrak{S}_c^{q,p}\in M(q,p)$ by
$\mathfrak{S}_r^{q,p} = (\mathfrak{S}_c^{p,q})^T$, $\mathfrak{S}_c^{q,p} = (\mathfrak{S}_r^{p,q})^T$.
Note that $\mathfrak{S}_r^{p,q}$ and $\mathfrak{S}_c^{q,p}$ are only defined if $q \ge 2$.
Set
\begin{align*}
&\mathfrak{S}^{p,p}_1& = \left[\begin{matrix}
\mathfrak{S}^{p}_1 & \is{0}_{p,q} \\
\is{0}_{q,p} & \is{0}_{q,q} 
\end{matrix}\right],\quad
&\mathfrak{S}^{q,q}_1& = \left[\begin{matrix}
\is{0}_{p,p} & \is{0}_{p,q} \\
\is{0}_{q,p} & \mathfrak{S}^{q}_1 
\end{matrix}\right],\quad \\
&\mathfrak{S}^{p,p}_2& = \left[\begin{matrix}
\mathfrak{S}^p_2 & \is{0}_{p,q} \\
\is{0}_{q,p} & \is{0}_{q,q}
\end{matrix}\right],\quad
&\mathfrak{S}^{q,q}_2& = \left[\begin{matrix}
\is{0}_{p,p} & \is{0}_{p,q} \\
\is{0}_{q,p} & \mathfrak{S}^{q}_2 
\end{matrix}\right],\quad \\
&\mathfrak{S}^{p,p}_3& = \left[\begin{matrix}
\mathfrak{S}^p_3 & \is{0}_{p,q} \\
\is{0}_{q,p} & \is{0}_{q,q}
\end{matrix}\right], \quad
&\mathfrak{S}^{q,q}_3& = \left[\begin{matrix}
\is{0}_{p,p} & \is{0}_{p,q} \\
\is{0}_{q,p} & \mathfrak{S}^{q}_3 
\end{matrix}\right],\quad \\
&\mathfrak{S}_4^{p,q}& = \left[\begin{matrix}
\is{0}_{p,p} & \mathfrak{S}_c^{p,q} \\
\is{0}_{q,p} & \is{0}_{q,q}
\end{matrix}\right],\quad 
&\mathfrak{S}_4^{q,p}& = \left[\begin{matrix}
\is{0}_{p,p} &  \mathfrak{S}_r^{p,q} \\
\is{0}_{q,p} & \is{0}_{q,q}
\end{matrix}\right],\quad \\
&\mathfrak{S}_5^{p,q}& = \left[\begin{matrix}
\is{0}_{p,p} & \is{0}_{p,q} \\
\mathfrak{S}_r^{q,p} & \is{0}_{q,q}
\end{matrix}\right],\quad 
&\mathfrak{S}_5^{q,p}& = \left[\begin{matrix}
\is{0}_{p,p} &  \is{0}_{p,q} \\
\mathfrak{S}_c^{q,p} & \is{0}_{q,q}
\end{matrix}\right].\quad
\end{align*}
The first column defines representative elements in each of the factors comprising $5 \mathfrak{s}_p$ that
lie in the image of of $(1,-1,0,\cdots,0) \in H_{p-1}$ by the natural map of $(H_{p-1},S_p)$ onto that factor; similarly
for the second column (with $p$ replaced by $q$).

\subsubsection{Block matrix decompositions related to $\mathfrak{s}_p \boxtimes \mathfrak{s}_q$}
Recall that $\mathfrak{s}_p \boxtimes \mathfrak{s}_q$ is the exterior tensor product of the $S_p$-representation $\mathfrak{s}_p$ and
the  $S_q$-representation $\mathfrak{s}_q$. The degree of $\mathfrak{s}_p \boxtimes \mathfrak{s}_q$ is $(p-1)(q-1)$. Since $p > k/2$, $p \ne q$ and so
$\mathfrak{s}_p \boxtimes \mathfrak{s}_q$ is irreducible.
Just as we view
$(M(p,q),S_p \times S_q)$ and $(M(q,p),S_p \times S_q)$ as isomorphic representations, we regard $\mathfrak{s}_q \boxtimes \mathfrak{s}_p$ as
the isomorphism class of an $S_p \times S_q$ representation and then $\mathfrak{s}_p \boxtimes \mathfrak{s}_q=\mathfrak{s}_q \boxtimes \mathfrak{s}_p$.

Assume $q \ge 2$. Define
\[
\mathfrak{H}_u^{p,q} = 
\left[\begin{matrix}
1 & -1 & 0 &\cdots & 0\\
-1 & 1 & 0 &\cdots & 0\\
0 & 0 & 0 & \cdots & 0\\
\cdots & \cdots & \cdots & \cdots & \cdots \\
0 & 0 & 0 & \cdots & 0
\end{matrix}\right] \in M(p,q),\quad \mathfrak{H}_l^{q,p} = (\mathfrak{H}_u^{p,q})^T \in M(q,p)
\]
Define
\begin{align*}
&\mathfrak{H}^{p,q}& = \left[\begin{matrix}
\is{0}_{p,p} & \mathfrak{H}_u^{p,q} \\
\is{0}_{q,p} & \is{0}_{q,q} 
\end{matrix}\right],\quad
&\mathfrak{H}^{q,p}& = \left[\begin{matrix}
\is{0}_{p,p} & \is{0}_{p,q} \\
\mathfrak{H}_l^{q,p} & \is{0}_{q,q}
\end{matrix}\right],\quad 
\end{align*}
Note that $\mathfrak{H}^{p,q}_1 \in M(p,q)$, $\mathfrak{H}_l^{q,p} \in M(q,p)$.  
The isotopic decomposition of  $(M(p,q),S_p \times S_q)$ (equivalently, $(M(q,p),S_p \times S_q)$ is 
\[
(\mathfrak{s}_p + \mathfrak{t})\boxtimes(\mathfrak{s}_q + \mathfrak{t}))=\mathfrak{s}_p\boxtimes\mathfrak{s}_q+\mathfrak{s}_p+\mathfrak{s}_q+\mathfrak{t}
\]
Hence $M(p,q)\oplus M(q,p)$ contributes $2\mathfrak{s}_p\boxtimes\mathfrak{s}_q+2\mathfrak{s}_p+2\mathfrak{s}_q+2\mathfrak{t}$ to
the isotypic decomposition of $(M(k,k), S_p \times S_q)$.

If $A:M(p,q) \arr M(q,p)$ is an $S_p \times S_q$-map, then $A(\mathfrak{H}_u^{p,q})= c \mathfrak{H}_l^{q,p}$, for some $c\in \real$. 
as an $S_p \times S_q$ representation (we switch the order of the action on the target). In particular, the linear isomorphism $H:M(p,q)\arr M(q,p)$, $A \mapsto A^T$, is
an  $S_p \times S_q$-map and $H(\mathfrak{H}^{p,q}_1) = \mathfrak{H}^{q,p}_1$. In order to compute spectrum associated to $2\mathfrak{s}_p\boxtimes\mathfrak{s}_q$,
we use the representative matrices $\mathfrak{H}^{p,q}, \mathfrak{H}^{q,p}$. Trivial factors add the representative matrices $\mathfrak{D}_3^{p,q}, \mathfrak{D}_3{q,p}$
and  $\mathfrak{s}_p, \mathfrak{s}_q$ add the 4 representative matrices $\mathfrak{S}_i^{r,s}$, where $i = 4,5$ and $r,s \in \{p,q\}$, $r \ne s$.  

All the algebra is now in place for computing the spectra of $S_p \times S_q$-maps of $M(k,k)$, where $p,q,k$ satisfy the conditions of Theorem~\ref{thm: isot2_body}.

	\section{Computation of the Hessian of $\cal{F}$.}\label{sec: C}

We assume that $d = k$ (the case $d>k$ is done is Section~\ref{sec: E}). 
%% I think it should go at the end rather than in this section
%%where it gets lost. But it can be moved to the end of the section if you wish.
We make use of the computations given in~\cite[4.3.1]{safran2017spurious} where the parameters are
viewed as column vectors rather than as row vectors, the natural choice for the matrix formalism. The result, however, is
independent of whatever viewpoint is adopted. Here we represent as columns (labelled by superscripts)
to keep compatibility with notation in~\cite{safran2017spurious}. The result we give applies to the case
$\WW = \VV$ although the Hessian formula~\cite[4.1.1]{safran2017spurious} is
not well-defined when $\WW = \VV$ (division by zero). We remark that $\mathcal{F}(\WW)$ is $C^2$ at $\WW =\VV$, but not real analytic,
or even smooth~\cite{ArjevaniField2020}.

It follows from our analysis that if none of the parameter vectors $\ww^j$ in $\WW$ has isotropy $\Delta S_k$,
then the Hessian depends only on (a) the angles between between parameter vectors and (b) angles between
parameter vectors and the the target parameters $\vv^j$, $j \in \ibr{k}$ determining $\VV$.  In particular, \emph{there is no
dependence on the norms $\|\ww^j\|$}. If $\WW$ has isotropy $\Delta S_{k-1}$, and $\ww^k \ne \is{0}$, then
a similar result holds but now with mild dependence on norms of parameters.
Isotropy groups which are not diagonal
often lead to parallel parameter vectors and loss of differentiability of $\mathcal{F}$ (see~\cite[Ex.~4.9]{ArjevaniField2020}).

Henceforth, we always assume that no rows of $\WW = [\ww^1,\cdots,\ww^k]$ are 
parallel. In particular, that
$\ww^j \ne \is{0}$ and $|\langle \ww^i,\ww^j \rangle| \ne \|\ww^i\|\|\ww^j\|$, $i,j \in \ibr{k}$, $i \ne j$.

\subsection{Formula for the Hessian $\cal{F}$ at $\mathbf{W} 
=[\mathbf{w}^1,\cdots,\mathbf{w}^k]$}

We recall some results and notation from~\cite[4.4.1]{safran2017spurious}. Specifically, for  non-parallel $\ww,\vv \in \real^k$,
let $\theta_{\ww,\vv}\in (0,\pi)$ denote the angle between $\ww, \vv$ and define
\begin{enumerate}
\item $\is{n}_{\ww,\vv} = \frac{\ww}{\|\ww\|} - \cos(\theta_{\ww,\vv})  \frac{\vv}{\|\vv\|}$.
\item $\is{\bar{n}}_{\ww,\vv} = \frac{\is{n}_{\ww,\vv}}{\|\is{n}_{\ww,\vv}\|}$.
\end{enumerate}
Note that
\begin{equation}\label{eq: nwv}
\|\is{n}_{\ww,\vv}\| = \sin(\theta_{\ww,\vv}).
\end{equation}
If $\vv, \ww$ are parallel but not zero, then $\is{n}_{\ww,\vv} = \is{0}$ and we \emph{define} $\is{\bar{n}}_{\ww,\vv} = \is{0}$---this
choice gives the correct value for the Hessian $H$ of $\cal{F}$ if $\WW=\VV$.

We write the Hessian $H$ of $\cal{F}$ as a $k \times k$-matrix of $k \times k$-blocks: $H = [H^{pq}]$. Since $H$ is symmetric,
$H^{pq} = (H^{qp})^T$, $p,q \in \ibr{k}$, and
$H^{pp}$ is symmetric. Each block $H^{pq}$ corresponds
to derivatives with respect to $\ww^p, \ww^q$.

Let $\is{I} = \is{I}_k \in M(k,k)$ denote the identity matrix. Given non-parallel $\ww,\vv\in \real^k$, define $h_1,h_2 \in M(k,k)$ by
\begin{align} \label{eq: h1}
h_1(\ww,\vv) & = \frac{\sin(\theta_{\ww,\vv})\|\vv\|}{2\pi \|\ww\|}\left(\is{I} - \frac{\ww 
\ww^T}{\|\ww\|^2} + \is{\bar{n}}_{\vv,\ww}\is{\bar{n}}_{\vv,\ww}^T\right), \\
\label{eq: h2}
h_2(\ww,\vv)&= \frac{1}{2\pi}\left( (\pi - \theta_{\ww,\vv})\is{I} + 
\frac{\is{\bar{n}}_{\ww,\vv} \vv^T}{\|\vv\|} + \frac{\is{\bar{n}}_{\vv,\ww} 
\ww^T}{\|\ww\|}\right).
\end{align}
%%To avoid discussion of special cases, it is convenient to set $h_1(\ww,\ww) = 0$,  if $\ww \ne \is{0}$.
\begin{lemma}[{\cite[Theorem 5]{safran2017spurious}}]\label{lem: Hess1}
The Hessian $H = [H^{pq}]$ of $\cal{F}$ at the critical point $\WW = [\ww^1,\cdots,\ww^k]$ is
given by
\begin{eqnarray*}
H^{pp}&=&\frac{1}{2} \is{I} + \sum_{q \in \is{k}} \left(h_1(\ww^p,\ww^q) - h_1(\ww^p,\vv^q)\right), \;p\in\ibr{k},\\
H^{pq}&=& h_2(\ww^p,\ww^q), \; p,q \in \ibr{k},\; p \ne q.
\end{eqnarray*}
\end{lemma}

\subsection{Expressions for $h_1, h_2$.}
We work towards obtaining more geometric expressions for the blocks $H^{pq}$. This will involve a careful analysis of the
terms $h_1, h_2$ in the preceeding lemma.
The term $h_1(\ww^p,\vv^p)$, used only in the description of the diagonal blocks, is particularly
tricky as  $\ww^p$ is often close to being parallel to $\vv^p$ in our applications.

Let $\langle \;,\;\rangle$ denote the standard Euclidean inner product on $\real^k$ and {\small$\veebar$} denote
the \emph{exclusive} or.
\begin{lemma}\label{lem :Hess2}
If $q\in \ibr{k}$, $\ww = [w_1,\cdots,w_k]^T\in\real^k$  and $\ww,\vv^q$ are not parallel, then
\begin{eqnarray*}
h_1(\ww,\vv^q)_{ij}&=&\frac{\sin(\alpha_{\ww q})}{2 \pi \|\ww\|} \left(\delta_{ij} -\frac{w_i w_j}{\|\ww\|^2} + K^{\ww q}_{ij}\frac{w_i w_j}{\|\ww\|^2}\right),\; (i,j) \ne (q,q) \\
h_1(\ww,\vv^q)_{qq} &=& \frac{\sin^3(\alpha_{\ww q})}{\pi\|\ww\|},
\end{eqnarray*}
where $\alpha_{\ww q} = \cos^{-1}\left(\frac{\langle \ww, \vv^q\rangle}{\|\ww\|}\right)$ and
\[
K^{\ww q}_{ij} = \begin{cases}
&-1,\; i\vb j= q,\\
& \frac{w^2_q}{\sum_{\ell \ne q} w_\ell^2} = \cot^2(\alpha_{\ww q}),\; i,j \ne q.
\end{cases}
\]
\end{lemma}
\proof The proof is a straightforward computation using~\RRef{eq: h1} and we only give details for $h_1(\ww,\vv^q)_{qq}$.
By~\RRef{eq: h1} and~\RRef{eq: nwv}, we have
\[
h_1(\ww,\vv^q)_{qq} =\frac{\sin(\alpha_{\ww q})}{2 \pi \|\ww\|} \left[ 1 - \frac{w_q^2}{\|\ww\|^2} + \left(1-\cos(\alpha_{\ww q})\frac{w_q}{\|\ww\|}\right)^2\!\big/\sin^2(\alpha_{\ww q})\right].
\]
Since $\cos(\alpha_{\ww q}) = \frac{w_q}{\|\ww\|}$,
$1-\cos(\alpha_{\ww q})\frac{w_q}{\|\ww\|}= 1 - \cos^2(\alpha_{\ww q}) = \sin^2(\alpha_{\ww q})$.
Hence $h_1(\ww,\vv^q)_{qq} = \frac{\sin(\alpha_{\ww q})}{2 \pi \|\ww\|} \left( 1 -  \cos^2(\alpha_{\ww q}) + \sin^2(\alpha_{\ww q})\right)$
giving the result since $1 -  \cos^2(\alpha_{\ww q}) + \sin^2(\alpha_{\ww q}) = 2 \sin^2(\alpha_{\ww q})$. \qed

\begin{rem}
$K^{\ww q}_{ij}$ is well-defined if $i,j \ne q$---since $\ww$ and $\vv^q$ are not parallel. Moreover,
even though $K^{\ww q}_{ij}$ may be large, because of the division by $\sum_{p\ne q} w_p^2$,
$|K^{\ww q}_{ij}w_iw_j|\le \|\ww\|/2$ by the Cauchy-Schwartz inequality.  This allows us to show the formula we
derive below for the Hessian applies when $\WW = \VV$, even though $\ww^i$ is parallel to $\vv^i$, $i \in \ibr{k}$, and that $\mathcal{F}$ is $C^2$ at $\WW=\VV$.
\end{rem}
\begin{lemma}\label{lem :Hess4}
(Notation and assumptions as above.)
For $i,j,p,q \in \ibr{k}$, $ p \ne q$,
\begin{eqnarray*}
h_1(\ww^p,\ww^q)_{ij}& =&\frac{\sin(\Theta_{pq})\|\ww^q\|}{2\pi \|\ww_p\|}\left(\delta_{ij} - \frac{w^p_iw^p_j}{\|\ww^p\|^2}\right) + \\
&&\hspace*{-1.00in} \frac{\|\ww^q\|}{2\pi \|\ww_p\|\sin(\Theta_{pq})}\left(\frac{w_i^qw_j^q}{\|\ww^q\|^2} - \cos(\Theta_{pq})\frac{w_i^pw_j^q+w_i^qw_j^p}{\|\ww^p\|\|\ww^q\|} +
\cos^2(\Theta_{pq})\frac{w_i^pw_j^p}{\|\ww^p\|^2}\right),
\end{eqnarray*}
where  $\Theta_{pq} = \cos^{-1}\left(\frac{\langle \ww^p, \ww^q\rangle}{\|\ww^p\|\|\ww^q\|}\right)$.
\end{lemma}
\proof A straightforward computation using~\RRef{eq: h1}. \qed

\begin{lemma}\label{lem :Hess5}
(Notation and assumptions as above.) For $i,j,p,q \in \ibr{k}$, $p \ne q$,
\[
h_2(\ww^p,\ww^q)_{ij} = \frac{(\pi - \Theta_{pq})\delta_{ij}}{2\pi} +  \frac{1}{2\pi\sin(\Theta_{pq})}L_{ij}^{pq},
\]
where
\[
L_{ij}^{pq} = \frac{w_i^pw_j^q+ 
w_i^qw_j^p}{\|\ww^p\|\|\ww^q\|}-\cos(\Theta_{pq})\left(\frac{w_i^pw_j^p}{\|\ww^p\|^2} + 
\frac{w_i^qw_j^q}{\|\ww^q\|^2}\right).
\]
\end{lemma}
\proof A straightforward computation using~\RRef{eq: h2}. \qed
\begin{rems}
(1) Since no columns of $\WW$ are parallel, division by $\sin(\Theta_{pq})$ is safe in both lemmas.
In our applications, $\sin(\Theta_{pq})$  will typically be close to $1$ for large $k$. \\
(2) If we let $\alpha_{p q}$ denote the angle between $\ww^p$ and $\vv^q$, $p,q \in\ibr{k}$, then all of the terms in  second lemma
can be written in terms of the angles $\alpha_{i q}$
and $\Theta_{pq}$ with no norm terms appearing ($w_q^p/\|\ww^p\|=\cos(\alpha_{p q})$). This is true in the first lemma if all $\ww^j$ have the same norm.
\end{rems}

\subsection{The Hessian of critical points with isotropy 
$\Delta 
S_k$.}\label{sec: isosk}

We give a formula for the Hessian at critical points $\WW = [\ww^1,\cdots,\ww^k]$ with isotropy $\Delta S_k$. We continue to assume $k = d$. Since
isotropy is $S_k$, columns can never be parallel: if two columns are parallel, then since $\WW$ is fixed by $\Delta S_k$, all columns must be equal and so the isotropy of $\WW$
is strictly bigger than $\Delta S_k$.

Since $\WW$ has isotropy $\Delta S_k$, $\|\ww^i\|$ is independent of $i \in \ibr{k}$ and we set $\|\ww^i\| = \tau$, $i\in \ibr{k}$.
Set $w^i_i = R$, $w^j_i = S$, $i,j \in \ibr{k}$, $j \ne i$, so that the diagonal entries of $\WW$ are all equal to $R$, the off-diagonal entries all equal to $S$. Since the
isotropy of $\WW$ is $\Delta S_k$, $S \ne R$.
Define the angles
\begin{enumerate}
\item $\Theta = \cos^{-1}\left(\frac{\langle \ww^i,\ww^j\rangle}{\tau^2}\right)$, $i,j \in \ibr{k}, i \ne j$.
\item $\alpha =  \cos^{-1}\left(\frac{\langle \ww^i,\vv^j\rangle}{\tau}\right)$, $i,j \in \ibr{k}$, $i \ne j$.
\item $\beta =  \cos^{-1}\left(\frac{\langle \ww^i,\vv^i\rangle}{\tau}\right)$, $i\in \ibr{k}$.
\end{enumerate}
and note that $\Theta,\, \alpha$ and $\beta$ are independent of $i,j \in \ibr{k}$.

For $i,j,p \in \ibr{k}$, we tabulate the possible values of $A_{ij}^p \defoo w_i^pw_j^p/\tau^2$.
\begin{align*}
A^p_{ij}& = R^2/\tau^2 = \cos^2(\beta),  \quad  i=j=p\\
A^p_{ij}& = RS/\tau^2 = \cos(\alpha)\cos(\beta),\quad  i \vb j = p\\
A^p_{ij}& = S^2/\tau^2 = \cos^2(\alpha),  \quad  i,j \ne p,
\end{align*}
and define
\begin{align*}
A^{pq}_{ij}& =  \cos^2(\alpha) + \cos^2(\beta),     &B^{pq}_{ij}=&2\cos(\alpha)\cos(\beta),\quad  i,j \in \{p,q\}, i = j\\
      & = 2\cos(\alpha)\cos(\beta), &                 =&\cos^2(\alpha) + \cos^2(\beta) ,\quad  i,j \in \{p,q\}, i \ne j \\
      & = \cos(\alpha)\cos(\beta)+\cos^2(\alpha),  &               =& \cos(\alpha)\cos(\beta)+\cos^2(\alpha) ,\quad  i\vb j \in \{p,q\}\\
      & = 2\cos^2(\alpha)     &              =&2\cos^2(\alpha) ,\quad  i,j \notin \{p,q\}.
\end{align*}
Note that $A^{pq}_{ij}, B^{pq}_{ij}$ are symmetric in $p,q$ and $i,j$.

\begin{lemma}[Off diagonal blocks]\label{lem: nondiagsk}
(Notation and assumptions as above.)
If $p,q,i,j \in \ibr{k}$, with $p \ne q$, then
\[
H^{pq}_{ij}=
\frac{(\pi-\Theta)\delta_{ij}}{2\pi} + \frac{B^{pq}_{ij} - \cos(\Theta)A^{pq}_{ij}}{2\pi\sin(\Theta)}.
\]
In particular $H^{pq} = H^{qp}$ and are symmetric matrices.
\end{lemma}
\proof Immediate from Lemma~\ref{lem :Hess5}, the definitions of $A^p_{ij}, A^{pq}_{ij}$ and $B^{pq}_{ij}$
and the symmetry of $A^{pq}_{ij}, B^{pq}_{ij}$ n $p,q$ and $i,j$. \qed

We need a preliminary result before we give a precise description of $[H^{ij}]$.
\begin{lemma}\label{lem: diagsk}
(Notation and assumptions as above.)
If $p,q,i,j \in \ibr{k}$, $p\ne q$, then
\begin{eqnarray*}
h_1(\ww^p,\ww^q)_{ij}&=&
\frac{\sin(\Theta)}{2\pi}\left(\delta_{ij} - \frac{w_i^pw_j^p}{\tau^2}\right) + \\
&& \frac{1}{2\pi\tau^2\sin(\Theta)}\left(w_i^qw_j^q - \cos(\Theta)(w_i^pw_j^q+w_i^qw_j^p) + \cos^2(\Theta) w_i^pw_j^p\right) \\
&=&\frac{\sin(\Theta)}{2\pi}\left(\delta_{ij} - A^p_{ij}\right) + 
\frac{1}{2\pi\sin(\Theta)}\left(A_{ij}^q - \cos(\Theta)B^{pq}_{ij} + \cos^2(\Theta) A_{ij}^p\right)
\end{eqnarray*}
Let $p,q,i,j \in \ibr{k}$. If $K^{pq}_{ij}\defoo K^{\ww^p q}_{ij}$, then
\begin{enumerate}
\item $p \ne q$
\begin{eqnarray*}
h_1(\ww^p,\vv^q)_{ij} & = & \frac{\sin(\alpha)}{2\pi\tau}\left(\delta_{ij} - A_{ij}^p+K^{pq}_{ij}A_{ij}^p\right) ,\; (i,j)\ne (q,q)\\
&= & \frac{\sin^3(\alpha)}{\pi\tau},\; i=j=q,
\end{eqnarray*}
where if $i,j \ne q$,
\[
K^{pq}_{ij}A_{ij}^p = \begin{cases}
&\cos^2(\beta)\cot^2(\alpha),\; i = j = p\\
& -\cos(\alpha)\cos(\beta),\; i,j \in \{p,q\},\, i \ne j \\
& \cos(\alpha)\cos(\beta)\cot^2(\alpha),\;i \vb j = p,\, i,j\ne q   \\
& -\cos^2(\alpha),\; i \vb j = q,\, i,j\ne p   \\
& \cos^2(\alpha)\cot^2(\alpha), \;i,j \notin \{p,q\}.   
\end{cases}
\]
\item $p = q$
\begin{eqnarray*}
h_1(\ww^q,\vv^q)_{ij} & = & \frac{\sin(\beta)}{2\pi\tau}\left(\delta_{ij} - A_{ij}^q+K^{qq}_{ij}A_{ij}^q\right) ,\; (i,j)\ne (q,q)\\
&= & \frac{\sin^3(\beta)}{\pi\tau},\; i=j=q,
\end{eqnarray*}
where if $i,j \ne q$,
\[
K^{qq}_{ij}A_{ij}^q = \begin{cases}
& \frac{\cos^2(\beta)}{k-1},\; i,j \ne q.\\
&-\cos(\alpha)\cos(\beta),\; i = q \ne j,\; j = q \ne i
\end{cases}
\]
\end{enumerate}
\end{lemma}
\proof We verify the statements concerning the terms $K^{qq}_{ij}A_{ij}^q$. First note that both
$K^{qq}_{ij}$ and $A_{ij}^q$ are symmetric in $i,j$.
If $i,j \ne q$, then $K_{ij}^{qq} = \frac{(w_q^q)^2}{\sum_{\ell \ne q} (w^q_\ell)^2} = \frac{R^2}{(k-1)S^2}$, and
$A_{ij}^q = \cos^2(\alpha) = \frac{S^2}{\tau^2}$. Hence $K_{ij}^{qq}A_{ij}^q =\frac{R^2}{\tau^2(k-1)}=\frac{\cos^2(\beta)}{k-1}$.
On the other hand if $i = q, j \ne q$, $K_{qj}^{qq} = -1$ and
$A_{qj}^q = \cos(\alpha)\cos(\beta)$. Hence $K^{qq}_{qj}A_{qj} = -\cos(\alpha)\cos(\beta)$.  \qed

\begin{prop}[Diagonal blocks]\label{prop: diagsk}
(Notation and assumptions as above.) Let $i,j,p \in \ibr{k}$.
\begin{enumerate}
\item
\begin{eqnarray*}
H^{pp}_{pp}&=&\frac{1}{2}+\frac{(k-1)\sin^2(\beta)}{2\pi}\left(\sin(\Theta)-\frac{\sin(\alpha)}{\tau}\right) - \frac{\sin^3(\beta)}{\pi\tau} + \\
&&\hspace*{-0.5in} \frac{(k-1)}{2\pi}\left(\frac{1}{\sin(\Theta)}\big(\cos(\alpha)-\cos(\Theta)\cos(\beta)\big)^2 - \frac{\cot(\alpha)\cos(\alpha)\cos^2(\beta)}{\tau}\right).
\end{eqnarray*}
\item  If $i \ne p$, then
\begin{eqnarray*}
H^{pp}_{ii}&=&\frac{1}{2}+\frac{(k-2)\sin^2(\alpha)}{2\pi}\left(\sin(\Theta)-\frac{\sin(\alpha)}{\tau}\right) + \\
&&  \frac{\sin^2(\alpha)}{\pi} \left(\frac{\sin(\Theta)}{2}- \frac{\sin(\alpha)}{\tau}\right) + 
\frac{(k-2)}{2\pi}\left(\frac{\cos^2(\alpha)}{\sin(\Theta)}(1-\cos(\Theta))^2 - \frac{\cot(\alpha)\cos^3(\alpha)}{\tau}\right) + \\
&&\frac{1}{2\pi \sin(\Theta)}(\cos(\beta) - \cos(\Theta)\cos(\alpha))^2 - \frac{\sin(\beta)}{2\pi\tau}\left(\sin^2(\alpha) + \frac{\cos^2(\beta)}{k-1}\right)
\end{eqnarray*}
\end{enumerate}
\begin{enumerate}
\item If $i \vb j = p $,  then
\begin{eqnarray*}
H^{pp}_{1j}& = & -\frac{(k-1)\cos(\alpha)\cos(\beta)}{2\pi}\left(\sin(\Theta)-\frac{\sin(\alpha)}{\tau}\right) + \\
&& \frac{(k-2)\cos(\alpha)}{2\pi}\left( \frac{\cos(\alpha) - \cos(\Theta)(\cos(\alpha)+\cos(\beta))+\cos^2(\Theta)\cos(\beta)}{\sin(\Theta)}\right) -\\
&& \frac{(k-2)}{2\pi\tau} \sin(\alpha)\cos(\alpha)\cos(\beta)\cot^2(\alpha) + \frac{\sin(\alpha)\cos(\alpha)\cos(\beta)}{2\pi\tau} + \\
&& \frac{1}{2\pi\sin(\Theta)}\big(\cos(\alpha)\cos(\beta) - \cos(\Theta)(\cos^2(\alpha)+\cos^2(\beta))+  \cos^2(\Theta)\cos(\alpha)\cos(\beta) \big) + \\
&&\frac{\cos(\alpha)\sin(\beta)\cos(\beta)}{\pi\tau}
\end{eqnarray*}
\item If $i,j \ne p$, then
\begin{eqnarray*}
H^{pp}_{ij}& = &-\frac{(k-1)\cos^2(\alpha)}{2\pi}\left(\sin(\Theta)-\frac{\sin(\alpha)}{\tau}\right) + \\
&& \frac{(k-3)\cos^2(\alpha)}{2\pi}\left(\frac{(1-\cos(\Theta))^2}{\sin(\Theta)} - \frac{\sin(\alpha)\cot^2(\alpha)}{\tau}\right) + \\
&& \frac{\cos(\alpha)}{\pi}\left(\frac{\cos(\beta) - \cos(\Theta)(\cos(\beta)+\cos(\alpha))+\cos^2(\Theta)\cos(\alpha)}{\sin(\Theta)}+ \frac{\sin(\alpha)\cos(\alpha)}{\tau}\right)+\\
&& \frac{\sin(\beta)}{2\pi\tau}\left(\cos^2(\alpha) -\frac{\cos^2(\beta)}{k-1}\right) 
\end{eqnarray*}
\end{enumerate}
\end{prop}
\begin{prop}[Off diagonal blocks]\label{prop: nondiagsk}
(Notation and assumptions as above.) Given $i,j,p,q \in \ibr{k}$, $p \ne q$.
\begin{enumerate}
\item $i \in \{p,q\}$,
\begin{eqnarray*}
H^{pq}_{ii}&=& \frac{\pi-\Theta}{2\pi} + \frac{1}{2\pi\sin(\Theta)}\big(2 \cos(\alpha)\cos(\beta) -\cos(\Theta)(\cos^2(\alpha)+\cos^2(\beta)\big)
\end{eqnarray*}
\item $i \notin \{p,q\}$,
\begin{eqnarray*}
H^{pq}_{ii}&=& \frac{\pi-\Theta}{2\pi} + \frac{\cos^2(\alpha)}{\pi\sin(\Theta)}\big(1-\cos(\Theta)\big)
\end{eqnarray*}
\item $i,j \in \{p,q\}$, $i \ne j$,
\begin{eqnarray*}
H^{pq}_{ij}&=& \frac{1}{2\pi\sin(\Theta)}\big(\cos^2(\alpha)+\cos^2(\beta)  -2\cos(\Theta)\cos(\alpha)\cos(\beta)\big)
\end{eqnarray*}
\item $i \vb j \in \{p,q\}$,
\begin{eqnarray*}
H^{pq}_{ij}&=&  \frac{\cos(\alpha)(\cos(\alpha)+\cos(\beta))}{2\pi\sin(\Theta)}\big(1-\cos(\Theta)\big)
\end{eqnarray*}
\item $i, j \notin \{p,q\}$,
\begin{eqnarray*}
H^{pq}_{ij}&=&  \frac{\cos^2(\alpha)}{\pi\sin(\Theta)}\big(1-\cos(\Theta)\big) \
\end{eqnarray*}
\end{enumerate}
\end{prop}
\proof Both results follow straightforwardly from the definitions of $A_{ij}^p,B_{ij}^{pq}$ and Lemmas~\ref{lem: nondiagsk}, \ref{lem: diagsk}. \qed

\begin{exam}[Spectrum of the Hessian at $\WW = \VV$]   \label{exam: global}  
For $i,j \in \ibr{k}$, let $\delta_{ij}$ be equal to $1$ if $i = j$,
and $0$ otherwise, and $\boldsymbol{\delta}_{ij} \in M(k,k)$ be the matrix with $i,j$ entry equal to $1$ and all other entries zero.
If $\WW = \VV$, take $\alpha = \Theta = \pi/2$ and $\beta = 0$ in Propositions~\ref{prop: diagsk}, ~\ref{prop: nondiagsk}.
The Hessian $H = [H^{pq}]$ of $\mathcal{F}$ at $\VV$ is then given by
\begin{eqnarray*}
H^{pp}&=&\frac{1}{2} \is{I},\; p\in \ibr{k},\\
H^{pq}&=& \frac{1}{4} \left(\is{I} + \frac{2}{\pi}(\boldsymbol{\delta}_{pq} + \boldsymbol{\delta}_{qp})\right), \; p\ne q.
\end{eqnarray*}
In particular
\begin{eqnarray*}
H^{pp}_{ij}&=&\begin{cases}
&\frac{1}{2},\; i = j\\
& 0, \; i \ne j
\end{cases}, \qquad p\in \ibr{k},\\
H^{pq}_{ij}&=&\begin{cases}
&\frac{1}{4},\; i = j\\
& \frac{1}{2\pi}, \; i,j\in \{p,q\},\; i \ne j,\\
& 0,\; \text{otherwise}
\end{cases} \qquad p,q\in\ibr{k}, p \ne q
\end{eqnarray*}
Let  $\is{r}_i$ denote row $i$ of $H$, $ i \in \ibr{3}$.

\subsubsection{The eigenvalue associated to $\mathfrak{x}_k$.}
Assume $k > 3$ and take $\mathfrak{X}^k \in \mathbb{A}_{k,2} \subset M(k,k)$ with vectorization $\overline{\mathfrak{X}^k}$ (by rows) as defined previously.
Computing $\langle \is{r}_2,\overline{\mathfrak{X}^k}\rangle$, we find that
\[
H(\mathfrak{X}^k)_{12} = \langle \is{r}_2,\overline{\mathfrak{X}^k}\rangle = \frac{1}{4} - \frac{1}{2\pi}.
\]
Since $\mathfrak{X}_{12} = 1$,  the eigenvalue $\lambda_{\mathfrak{x}}$ associated to $\mathfrak{x}_k$ is $\frac{1}{4}-\frac{1}{2\pi}$,
and has multiplicity $(k-1)(k-2)/2$.
\subsubsection{The eigenvalue associated to $\mathfrak{y}_k$.}
Using the same method as above, $\lambda_{\mathfrak{y}} = \frac{1}{4} + \frac{1}{2\pi}$, for all $k \ge 5$.
\subsubsection{The eigenvalues associated to $\mathfrak{t}$.}
In this case, we compute $\langle \is{r}_i,\mathfrak{D}^k_j\rangle$, for $i,j \in \is{2}$ (see Example\ref{ex: diags} for $\mathfrak{D}_j^k$), to find the matrix
$
\left[\begin{matrix}
\frac{1}{2} + \frac{k-1}{2\pi} & \frac{k-1}{4} \\
\frac{1}{4} & \frac{1}{2\pi} + \frac{k}{4}
\end{matrix}\right]
$
giving the eigenvalues $\lambda^1_{\mathfrak{t}}, \lambda^2_{\mathfrak{t}}$ associated to the factor $2 \mathfrak{t}$. For $k = 6$,
we find $\lambda^1_{\mathfrak{t}} = 0.8896627389$, $\lambda^2_{\mathfrak{t}} =  2.0652669197$. As functions of $k$,  $\lambda^i_{\mathfrak{t}}$
monotonically increase like $c_i k$, where $c_1 \approx 0.16$ and $c_2 = 0.25$.

\subsubsection{The eigenvalues associated to $\mathfrak{s}_k$.}
Denote the matrix associated to the factor $3\mathfrak{s}_k$ by $B = [\beta_{ij}]\in M(3,3)$.  That is,
\[
H(\mathfrak{S}_i) = \beta_{i1}\mathfrak{S}_1+ \beta_{i2}\mathfrak{S}_2+\beta_{i3}\mathfrak{S}_3, \; i \in \is{3}.
\]
Since $H(\mathfrak{S}_i)_j = \langle \is{r}_j,\mathfrak{S}_i\rangle$, it follows that for $i\in \is{3}$,
\[ 
\langle \is{r}_1,\mathfrak{S}_i\rangle  =  \beta_{i1},\quad
\langle \is{r}_2,\mathfrak{S}_i\rangle  =  2\beta_{i2},\quad
\langle \is{r}_3,\mathfrak{S}_i\rangle  =  \beta_{i2} + \beta_{i3}
\]
where the factor $2$ in the second equation occurs since
the $12$-component of $\mathfrak{S}_2$ is $2$.
Setting $h_{ij} = \langle \is{r}_j,\mathfrak{S}_i\rangle$,
\[
[h_{ij}] = \left[\begin{matrix}
\frac{1}{2}-\frac{1}{2\pi}&-\frac{1}{4} &0 \\
-\frac{k}{4}&\frac{k+2}{4} -\frac{1}{\pi} &\frac{1}{4}-\frac{1}{2\pi} \\
\frac{k-2}{4}&-\frac{k-2}{4} & \frac{1}{4}+\frac{1}{2\pi}
\end{matrix}\right],
\]
and so
\[
B = \left[\begin{matrix}
\frac{1}{2}-\frac{1}{2\pi} & -\frac{1}{8} & \frac{1}{8} \\
-\frac{k}{4}&\frac{k+2}{8}-\frac{1}{2\pi} &  -\frac{k}{8}\\
\frac{k-2}{4}& -\frac{k-2}{8} & \frac{k}{8} + \frac{1}{2\pi}
\end{matrix}\right].
\]
Since this equation has real roots for all $k \ge 5$, we can solve in terms of 
trigonometric functions using the formula of François Vi\'ete.
From this we find that for $k=6$, the eigenvalues are
\[
\lambda_{\mathfrak{s}}^1 = 1.712918525755,\;\lambda_{\mathfrak{s}}^2 = 0.287081474245,\; \lambda_{\mathfrak{s}}^3=0.090845056908.
\]
Numerical examination of eigenvalues for different values of $k$ reveals that the last eigenvalue is constant and equal to
$\frac{1}{4}- \frac{1}{2\pi}$---the same as $\lambda_{\mathfrak{x}}$. It may be shown that the
characteristic equation of $B$ has the factorization
\[
\left(\lambda - \frac{1}{4} + \frac{1}{2\pi}\right)\left(\lambda^2 -(\frac{k}{4}+ 
\frac{1}{2})\lambda + \frac{1}{16}(k - 4 \pi^2 + 4 \pi + 1)\right).
\]
Analysis of the roots of the quadratic term reveal that $\lambda_{\mathfrak{s}}^1 = \frac{k+1}{4} + O(k^{-1})$, and
$\lambda_{\mathfrak{s}}^2 = \frac{1}{4}+O(k^{-1})$ is monotone decreasing with limit $0.25$.
In particular, the eigenvalues of the Hessian are uniformly bounded above zero.
\end{exam}
\subsection{The Hessian at critical points with isotropy $S_{k-1}$.}
We assume  $\WW = [\ww^1,\cdots,\ww^k]$ is a critical point of $\mathcal{F}$ with isotropy $S_{k-1}$ and that $\ww_k \ne \is{0}$.
The isotropy $S_{k-1}$ then guarantees that no two columns of $\WW$ are parallel. We give an angle representation of the Hessian at $\WW$.
In this case, we need $7$ angles which we describe below.
Since $\WW$ has isotropy $S_{k-1}$, $\|\ww^i\|$ is independent of $i \in \ibr{k-1}$. Set $\|\ww^i\| = \tau $, $i< k$, and
$\|\ww^k\| = \tau_k$.

Define the angles
\begin{enumerate}
\item $\Theta = \cos^{-1}\left(\frac{\langle \ww^i,\ww^j\rangle}{\tau^2}\right)$, $i,j < k$, $i \ne j$.
\item $\Lambda = \cos^{-1}\left(\frac{\langle \ww^i,\ww^k\rangle}{\tau\tau_k}\right)$, $i < k$.
\item $\alpha_{ii} =  \cos^{-1}\left(\frac{\langle \ww^i,\vv^i\rangle}{\tau}\right)=\cos^{-1}\left(\frac{\ww_i^i}{\tau}\right)$, $i<k$.
\item $\alpha_{ij} =  \cos^{-1}\left(\frac{\langle \ww^i,\vv^j\rangle}{\tau}\right)=\cos^{-1}\left(\frac{\ww^i_j}{\tau}\right)$, $i,j<k$, $i \ne j$.
\item $\alpha_{ik} =  \cos^{-1}\left(\frac{\langle \ww^i,\vv^k\rangle}{\tau}\right)=\cos^{-1}\left(\frac{\ww^i_k}{\tau}\right)$, $i<k$.
\item $\alpha_{kk} =  \cos^{-1}\left(\frac{\langle \ww^k,\vv^k\rangle}{\tau_k}\right)=\cos^{-1}\left(\frac{\ww^k_k}{\tau_k}\right)$.
\item $\alpha_{kj} =  \cos^{-1}\left(\frac{\langle \ww^k,\vv^j\rangle}{\tau_k}\right)=\cos^{-1}\left(\frac{\ww_j^k}{\tau_k}\right)$, $j < k$.
\end{enumerate}
So as to simplify and shorten some of the expressions involved in the description of the Hessian, we set
\begin{enumerate}
\item $\cos(\alpha_{ij}) = \cs_{ij}$, $\cos(\alpha_{ik}) = \cs_{ik}$, and $\cos(\alpha_{ii}) = \cs_{ii}$, $i,j < k$.
\item $\cos(\alpha_{kj}) = \cs_{kj}$ and $\cos(\alpha_{kk}) = \cs_{kk}$.
\item $\cot(\alpha_{ij}) =\ct_{ij}$, $\cot(\alpha_{ij}) = \ct_{ik}$, and $\cot(\alpha_{kj}) =\ct_{kj}$
\item $\sin(\alpha_{ij}) =\sn_{ij}$, $\sin(\alpha_{ik}) = \sn_{ik}$ and $\sin(\alpha_{ii}) = \sn_{ii}$
\item $\sin(\alpha_{kj}) = \sn_{kj}$ and $\sin(\alpha_{kk}) = \sn_{kk}$.
\end{enumerate}

Along similar lines to the previous section, we define and tabulate the values of $A_{ij}^p$, $A_{ij}^{pq}$ and $B_{ij}^{pq}$, $i,j,p\in\ibr{k}$.
For $i,j < k$, define $\rho = w_{ii}^i$, $\eps = w_{ij}^i$, $\zeta = w_{ik}^i$, $\eta = w_{kj}^i$ and  $\nu = w_{kk}^k$. Note that by $S_{k-1}$ symmetry,
$\rho, \eps,\zeta, \eta$ do not depend on the choice of $i,j \in \ibr{k-1}$.
\begin{enumerate}
\item $p < k$.
\begin{align*} 
A^p_{ij}& = \rho^2/\tau^2 = \cs^2_{ii},  \quad  i=j=p\\
A^p_{ij}& = \rho\eps/\tau^2 = \cs_{ii}\cs_{ij},\quad  i,j < k,\; i \vb j = p\\
A^p_{ij}& = \rho\zeta/\tau^2 = \cs_{ii}\cs_{ik},\quad  i \vb j = k,\; i \vb j = p\\
A^p_{ij}& = \eps^2/\tau^2 = \cs^2_{ij},  \quad i,j < k,\;i,j \ne p \\
A^p_{ij}& = \eps\zeta/\tau^2 = \cs_{ij}\cs_{ik},  \quad i \vb j = k,\; i,j \ne p \\
A^p_{ij}& = \zeta^2/\tau^2 = \cs^2_{ik},  \quad i = j = k
\end{align*}
\item $p = k$.
\begin{align*}
A^k_{ij}& = \nu^2/\tau_k^2 = \cs^2_{kk},  \quad  i=j=k\\
A^k_{ij}& = \nu\eta/\tau_k^2 = \cs_{kk}\cs_{kj},\quad i \vb j = k\\
A^k_{ij}& = \eta^2/\tau_k^2 = \cs^2_{kj},  \quad i,j < k
\end{align*}
\end{enumerate}
\begin{enumerate}
\item If $p,q,i,j < k$, $ p \ne q$, define
\begin{align*}
A^{pq}_{ij}& =  \cs^2_{ii} + \cs^2_{ij},     &B^{pq}_{ij}=&2\cs_{ii}\cs_{ij},\quad  i=j \in \{p,q\}\\
      & = 2\cs_{ii}\cs_{ij}, &                 =&\cs^2_{ii} + \cs^2_{ij} ,\quad  i,j \in \{p,q\}, i \ne j \\
      & = \cs_{ij}\cs_{ii}+\cs^2_{ij},  &               =& \cs_{ij}\cs_{ii}+\cs^2_{ij} ,\quad  i\vb j \in \{p,q\}\\
      & = 2\cs^2_{ij}     &              =&2\cs^2_{ij} ,\quad  i,j \notin \{p,q\}
\end{align*}
\item If $p,q< k$, $p \ne q$, $i \vb j = k$, define
\begin{align*}
A^{pq}_{ij}      & = \cs_{ik}(\cs_{ii} +\cs_{ij})     & B^{pq}_{ij}             =&\cs_{ik}(\cs_{ii} +\cs_{ij}) ,\quad  i\vb j \in \{p,q\}\\
      & = 2\cs_{ik}\cs_{ij}     &              =&2\cs_{ik}\cs_{ij},\quad  i, j \notin \{p,q\} 
\end{align*}
\item $p,q< k$, $p \ne q$, $i=j=k$, define
\begin{align*}
    A^{pq}_{kk}  & = 2\cs^2_{ik}     & B^{pq}_{kk}             =&2\cs^2_{ik} 
\end{align*}
\item
If $p \vb q = k$, and $i,j < k$ define
\begin{align*}
A^{pq}_{ij}& =  \cs^2_{kj} + \cs^2_{ii},     &B^{pq}_{ij}=&2\cs_{kj}\cs_{ii},\quad  i=j \in \{p,q\}\\
      & = \cs^2_{kj}+\cs_{ij}\cs_{ii} &                 =&\cs_{kj}(\cs_{ij} + \cs_{ii}) ,\quad  i\vb j \in \{p,q\} \\
      & = \cs^2_{kj}+\cs^2_{ij},  &               =& 2\cs_{kj}\cs_{ij} ,\quad  i, j \notin \{p,q\}
\end{align*}
\item
If $p \vb q = k$, and $i\vb j = k$ define
\begin{align*}
\hspace*{-0.3in} A^{pq}_{ij}& =  \cs_{kk}\cs_{kj} + \cs_{ik}\cs_{ii},     &B^{pq}_{ij}=&\cs_{kk}\cs_{ii}+\cs_{kj}\cs_{ik},\;  i,j \in \{p,q\}\\
\hspace*{-0.3in}       & =    \cs_{kk}\cs_{kj} + \cs_{ik}\cs_{ij} &                 =&\cs_{kk}\cs_{ij}+\cs_{kj} \cs_{ik},\;  i\vb j \in \{p,q\}
\end{align*}
\item If $p \vb q = k$, and $i=j = k$ define
\begin{align*}
A^{pq}_{kk}      & = \cs^2_{kk}+\cs^2_{ik},  &   B^{pq}_{kk}            =& 2\cs_{kk}\cs_{ik} 
\end{align*}
\end{enumerate}

Note that $A^{pq}_{ij}, B^{pq}_{ij}$ are symmetric in $p,q$ and $i,j$.

\begin{prop}[Off diagonal blocks]
(Notation and assumptions as above.)
If $p,q,i,j \in \ibr{k}$, with $p \ne q$. then
\begin{enumerate}
\item If $ p, q < k$,
\begin{eqnarray*}
H^{pq}_{ij} &=& \frac{(\pi-\Theta)\delta_{ij}}{2\pi} + \frac{B^{pq}_{ij} - \cos(\Theta)A^{pq}_{ij}}{2\pi\sin(\Theta)}.
\end{eqnarray*}
\item If $p\vb q=k$,
\begin{eqnarray*}
H^{pq}_{ij} &=& \frac{(\pi-\Lambda)\delta_{ij}}{2\pi} + \frac{B^{pq}_{ij} - \cos(\Lambda)A^{pq}_{ij}}{2\pi\sin(\Lambda)}.
\end{eqnarray*}
\end{enumerate}
In particular,  $H^{pq} = H^{qp}$ and are symmetric matrices.
\end{prop}
\proof Along exactly the same lines as that of Lemma~\ref{lem: nondiagsk}. \qed

Before giving the main lemma for computation of the terms $h_1(\ww^p,\ww^q)$, and $h_1(\ww^p,\vv^q)$, we need to
extend the definition of $K^{pq}_{ij}$ to allow for $S_{k-1}$ symmetry.

Given $i,j,p,q \in \ibr{k}$, $ p \ne q$, define for $(i,j) \ne (q,q)$,
\[
K^{pq}_{ij} = \begin{cases}
&-1, i \vb j = q\\
& \ct^2_{ij}, p \ne q,\; p,q < k,\; i,j \ne q\\
& \ct^2_{ik}, q=k,\; p < k,\; i,j \ne q\\
& \ct^2_{kj}, p=k,\; q < k,\; i,j \ne q
\end{cases}
\]
In case $p = q$, it is more convenient to give the values of $K^{qq}_{ij}A_{ij}^q$ rather than $K^{qq}_{ij}$, for $(i,j) \ne (q,q)$
($K^{qq}_{qq}A^q_{qq}$ is not defined).
Given $i,j,p,q \in \ibr{k}$, with $p = q$, and $(i,j) \ne (q,q)$,
\begin{enumerate}
\item If $q \ne k$, then
\[
K^{qq}_{ij}A_{ij}^q = \begin{cases}
& \frac{\cs^2_{ii}\cs^2_{ij}}{(k-2)\cs^2_{ij} + \cs^2_{ik}},\; i,j \notin\{q,k\}\\
& \vspace*{-0.1in} \\
& \frac{\cs^2_{ii}\cs^2_{ik}}{(k-2)\cs^2_{ij} + \cs^2_{ik}},\; i,j = k\\
& \vspace*{-0.1in} \\
& \frac{\cs^2_{ii}\cs_{ij}\cs_{ik}}{(k-2)\cs^2_{ij} + \cs^2_{ik}},\; i\vb j = k,\; i,j \ne q\\ 
& \vspace*{-0.1in} \\
& -\cs_{ij}\cs_{ii},\; i \vb j = q,\; i,j \ne k\\
& -\cs_{ik}\cs_{ii},\; i \vb j = q,\;i\vb j = k
\end{cases}
\]
\item $q = k$, then
\[
K^{kk}_{ij}A_{ij}^k = \begin{cases}
& \frac{\cs^2_{kk}}{k-1},\; i,j \ne k,\\
& -\cs_{kj}\cs_{kk},\; i \vb j = k
\end{cases}
\]
\end{enumerate}
\begin{rem}
The expressions $K^{qq}_{ij}A^q_{ij}$ are all bounded by $1$. For the type II critical points of $\cal{F}$, $K^{qq}_{ij}A^q_{ij}$ may be very small.
For example, if $q \ne k$ and $i,j \notin\{q,k\}$, then $K^{qq}_{ij}A^q_{ij} = O(k^{-2})$. On the other hand if $i,j = k \ne q$,
$K^{qq}_{ij}A^q_{ij}\approx 1$ for large $k$. \rend
\end{rem}

\begin{lemma}
(Notation and assumptions as above.)
If $p,q,i,j \in \ibr{k}$, $p\ne q$, then
\begin{enumerate}
\item If $p,q < k$
\begin{eqnarray*}
h_1(\ww^p,\ww^q)_{ij} &=&\frac{\sin(\Theta)}{2\pi}\left(\delta_{ij} - A^p_{ij}\right) + 
\frac{1}{2\pi\sin(\Theta)}\left(A_{ij}^q - \cos(\Theta)B^{pq}_{ij} + \cos^2(\Theta) A_{ij}^p\right)
\end{eqnarray*}
\item if $p < k, q = k$
\begin{eqnarray*}
h_1(\ww^p,\ww^k)_{ij} &=&\frac{\tau_k\sin(\Lambda)}{2\pi\tau}\left(\delta_{ij} - A^p_{ij}\right) + 
\frac{\tau_k}{2\pi\tau\sin(\Lambda)}\left(A_{ij}^k - \cos(\Lambda)B^{pk}_{ij} + \cos^2(\Lambda) A_{ij}^p\right)
\end{eqnarray*}
\item if $p = k, q < k$.
\begin{eqnarray*}
h_1(\ww^k,\ww^q)_{ij} &=&\frac{\tau\sin(\Lambda)}{2\pi\tau_k}\left(\delta_{ij} - A^k_{ij}\right) + 
\frac{\tau}{2\pi\tau_k\sin(\Lambda)}\left(A_{ij}^q - \cos(\Lambda)B^{kq}_{ij} + \cos^2(\Lambda) A_{ij}^k\right)
\end{eqnarray*}
\end{enumerate}
If $p,q,i,j \in \ibr{k}$, then
\begin{enumerate}
\item $p \ne q$, $p, q < k$.
\begin{eqnarray*}
h_1(\ww^p,\vv^q)_{ij} & = & \frac{\sin(\alpha_{ij})}{2\pi\tau}\left(\delta_{ij} - A_{ij}^p+K^{pq}_{ij}A_{ij}^p\right) ,\; (i,j)\ne (q,q)\\
&= & \frac{\sin^3(\alpha_{ij})}{\pi\tau},\; i=j=q,
\end{eqnarray*}
\item $p \ne q$, $p < k = q$,
\begin{eqnarray*}
h_1(\ww^p,\vv^k)_{ij} & = & \frac{\sin(\alpha_{ik})}{2\pi\tau}\left(\delta_{ij} - A_{ij}^p+K^{pk}_{ij}A_{ij}^p\right) ,\; (i,j)\ne (k,k)\\
&= & \frac{\sin^3(\alpha_{ik})}{\pi\tau},\; i=j=k,
\end{eqnarray*}
\item $q < k = p$,
\begin{eqnarray*}
h_1(\ww^k,\vv^q)_{ij} & = & \frac{\sin(\alpha_{kj})}{2\pi\tau_k}\left(\delta_{ij} - A_{ij}^k+K^{kq}_{ij}A_{ij}^k\right) ,\; (i,j)\ne (q,q)\\
&= & \frac{\sin^3(\alpha_{kj})}{\pi\tau_k},\; i=j=q,
\end{eqnarray*}
\item $p = q < k$,
\begin{eqnarray*}
h_1(\ww^p,\vv^p)_{ij} & = & \frac{\sin(\alpha_{ii})}{2\pi\tau}\left(\delta_{ij} - A_{ij}^p+K^{pp}_{ij}A_{ij}^p\right) ,\; (i,j)\ne (p,p)\\
&= & \frac{\sin^3(\alpha_{ii})}{\pi\tau},\; i=j=p,
\end{eqnarray*}
\item $p = q = k$,
\begin{eqnarray*}
h_1(\ww^k,\vv^k)_{ij} & = & \frac{\sin(\alpha_{kk})}{2\pi\tau_k}\left(\delta_{ij} - A_{ij}^k+K^{kk}_{ij}A_{ij}^k\right) ,\; (i,j)\ne (k,k)\\
&= & \frac{\sin^3(\alpha_{kk})}{\pi\tau_k},\; i=j=k,
\end{eqnarray*}

\end{enumerate}
\end{lemma}

\begin{prop}[Off diagonal blocks] \label{prop: off_diag}
Assume $p,q,i,j \in \ibr{k}$ and $p \ne q$.
\begin{enumerate}
\item[(A)]
If $p,q < k$, $i = j$, then if
\begin{enumerate}
\item $i\notin \{p,q\}$, $i< k$,
\begin{eqnarray*}
H^{pq}_{ii}&=& \frac{\pi-\Theta}{2\pi} + \frac{\cs^2_{ij}(1 -\cos(\Theta))}{\pi\sin(\Theta)}
\end{eqnarray*}
\item $i \in \{p,q\}$,
\begin{eqnarray*}
H^{pq}_{ii}&=& \frac{\pi-\Theta}{2\pi} + \frac{2\cs_{ii}\cs_{ij}-(\cs^2_{ii}+\cs^2_{ij})\cos(\Theta)}{2\pi\sin(\Theta)}
\end{eqnarray*}
\item $i =k$,
\begin{eqnarray*}
H^{pq}_{kk}&=& \frac{\pi-\Theta}{2\pi} + \frac{\cs^2_{ik}(1 -\cos(\Theta))}{\pi\sin(\Theta)}
\end{eqnarray*}
\end{enumerate}
\item[(B)] If $p\vb q = k$, $i = j$,  then if
\begin{enumerate}
\item $i< k$, $i = p\vb q$, %%(a)
\begin{eqnarray*}
H^{pq}_{ii}&=&  \frac{\pi-\Lambda}{2\pi} + \frac{2\cs_{ii}\cs_{kj}-(\cs^2_{ii}+\cs^2_{kj})\cos(\Lambda)}{2\pi\sin(\Lambda)}
\end{eqnarray*}
\item $i< k$, $i \notin \{p,q\}$, %%(b)
\begin{eqnarray*}
H^{pq}_{ii}&=&  \frac{\pi-\Lambda}{2\pi} + \frac{2\cs_{ij}\cs_{kj}-(\cs^2_{ij}+\cs^2_{kj})\cos(\Lambda)}{2\pi\sin(\Lambda)}
\end{eqnarray*}
\item $i = k$,
\begin{eqnarray*}
H^{pq}_{kk}&=&   \frac{\pi-\Lambda}{2\pi} + \frac{2\cs_{ik}\cs_{kk}-(\cs^2_{ik}+\cs^2_{kk})\cos(\Lambda)}{2\pi\sin(\Lambda)}
\end{eqnarray*}
\end{enumerate}
\item[(C)] If $p,q < k$, $i \ne j$, then if
\begin{enumerate}
\item $i,j\notin \{p,q\}$, $i,j< k$,  %% (a)
\begin{eqnarray*}
H^{pq}_{ij}&=&\frac{\cs^2_{ij}(1 -\cos(\Theta))}{\pi\sin(\Theta)}
\end{eqnarray*}
\item $i\vb j \in \{p,q\}$, $i,j< k$,  %% (b)
\begin{eqnarray*}
H^{pq}_{ij}&=& \frac{\cs_{ij}\cs_{ii}+ \cs^2_{ij}}{2\pi\sin(\Theta)}\big(1-\cos(\Theta)\big)
\end{eqnarray*}
\item $i,j \in \{p,q\}$, $i,j< k$, %% (c)
\begin{eqnarray*}
H^{pq}_{ij}&=& \frac{1}{2\pi\sin(\Theta)}\big(\cs^2_{ii}+\cs^2_{ij}  -2\cos(\Theta)\cs_{ii}\cs_{ij}\big)
\end{eqnarray*}
\item $i \vb j \in \{p,q\}$, $i\vb j =k$, %%(d)
\begin{eqnarray*}
H^{pq}_{ij}&=& \frac{\cs_{ik}(\cs_{ii}+\cs_{ij})}{2\pi\sin(\Theta)}\big(1-\cos(\Theta)\big) 
\end{eqnarray*}
\item $i, j \notin \{p,q\}$, $i\vb j =k$, %%(e)
\begin{eqnarray*}
H^{pq}_{ij}&=& \frac{\cs_{ik}\cs_{ij}}{\pi\sin(\Theta)}\big(1-\cos(\Theta)\big)
\end{eqnarray*}
\end{enumerate}
\item[(D)] If $p\vb q = k$, $i \ne j$, then if
\begin{enumerate}
\item $i,j\notin \{p,q\}$, $i,j< k$, %%(a)
\begin{eqnarray*}
H^{pq}_{ij}&=&  \frac{2\cs_{ij}\cs_{kj}-(\cs^2_{ij}+\cs^2_{kj})\cos(\Lambda)}{2\pi\sin(\Lambda)} 
\end{eqnarray*}
\item $i\vb j \in \{p,q\}$, $i,j< k$, %%(b)
\begin{eqnarray*}
H^{pq}_{ij}&=&  \frac{\cs_{kj}(\cs_{ij}+\cs_{ii}) - \cos(\Lambda)(\cs^2_{kj}+\cs_{ij}\cs_{ii})}{2\pi\sin(\Lambda)}
\end{eqnarray*}
\item $i\vb j =k$, $i,j \in \{p,q\}$. %%(c)
\begin{eqnarray*}
H^{pq}_{ij}&=&  \frac{\cs_{kk}\cs_{ii}+\cs_{kj}\cs_{ik} - \cos(\Lambda)(\cs_{kk}\cs_{kj}+\cs_{ik}\cs_{ii})}{2\pi\sin(\Lambda)}
\end{eqnarray*}
\item $i\vb j=  k$, and $i \vb j \notin \{p,q\}$. %%(d)
\begin{eqnarray*}
H^{pq}_{ij}&=&  \frac{\cs_{kk}\cs_{ij}+\cs_{ik}\cs_{kj} - \cos(\Lambda)(\cs_{kk}\cs_{kj}+\cs_{ik}\cs_{ij})}{2\pi\sin(\Lambda)}
\end{eqnarray*}
\end{enumerate}

\end{enumerate}
In particular, $H^{pq} = H^{qp}$ and the matrices $H^{pq}$ are all symmetric.
\end{prop}

\begin{prop}[Diagonal blocks] \label{prop: diag}
Assume $p,i,j \in \ibr{k}$.
\begin{enumerate}
\item[(A)]
If $i = j$, then if
\begin{enumerate}
\item $i\notin\{ p,k\}$, $p < k$,
\begin{eqnarray*}
H^{pp}_{ii}&=& \frac{1}{2}+\frac{(k-2)\sn^2_{ij}}{2\pi}\left(\sin(\Theta) - \frac{\sn_{ij}}{\tau}\right) + \\
&& \frac{(k-3)}{2\pi}\left(\frac{\cs^2_{ij}(1-\cos(\Theta))^2}{\sin(\Theta)} - \frac{\sn_{ij}\ct^2_{ij}\cs^2_{ij}}{\tau}\right) + \\
&& \frac{(\cs_{ii} - \cos(\Theta)\cs_{ij})^2}{2\pi\sin(\Theta)} - \frac{\sn^3_{ij}}{2\pi\tau} + 
 \frac{\sn^2_{ij}(\tau_k\sin(\Lambda) - \sn_{ik})}{2\pi\tau} - \frac{\sn_{ik}\ct^2_{ik}\cs^2_{ij}}{2\pi\tau} + \\
&& \frac{\tau_k}{2\pi\tau\sin(\Lambda)}\big(\cs_{kj} - \cos(\Lambda)\cs_{ij}\big)^2 - 
 \frac{\sn_{ii}}{2\pi\tau}\left(\sn^2_{ij} + \frac{\cs^2_{ii}\cs^2_{ij}}{(k-2)\cs^2_{ij}+\cs^2_{ik}}\right)
\end{eqnarray*}
\item $i = p$, $p < k$,
\begin{eqnarray*}
H^{pp}_{pp}&=& \frac{1}{2}+\frac{(k-2)\sn^2_{ii}}{2\pi}\left(\sin(\Theta)-\frac{\sn_{ij}}{\tau}\right)+ \\
&& \frac{(k-2)}{2\pi}\left(\frac{\big(\cs_{ij}-\cos(\Theta)\cs_{ii}\big)^2}{\sin(\Theta)} - \frac{\sn_{ij}\cs_{ii}^2\ct^2_{ij}}{\tau}\right) +\\
&& \frac{\sn^2_{ii}}{2\pi\tau}\big(\tau_k\sin(\Lambda) - \sn_{ik}\big) - \frac{\sn^3_{ii}}{\pi\tau} + \\
&& \frac{\tau_k}{2\pi \tau \sin(\Lambda)}\big(\cs_{kj} - \cos(\Lambda)\cs_{ii}\big)^2 - \frac{\sn_{ik}\ct^2_{ik}\cs^2_{ii}}{2\pi\tau}
\end{eqnarray*}
\item if $i =k$, $p < k$,
\begin{eqnarray*}
H^{pp}_{kk}&=& \frac{1}{2} +  \frac{(k-2)\sn^2_{ik}}{2\pi}\left(\sin(\Theta)-\frac{\sn_{ij}}{\tau}\right) + \\
&& \frac{(k-2)\cs^2_{ik}}{2\pi}\left(\frac{(1-\cos(\Theta))^2}{\sin(\Theta)} -  \frac{\sn_{ij}\ct^2_{ij}}{\tau}\right) + \\
&& \frac{\sn_{ik}^2}{2\pi\tau} \big(\tau_k\sin(\Lambda) - \sn_{ii}\big) -  \frac{\sn^3_{ik}}{\pi\tau} + \\
&& \frac{\tau_k}{2\pi\tau\sin(\Lambda)}\big( \cs_{kk}-\cos(\Lambda)\cs_{ik}\big)^2 - \\
&& \frac{\sn_{ii}}{2\pi\tau}\left(\frac{\cs^2_{ii}\cs^2_{ik}}{(k-2)\cs^2_{ij}+\cs^2_{ik}}\right)
\end{eqnarray*}
\item $i\ne k$, $p = k$,
\begin{eqnarray*}
H^{kk}_{ii}&=& \frac{1}{2} + \frac{(k-1)\sn^2_{kj}}{2\pi\tau_k}\big(\tau \sin(\Lambda)-\sn_{kj}\big)+\\
&&\frac{(k-2)}{2\pi\tau_k}\left(\frac{\tau(\cs_{ij} - \cos(\Lambda)\cs_{kj})^2}{\sin(\Lambda)}-\sn_{kj}\cs^2_{kj}\ct^2_{kj}\right) - \\
&& \frac{\sn^3_{kj}}{2\pi \tau_k} - \frac{\sn_{kk}}{2\pi\tau_k}\left(\sn^2_{kj} + \frac{\cs^2_{kk}}{k-1}\right) + \\
&& \frac{\tau}{2\pi\tau_k\sin(\Lambda)}\big(\cs_{ii}-\cos(\Lambda)\cs_{kj}\big)^2
\end{eqnarray*}
\item $i= k$, $p = k$,
\begin{eqnarray*}
H^{kk}_{kk}&=& \frac{1}{2}+\frac{(k-1)\sn^2_{kk}}{2\pi\tau_k}\big(\tau\sin(\Lambda) - \sn_{kj}\big) + \\
&& \frac{(k-1)}{2\pi\tau_k}\left(\frac{\tau(\cs_{ik}-\cos(\Lambda)\cs_{kk})^2}{\sin(\Lambda)} - \ct^2_{kj}\cs^2_{kk}\sn_{kj}\right) - \frac{\sn^3_{kk}}{\pi\tau_k}   
\end{eqnarray*}
\end{enumerate}
\item[(B)]
If  $i \ne j$, then if
\begin{enumerate}
\item $i,j \notin \{p,k\}$, $p < k$,  %% (a) Typical case. This is ZERO when W = V. Most terms are *very* small for large k
\begin{eqnarray*}
H_{ij}^{pp}&=& \frac{\cs^2_{ij}(k-2)}{2\pi}\left(\frac{\sn_{ij}}{\tau} - \sin(\Theta)\right) +\frac{(k-4)\cs^2_{ij}}{2\pi\sin(\Theta)}\big(1-\cos(\Theta)\big)^2 - \\
&& \frac{(k-4)\sn_{ij}\cs^2_{ij}\ct^2_{ij}}{2\pi\tau} + \frac{\sn_{ij}\cs^2_{ij}}{\pi\tau} + \\
&& \frac{\cs_{ij}}{\pi\sin(\Theta)}\big(\cs_{ii} - \cos(\Theta)(\cs_{ii}+ \cs_{ij}) + \cos^2(\Theta)\cs_{ij}\big)  + \\
&&\frac{\cs^2_{ij}}{2\pi\tau}\big(\sn_{ik} - \tau_k\sin(\Lambda)\big) + \frac{\tau_k}{2\pi\tau\sin(\Lambda)}\big(\cs_{kj} - \cos(\Lambda)\cs_{ij}\big)^2 - \\
&& \frac{\sn_{ik}}{2\pi\tau}\ct^2_{ik}\cs^2_{ij}+\frac{\sn_{ii}}{2\pi\tau}\left(\cs^2_{ij} -\frac{\cs^2_{ii}\cs^2_{ij}}{(k-2)\cs^2_{ij}+\cs^2_{ik}}\right)
\end{eqnarray*}
\item $i\vb j = p$, $i,j \ne k$, $p < k$. %%(b)
\begin{eqnarray*}
H_{ip}^{pp}&=& \frac{(k-2)\cs_{ij}\cs_{ii}}{2\pi}\left(\frac{\sn_{ij}}{\tau}-\sin(\Theta)\right)  + \\
&&\frac{(k-3)\cs_{ij}}{2\pi\sin(\Theta)} \big(\cs_{ij} - \cos(\Theta)(\cs_{ii}+\cs_{ij}) + \cos^2(\Theta)\cs_{ii}\big) - \\
&&\frac{(k-3)\sn_{ij}\cs_{ij}\cs_{ii}\ct^2_{ij}}{2\pi\tau} + \\
&&\frac{\big(\cs_{ii}\cs_{ij}-\cos(\Theta)(\cs^2_{ii}+\cs^2_{ij}) + \cos^2(\Theta)\cs_{ii}\cs_{ij}\big)}{2\pi\sin(\Theta)} +  \frac{\sn_{ij}\cs_{ij}\cs_{ii}}{2\pi\tau}  +\\
&& \frac{\cs_{ij}\cs_{ii}}{2\pi\tau}\left(\sn_{ik}-\tau_k\sin(\Lambda)\right)- \frac{\sn_{ik}}{2\pi\tau}\cs_{ii}\cs_{ij}\ct^2_{ik} +\\
&& \frac{\tau_k}{2\pi\tau\sin(\Lambda)} \big(\cs^2_{kj}-\cos(\Lambda)(\cs_{kj}[\cs_{ij}+\cs_{ii}]) + \cos^2(\Lambda)\cs_{ii}\cs_{ij}\big) +  \frac{\sn_{ii}}{\pi\tau}\cs_{ii}\cs_{ij}
\end{eqnarray*}
\item $i \vb j = k$, $i,j \ne p$, $p < k$,  %%(c)
\begin{eqnarray*}
H^{pp}_{ik}&=& \frac{(k-2)\cs_{ij}\cs_{ik}}{2\pi}\left(\frac{\sn_{ij}}{\tau} - \sin(\Theta)\right) + \\
&& \frac{(k-3)\cs_{ij}\cs_{ik}}{2\pi\sin(\Theta)}\big(1-\cos(\Theta)\big)^2-\frac{(k-3)\sn_{ij}\cs_{ij}\cs_{ik}\ct^2_{ij}}{2\pi\tau}+\\
&& \frac{\cs_{ik}\big(\cs_{ii} -\cos(\Theta)(\cs_{ii} + \cs_{ij}) + \cos^2(\Theta) \cs_{ij}\big)}{2\pi\sin(\Theta)} +  \frac{\sn_{ij}\cs_{ij}\cs_{ik}}{2\pi\tau} + \\
&&  \frac{\cs_{ij}\cs_{ik}}{2\pi\tau}\big(\sn_{ik}-\tau_k\sin(\Lambda) \big)  + \\
&&  \frac{\tau_k\big(\cs_{kk}\cs_{kj}-\cos(\Lambda)(\cs_{kk}\cs_{ij}+\cs_{ik}\cs_{kj})+\cos^2(\Lambda)\cs_{ij}\cs_{ik}\big)}{2\pi\tau\sin(\Lambda)}+ \\
&&  \frac{\sn_{ik}\cs_{ik}\cs_{ij}}{2\pi\tau}+ \frac{\sn_{ii}}{2\pi\tau}\left(\cs_{ij}\cs_{ik} - \frac{\cs^2_{ii}\cs_{ij}\cs_{ik}}{(k-2)\cs^2_{ij}+\cs^2_{ik}}\right)
\end{eqnarray*}
\item $i,j \in \{p,k\}$,  $p < k$,  %%(d)
\begin{eqnarray*}
H^{pp}_{pk}&=&\frac{(k-2)\cs_{ii}\cs_{ik}}{2\pi} \left(\frac{\sn_{ij}}{\tau} - \sin(\Theta)\right) + \\
&&\frac{(k-2)\cs_{ik}}{2\pi\sin(\Theta)}\big(\cs_{ij} - \cos(\Theta)(\cs_{ii}+\cs_{ij})+ \cos^2(\Theta)\cs_{ii}\big) - \\
&& \frac{(k-2)\sn_{ij}\cs_{ii}\cs_{ik}\ct^2_{ij}}{2\pi\tau}+\frac{\cs_{ii}\cs_{ik}}{2\pi\tau}\big(\sn_{ik} - \tau_k\sin(\Lambda)\big)+\\
&& \frac{\tau_k\big(\cs_{kk}\cs_{kj}-\cos(\Lambda)(\cs_{ii}\cs_{kk}+\cs_{ik}\cs_{kj})+\cos^2(\Lambda)\cs_{ii}\cs_{ik}\big)}{2\pi\tau\sin(\Lambda)} + \\
&& \frac{\sn_{ik}\cs_{ii}\cs_{ik}}{2\pi\tau} + \frac{\sn_{ii}\cs_{ii}\cs_{ik}}{\pi\tau}
\end{eqnarray*}
\item $i,j \ne p$, $p = k$, %%(e)
\begin{eqnarray*}
H^{kk}_{ij}&=& \frac{(k-1)\cs^2_{kj}}{2\pi\tau_k}\big(\sn_{kj}-\tau \sin(\Lambda)\big) + \frac{(k-3)\tau}{2\pi\tau_k\sin(\Lambda)}\big(\cs_{ij} - \cos(\Lambda)\cs_{kj}\big)^2 - \\
&& \frac{(k-3)\sn_{kj}\ct^2_{kj}\cs^2_{kj}}{2\pi\tau_k} + \frac{\sn_{kj}\cs^2_{kj}}{\pi\tau_k} + \\
&& \frac{\tau}{\pi\tau_k\sin(\Lambda)} \big(\cs_{ii}\cs_{ij} - \cos(\Lambda)\cs_{kj}(\cs_{ii}+\cs_{ij}) + \cos^2(\Lambda)\cs^2_{kj}\big) + \\
&& \frac{\sn_{kk}}{2\pi\tau_k}\left(\cs^2_{kj} - \frac{\cs^2_{kk}}{k-1}\right)
\end{eqnarray*}
\item $i\vb j = p$, $p = k$, %%(f)
\begin{eqnarray*}
H^{kk}_{ik}&=&  \frac{(k-1)\cs_{kk}\cs_{kj}}{2\pi\tau_k}\big(\sn_{kj} -\tau\sin(\Lambda)\big) - \frac{(k-2)\sn_{kj}}{2\pi\tau_k}\ct^2_{kj}\cs_{kk}\cs_{kj} +\\
&&  \frac{(k-2)\tau\big(\cs_{ij}\cs_{ik}-\cos(\Lambda)(\cs_{kk}\cs_{ij}+\cs_{ik}\cs_{kj})+\cos^2(\Lambda)\cs_{kk}\cs_{kj}\big)}{2\pi\tau_k\sin(\Lambda)} + \\
&&  \frac{\sn_{kj}}{2\pi\tau_k}\cs_{kk}\cs_{kj} + \frac{\sn_{kk}}{\pi\tau_k}\cs_{kk}\cs_{kj} +\\
&&  \frac{\tau}{2\pi\tau_k\sin(\Lambda)}\big(\cs_{ii}\cs_{ik}-\cos(\Lambda)(\cs_{kk}\cs_{ii}+\cs_{ik}\cs_{kj})+\cos^2(\Lambda)\cs_{kk}\cs_{kj}\big)
\end{eqnarray*}
\end{enumerate}
\end{enumerate}
\end{prop}

	\section{Estimating the Hessian spectrum}\label{sec: D}
The stage is now set for deriving the estimates for the Hessian spectrum. We first present a 
detailed derivation of the spectrum of type II minima which follows 
along the same lines of \pref{exam: global}, and 
then briefly state the adjustments needed 
for the analysis of types A and I. 
%In \pref{sec: evs_emp}, we provide numerical 
%computations of the different spectra. 
%for 
%$k=5\dots15,20,50,100$. 
%Empirical results confirm our analysis (see \pref{sec: empirical}).

\subsection{The spectrum at type II minima} \label{sec: 
type2spectrum}
First, we use the infinite series representation for type II minima given in \pref{lem: 
asym} to 
obtain the following estimates
(notations as in \pref{prop: 
off_diag} and \pref{prop: diag}):
\begin{enumerate}
	\item $\cos(\Theta) = (2e_4 + 4)k^{-2}+ 2e_5 k^{-\frac{5}{2}}$.
	\item $\sin(\Theta) = 1 + O(k^{-4})$
	\item $\cos(\Lambda) =-(e_4+2)k^{-1}-e_5k^{-\frac{3}{2}}$.
	\item $\sin(\Lambda) = 1- \frac{(e_4 + 2)^2}{2k^2}- (e_4 + 
	2)e_5 
	k^{-\frac{5}{2}}$.
	\item $\cs_{ii} =1 - \frac{2}{k^2}$.
%	\item $\cs_{ii}^2 =1 - \frac{4}{k^2}$.
	\item $\sn_{ii} = 2k^{-1} + (\frac{e_4^2}{4} + 2-d_2)k^{-2}$.
%	\item $\sn_{ii}^2 = 4k^{-2} + (e_4^2 + 4(2-d_2))k^{-3}$.
%	\item $\sn_{ii}^3 = 8k^{-3} + (3e_4^2 + 12(2-d_2))k^{-4}$.
	\item $\cs_{ij} =\frac{e_4}{k^2} + e_5 k^{-\frac{5}{2}}$.
%	\item $\cs_{ij}^2 =\frac{e^2_4}{k^4}$.  
	\item $\sn_{ij} = 1 + O(k^{-4})$.
%	\item $\sn_{ij}^p = 1 + O(k^{-4})$, $p > 1$.
	\item $\cs_{ik} =2k^{-1} + (2-d_2)k^{-2}$.
%	\item $\cs_{ik}^2 =4k^{-2} + 4(2-d_2)k^{-3}$.
	\item $\sn_{ik} = 1- 2k^{-2}$.
%	\item $\sn_{ik}^2 = 1- 4k^{-2}$.
	\item $\cs_{kk}  =-1 + \frac{e_4^2}{2k} + 
	e_4e_5k^{-\frac{3}{2}}$.
%	\item $\cs_{kk}^2  =1 - \frac{e_4^2}{k} - 
%	2e_4e_5k^{-\frac{3}{2}}$.
	\item $\sn_{kk} = -\frac{e_4}{\sqrt{k}} -e_5 k^{-1}$.
%	\item $\sn_{kk}^2 = \frac{e^2_4}{k} +2e_4e_5 k^{-\frac{3}{2}}$.
%	\item $\sn_{kk}^3 = -e^3_4k^{-\frac{3}{2}}  -3e^2_4e_5 k^{-2}$.
	\item $\cs_{kj} =-e_4k^{-1} -e_5k^{-\frac{3}{2}}$.
%	\item $\cs_{kj}^2 =e_4^2k^{-2} +2e_4e_5k^{-\frac{5}{2}}$.
	\item $\sn_{kj} = 1-\frac{e_4^2}{2k^2}$.
%	\item $\sn^2_{kj} = 1-\frac{e_4^2}{k^2}$.
%	\item $\cs_{ii}\cs_{ij} = \frac{e_4}{k^2} - \frac{2e_4}{k^4} $.
%	\item $\cs_{ii}\cs_{kj} = -e_4k^{-1} -e_5k^{-\frac{3}{2}}$.
%	\item $\cs_{ii}\cs_{ik} = 2k^{-1} + (2-d_2)k^{-2}$.
%	\item $\cs_{ii}\cs_{kk} = -1 + \frac{e_4^2}{2k} + 
%	e_4e_5k^{-\frac{3}{2}}$.
%	\item $\cs_{ij}\cs_{ik} = \frac{2e_4}{k^3} $.
%	\item $\cs_{ij}\cs_{kj} = -e_4^2k^{-3}$.
%	\item $\cs_{ij}\cs_{kk} = -\frac{e_4}{k^2} - e_5 
%	k^{-\frac{5}{2}}$.
%	\item $\cs_{ik}\cs_{kj} = -e_4^2k^{-3}$.
%	\item $\cs_{ik}\cs_{kk} =-2k^{-1}+(-2+d_2+e_4^2)k^{-2}$.
%	\item $\cs_{kj}\cs_{kk} =e_4k^{-1} + e_5 k^{-\frac{3}{2}}$.
%	\item $\ct_{ij}^2 = e_4^2 k^{-4}$.
%	\item $\ct_{ik}^2 = 4k^{-2}+4(2-d_2)k^{-3}$.
%	\item $\ct_{kj}^2 = e_4^2k^{-2} +2e_4e_5k^{-\frac{5}{2}}$.
%	\item $A = (k-1)\cs_{ij}^2 + \cs_{ik}^2 = 4k^{-2} + (e_4^2+ 
%	4(2-d_2))k^{-3}$.
%	\item $\cs_{ii}^2\cs_{ij}\cs_{ik}/A  =\frac{e_4}{2k} + 
%	\frac{e_5}{2} 
%	k^{-\frac{3}{2}}-\frac{e_4(e_4^2 + 4(2-d_2))}{8}k^{-2}$.
%	\item $\cs_{ii}^2\cs^2_{ij}/A  
%	=\frac{e_4^2}{4k^2}-\frac{e_4^2(e_4^2+ 
%		4(2-d_2))}{16k^3}$.
%	\item $\cs_{ii}^2\cs^2_{ik}/A  =1 - \frac{e_4^2}{4k} + 
%	O(k^{-2})$.
	\item
	$\tau = 1 + (c_4+2)k^{-2} + c_5 k^{-\frac{5}{2}}$
	\item
	$\tau^{-1} = 1 - (c_4 + 2)k^{-2} - c_5 k^{-\frac{5}{2}}$.
	\item 	$\tau_k= 1 + \frac{e_4^2 -2d_2}{2}k^{-1} 
	+(e_4e_5-d_3)k^{-\frac{3}{2}}$
	\item $\tau_k^{-1}= 1 - \frac{e_4^2 -2d_2}{2}k^{-1} 
	-(e_4e_5-d_3)k^{-\frac{3}{2}}$
%	\item $\tau_k^{-1}\tau  =  1 - \frac{e_4^2-2d_2}{2}k^{-1} 
%	-(e_4e_5-d_3)k^{-\frac{3}{2}}$
%	\item $\tau_k\tau^{-1}  =  1 + \frac{e_4^2-2d_2}{2}k^{-1} 
%	+(e_4e_5-d_3)k^{-\frac{3}{2}}$
	%(\tau_k\tau)^{-1} &=& 1 -  \frac{e_4^2-2d_2}{2}k^{-1} 
	%-(e_4e_5-d_3)k^{-\frac{3}{2}}. 
\end{enumerate}
Next, we use these `primitive' estimates to compute the Hessian entries. The estimates 
for the entries of off-diagonal blocks are obtained through the respective 
expressions in \pref{prop: off_diag}, see \pref{table: est_off}.
\begin{table}[H]
\begin{tabular}{l|l|l|l}
Case& Sub-case& Hessian entry & Estimate\\\hline 
A& a& ${H^{pq}_{ii}}$ & $\frac{1}{4} + O\left(k^{-2}\right)$ \\
A& b& ${H^{pq}_{ii}}$ & $\frac{1}{4} + O\left(k^{-2}\right)$ \\
A& c& ${H^{pq}_{kk}}$ & $\frac{1}{4} + O\left(k^{-2}\right)$ \\
B& a& ${H^{pq}_{ii}}$ & $\frac{1}{4} - \frac{e_{4}}{ \pi} k^{-1} - \frac{e_{5} 
k^{-1.5}}{\pi} + 
O\left(k^{-2}\right)$ \\
B& b& ${H^{pq}_{ii}}$ & $\frac{1}{4} + \left(- \frac{e_{4}}{2 \pi} 
- \frac{1}{\pi}\right)k^{-1} - \frac{e_{5} k^{-1.5}}{2 \pi} + 
O\left(k^{-2}\right)$ \\
B& c& ${H^{pq}_{ii}}$ & $\frac{1}{4} - \frac{2}{\pi}k^{-1}  + 
O\left(k^{-2}\right)$ \\
C& a& ${H^{pq}_{ij}}$ & $O\left(k^{-2}\right)$ \\
C& b& ${H^{pq}_{ij}}$ & $O\left(k^{-2}\right)$ \\
C& c& ${H^{pq}_{ij}}$ & $\frac{1}{2 \pi} + O\left(k^{-2}\right)$ \\
C& d& ${H^{pq}_{ij}}$ & $\frac{k^{-1}}{\pi} + O\left(k^{-2}\right)$ \\
C& e& ${H^{pq}_{ij}}$ & $O\left(k^{-2}\right)$ \\
D& a& ${H^{pq}_{ij}}$ & $O\left(k^{-2}\right)$ \\
D& b& ${H^{pq}_{ij}}$ & $- \frac{e_{5} k^{-1.5}}{2 \pi} - \frac{e_{4} 
	k^{-1}}{2 \pi} + O\left(k^{-2}\right)$ \\
D& c& ${H^{pq}_{ij}}$ & $- \frac{1}{2 \pi} + \frac{e_{4} e_{5} 
	k^{-1.5}}{2 \pi} + \frac{e_{4}^{2} k^{-1}}{4 \pi} + 
O\left(k^{-2}\right)$ \\
D& d& ${H^{pq}_{ij}}$ & $O\left(k^{-2}\right)$ \\
\end{tabular}
\caption{Estimates for the entries of off-diagonal blocks of the Hessian using the 
expression derived in \pref{prop: off_diag}.   }\label{table: est_off}
\end{table}
Similarly, the estimates for the entries of the diagonal blocks are obtained through 
the relevant expressions in \pref{prop: diag}.
\begin{table}[H]
\begin{tabular}{l|lll}
Hessian entry & Estimate\\\hline 
${H^{pp}_{ii}}$ & $0.5 + k^{-1.5} \left(\frac{0.5 c_{5}}{\pi} 
- \frac{0.5 d_{3}}{\pi} + \frac{0.5 e_{4}}{\pi} e_{5}\right) 
+ k^{-1.0} \left(\frac{0.5 c_{4}}{\pi} - \frac{0.5 
	d_{2}}{\pi} + \frac{0.25 e_{4}^{2}}{\pi}\right) + 
\mathcal{O}\left(k^{-2}\right)$ \\
${H^{pp}_{pp}}$ & $0.5 + \mathcal{O}\left(k^{-2}\right)$ \\
${H^{pp}_{kk}}$ & $0.5 + k^{-1.5} \left(\frac{0.5 c_{5}}{\pi} 
- \frac{1.0 d_{3}}{\pi}\right) + k^{1.0} \left(\frac{0.5 
	c_{4}}{\pi} - \frac{1.0 d_{2}}{\pi} + \frac{1.0}{\pi}\right) 
+ \mathcal{O}\left(k^{-2}\right)$ \\
${H^{kk}_{ii}}$ & $0.5 + k^{-1.5} \left(\frac{0.5 c_{5}}{\pi} 
+ \frac{0.5 d_{2}}{\pi} e_{4} - \frac{0.25 e_{4}^{3}}{\pi} - 
\frac{0.5 e_{4}}{\pi} e_{5} + \frac{e_{4}}{2 \pi} - 
\frac{1.0 e_{5}}{\pi}\right)$\\
& $+ k^{-1.0} \left(\frac{0.5  
	c_{4}}{\pi} - \frac{1.0 e_{4}}{\pi} + \frac{e_{5}}{2 
	\pi}\right) + \frac{e_{4} k^{-0.5}}{2 \pi} + 
\mathcal{O}\left(k^{-2}\right)$ \\
${H^{kk}_{kk}}$ & $0.5 + \frac{e_{4}^{3} k^{-1.5}}{\pi} + 
\mathcal{O}\left(k^{-2}\right)$ \\
${H^{pp}_{ij}}$ & $\mathcal{O}\left(k^{-2}\right)$ \\
${H^{pp}_{ip}}$ & $\mathcal{O}\left(k^{-2}\right)$ \\
${H^{pp}_{ik}}$ & $k^{-1.0} \left(\frac{0.5 e_{4}}{\pi} + 
\frac{1}{\pi}\right) + \frac{0.5 e_{5}}{\pi} k^{-1.5} + 
\mathcal{O}\left(k^{-2}\right)$ \\
${H^{pp}_{pk}}$ & $\mathcal{O}\left(k^{-2}\right)$ \\
${H^{kk}_{ij}}$ & $\frac{e_{4} k^{-1.5}}{2 \pi} + 
\mathcal{O}\left(k^{-2}\right)$ \\
${H^{kk}_{ik}}$ & $- \frac{e_{4}^{2} k^{-1.5}}{\pi} + 
\mathcal{O}\left(k^{-2}\right)$ 
\end{tabular}
\caption{Estimates for the entries of diagonal blocks of the Hessian using \pref{prop: diag}.   }
\end{table}
Our next goal is to compute the product of the Hessian of type II minima by the 
representative vectors described in \pref{sec: iso_decomp_Mkk}. 

\paragraph{The eigenvalues $\lambda_\mathfrak{x}$ and 
	$\lambda_\mathfrak{y}$.}
Computing $H(\mathfrak{X}^{k-1,1})_{12} = 
\langle\mathbf{r}_2,\mathfrak{X}^{k-1,1}\rangle$, we find that
\[
H(\mathfrak{X}^{k-1,1})_{12} 
= H^{11}_{22} - H^{11}_{ij} - H^{12}_{12} + 2H^{12}_{1j} - 
H^{12}_{33}=\frac{1}{4}-\frac{1}{2\pi}-\frac{1}{\pi k}
+ O(k^{-2}).
\]
Since  $\mathfrak{X}^{k-1,1}_{12} = 1$, $\lambda_\mathfrak{x} 
=\frac{1}{4} 
-\frac{1}{2\pi} - \frac{1}{\pi k} + O(k^{-2})$.
Along similar lines. we find that 
\begin{align*}
H(\mathfrak{Y}^{k-1,1})_{12} =&
(k-4)H^{11}_{22} - (k-4)H^{11}_{ij} + (k-4)H^{12}_{12}\\
&- 2(k-4))H^{12}_{1j} - 
(k-4)H^{12}_{33} + 2(k-4)H^{12}_{ij}\\
=&\frac{1}{4}+\frac{1}{2\pi}-\frac{1}{\pi k}+ O(k^{-2}),
\end{align*}
by which we conclude  
$\lambda_\mathfrak{y} =\frac{1}{4} 
+\frac{1}{2\pi} - \frac{1}{\pi k} + O(k^{-2})$.
\begin{rems}
	(1) The $1/k$ term that occurs for both eigenvalues appears to 
	be a 
	correction in going from $k$ to $k-1$---to the $(k-1)^2$ block 
	in 
	$M(k,k)$.\\
	(2) Using the results in~\cite{ArjevaniField2020} it is not 
	difficult to 
	compute the coefficient of $k^{-2}$ in power series (in 
	$1/\sqrt{k}$) for 
	both eigenvalues. 
	Although the coefficient of $k^{-\frac{3}{2}}$ is zero, the 
	coefficients 
	of higher order fractional powers of $1/k$ are typically 
	non-zero. \rend
\end{rems}
\paragraph{The eigenvalues associated to $\mathfrak{s}_k$.}
As in the case of $\W=\V$ (Example \pref{exam: global}), we denote the matrix 
associated to the factor 
$3\mathfrak{s}_k$ by $B_{\fs}\in 
M(3,3)$, and use \pref{table: hess_by_std} to show that
modulo $o(1)$ terms,
\begin{align*}
B_{\fs}= \begin{pmatrix}
0.5& -0.25 k& 0.25k& 0& 0.25\\
-0.125&   0.125k + 0.5&       -0.125k&       0&  -0.125\\
0.125& -0.125k - 0.25& 0.125k + 0.25&       0&   0.125\\
0&0&0&0.25&-0.5/\pi\\
0.25&-0.25k&0.25k&-0.5/\pi&0.5
\end{pmatrix}
\end{align*}
This allows us to compute the coefficients of the linear term for eigenvalues 
of the form $a+bk+o(1)$. Indeed, taking the limit of $B_{\fs}/k$ for 
$k\to\infty$, we have that the $b$-coefficients are the eigenvalues of 
\begin{align*}
%B_{\fs}' = 
\begin{pmatrix}
0& -0.25&  0.25& 0& 0\\
0&  0.125& -0.125& 0& 0\\
0& -0.125&  0.125& 0& 0\\
0&      0&      0& 0& 0\\
0&  -0.25&   0.25& 0& 0
\end{pmatrix}
\end{align*}
which are zero, except for a single eigenvalue which equals $0.25$. 
\paragraph{The eigenvalues associated to $\mathfrak{t}_k$.}
The matrix associated to the factor $3\mathfrak{t}_k$, $B_{\ft} = [\beta_{ij}]\in 
M(3,3)$ is computed through Table \pref{table: hess_by_tri}. Modulo $o(1)$ 
terms, we have
\begin{align*}
B_{\ft}= 
\begin{pmatrix}
0.5k/\pi + 0.5&       0.25k&0.25&0&-0.5/\pi\\
0.25&0.25k + 0.5&0.25&0&0\\
0.25&0.25k&0.5&-0.5/\pi&0\\
0&0&-0.5/\pi&0.25k + 0.5&0.25\\
-0.5k/\pi + 1.0&1.0&1.0&0.25k + 1.0&0.5
\end{pmatrix}
\end{align*}
This allows us to easily compute the coefficients of the linear term for eigenvalues 
of the form $a+bk+o(1)$. Indeed, this follows by computing the spectrum of $B_{\ft}/k$ where 
$k\to\infty$, 
\begin{align*}
%B_{\ft}' = 
\begin{pmatrix}
0.5/\pi&0.25&0&0&0\\
0&0.25&0&0&0\\
0&0.25&0&0&0\\
0&0&0&0.25&0\\
-0.5/\pi&0&0&0.25&0
\end{pmatrix}
\end{align*}
which is 0, $\frac{1}{4}$ and $\frac{1}{2\pi}$ of multiplicity $2,1,2$, respectively.
\begin{table}
\begin{tabular}{l|ll}
	&$\fS_1^{k-1,k-1}$ \\\hline
	$H(\cdot)_{11}$&$H^{11}_{11} - H^{12}_{12}$\\
	$H(\cdot)_{12}$&$H^{11}_{1j} - H^{12}_{11}$\\
	$H(\cdot)_{13}$&$H^{11}_{1j} - H^{12}_{1j}$\\
	$H(\cdot)_{1k}$&$H^{11}_{1k} - H^{12}_{1k}$\\
	$H(\cdot)_{k1}$&$H^{1k}_{11} - H^{1k}_{1j}$\\\hline
	&$\fS_2^{k-1,k-1}$\\\hline
	$H(\cdot)_{11}$&$(k-1)H^{11}_{1j} - (k-1)H^{12}_{11}$\\
	$H(\cdot)_{12}$&$2H^{11}_{22} + (k-3)H^{11}_{ij} - 2H^{12}_{12} - 
	2(k-3)H^{12}_{1j} + (k-3)H^{12}_{33}$\\
	$H(\cdot)_{13}$&$H^{11}_{22} + (k-2)H^{11}_{ij} - H^{12}_{12} - (k-3)H^{12}_{1j} - 
	H^{12}_{33}$\\
	$H(\cdot)_{1k}$&$(k-1)H^{11}_{ik} - (k-1)H^{12}_{1k}$\\
	$H(\cdot)_{k1}$&$(k-1)H^{1k}_{1j} - (k-1)H^{1k}_{22}$\\\hline
	&$\fS_3^{k-1,k-1}$\\\hline
	$H(\cdot)_{11}$&$(k-3)H^{11}_{1j} + (k-3)H^{12}_{11} - 2(k-3)H^{12}_{1j}$\\
	$H(\cdot)_{12}$&$(k-3)H^{11}_{ij} - (k-3)H^{12}_{33}$\\
	$H(\cdot)_{13}$&$H^{11}_{22} + (k-4)H^{11}_{ij} + H^{12}_{12} + (k-5)H^{12}_{1j} - 
	H^{12}_{33} - 2(k-4)H^{12}_{ij}$\\
	$H(\cdot)_{1k}$&$(k-3)H^{11}_{ik} + (k-3)H^{12}_{1k} - 2(k-3)H^{12}_{ik}$\\
	$H(\cdot)_{k1}$&$(k-3)H^{1k}_{1j} + (k-3)H^{1k}_{22} - 2(k-3)H^{1k}_{ij}$\\\hline
	&$\fS_4^{k-1,1}$\\\hline
	$H(\cdot)_{11}$&$H^{11}_{1k} - H^{12}_{1k}$\\
	$H(\cdot)_{12}$&$H^{11}_{ik} - H^{12}_{1k}$\\
	$H(\cdot)_{13}$&$H^{11}_{ik} - H^{12}_{ik}$\\
	$H(\cdot)_{1k}$&$H^{11}_{kk} - H^{12}_{kk}$\\
	$H(\cdot)_{k1}$&$H^{1k}_{1k} - H^{1k}_{ik}$\\\hline
	&$\fS_5^{1,k-1}$\\\hline
	$H(\cdot)_{11}$&$H^{1k}_{11} - H^{1k}_{1j}$\\
	$H(\cdot)_{12}$&$H^{1k}_{1j} - H^{1k}_{22}$\\
	$H(\cdot)_{13}$&$H^{1k}_{1j} - H^{1k}_{ij}$\\
	$H(\cdot)_{1k}$&$H^{1k}_{1k} - H^{1k}_{ik}$\\
	$H(\cdot)_{k1}$&$H^{kk}_{11} - H^{kk}_{ij}$\\
\end{tabular}
\caption{Product of the Hessian matrix by the representative vectors of the standard 
	representation, as described in \pref{sec: iso_decomp_Mkk}} \label{table: 
	hess_by_std}
\end{table}

\begin{table}[H]
\begin{tabular}{l|ll}
	&$\fD_1^{k-1,k-1}$ \\\hline
	$H(\cdot)_{11}$&
	$H^{11}_{11} + (k-2)H^{12}_{12}$
	\\
	$H(\cdot)_{12}$& $H^{11}_{1i} + H^{12}_{11} + (k-3)H^{12}_{1j}$
	\\
	$H(\cdot)_{1k}$& $H^{11}_{1k} + (k-2) H^{12}_{1k}$ \\
	$H(\cdot)_{k1}$& $H^{1k}_{11} + (k-2) H^{1k}_{1j}$\\
	$H(\cdot)_{kk}$& $(k-1)H^{1k}_{1k}$\\
	\hline
	&$\fD_2^{k-1,k-1}$\\\hline
	$H(\cdot)_{11}$&$(k-2)H^{11}_{1j} + (k-2)H^{12}_{11} + (k-2)(k-3)H^{12}_{1j}$\\
	$H(\cdot)_{12}$&$H^{11}_{22} + (k-3)H^{11}_{ij} + H^{12}_{12} + 2(k-3)H^{12}_{1j} 
	+ (k-3)H^{12}_{33} + (k^2-7k+8)H^{12}_{ij}$\\
	$H(\cdot)_{1k}$&$(k-2)H^{11}_{ik} + (k-2)H^{12}_{1k} + (k-2)(k-3)H^{12}_{ik}$\\
	$H(\cdot)_{k1}$&$(k-2)H^{1k}_{1j} + (k-2)H^{1k}_{22} + (k-2)(k-3)H^{1k}_{ij}$ \\
	$H(\cdot)_{kk}$&$(k-1)(k-2)H^{1k}_{ik}$\\\hline
	&$\fD_3^{k-1,1}$\\\hline
	$H(\cdot)_{11}$&$H^{1k}_{11} + (k-2)H^{1k}_{1j}$\\
	$H(\cdot)_{12}$&$H^{1k}_{1j} + H^{1k}_{22} + (k-3)H^{1k}_{ij}$ \\
	$H(\cdot)_{1k}$&$H^{1k}_{1k} + (k-2)H^{1k}_{ik}$\\
	$H(\cdot)_{k1}$&$H^{kk}_{11} + (k-2)H^{kk}_{ij}$\\
	$H(\cdot)_{kk}$&$(k-1)H^{kk}_{ik}$\\\hline
	&$\fD_3^{1,k-1}$\\\hline
	$H(\cdot)_{11}$&$H^{11}_{1k} + (k-2)H^{12}_{1k}$\\
	$H(\cdot)_{12}$&$H^{11}_{ik} + H^{12}_{1k} + (k-3)H^{12}_{ik}$\\
	$H(\cdot)_{1k}$&$H^{11}_{kk} + (k-2)H^{12}_{kk}$\\
	$H(\cdot)_{k1}$&$H^{1k}_{1k} + (k-2)H^{1k}_{ik}$\\
	$H(\cdot)_{kk}$&$(k-1)H^{1k}_{kk}$\\\hline
	&$\fD_1^{1,1}$\\\hline
	$H(\cdot)_{11}$&$H^{1k}_{1k}$\\
	$H(\cdot)_{12}$&$H^{1k}_{ik}$\\
	$H(\cdot)_{1k}$&$H^{1k}_{kk}$\\
	$H(\cdot)_{k1}$&$H^{kk}_{ik}$\\
	$H(\cdot)_{kk}$&$H^{kk}_{kk}$\\
\end{tabular}
\caption{Product of the Hessian matrix by the representative vectors of the trivial 
representation, as described in \pref{sec: iso_decomp_Mkk}\label{table: 
	hessian_by_tri}}\label{table: 
hess_by_tri}
\end{table}
\subsection{The spectrum of type A and type I minima} \label{section: 
spec}
The computation of the Hessian spectrum at types A and I uses the estimates 
derived in \cite{ArjevaniField2020}, which we provide here for convenience. Modulo high-order terms, we 
have (notations as in 
\pref{lem: asym})
\begin{align*}
\text{Type A: }\quad & \xi_1, \xi_5 \sim -1 + 2k^{-1}  + 
\prn*{\frac{8}{\pi}-4}k^{-2}    ,\quad \xi_2, \xi_3, 
\xi_4\sim 
2k^{-1}  + \prn*{\frac{4}{\pi}-2}k^{-2}.
\end{align*}
\[
\text{Type I: }\quad \xi_1 = -1 + \sum_{n=2}^\infty c_n 
k^{-\frac{n}{2}},\quad
\xi_2 = \sum_{n=2}^\infty e_n k^{-\frac{n}{2}},\quad \xi_5 = 1 + 
\sum_{n=2}^\infty d_n k^{-\frac{n}{2}}, \]
\[
\xi_3 = \sum_{n=2}^\infty f_n k^{-\frac{n}{2}},\quad
\xi_4 = \sum_{n=4}^\infty g_n k^{-\frac{n}{2}},
\]
where
\[
\begin{matrix}
c_2 = & 2              &d_2= &\frac{8(\pi - 1)}{\pi^2}& e_2   = & 
2                  &f_2  =&0 &g_2 =& 2-\frac{4}{\pi} \\
c_3 = & 0              &d_3=& -4.798751&e_3   = & 
0              &f_3=&0                      &g_3 
=& \frac{32}{\pi^2}\prn*{\frac{1}{\pi}-1} \\
c_4 =&\frac{16}{\pi}-4 &&                             & e_4   
= & 
\frac{8}{\pi}-2 & f_4=& \frac{16}{\pi^2} - 
\frac{12}{\pi}                   & \\
c_5 = &4.441691&&&e_5 =& \frac{8(\pi^2 +4(\pi-1))}{\pi^3}&f_5 
=& 6.205827
\end{matrix}
\]
The rest of the derivation follows along the same lines of type II minima. \\

Let us show how to compute the 3 distinct eigenvalues of type A which are related to 
the standard representation. Here, the matrix associated with the $3\fs$ factor is 
\begin{align*}
M = \begin{pmatrix}
-0.5/\pi + 0.5&2.0/\pi - 0.25k&-0.5 + 0.25k\\
-0.125&0.5/\pi + 0.25 + 0.125k&0.25 - 0.125k\\
0.125&-0.125k&-0.5/\pi + 0.125k
\end{pmatrix}.
\end{align*}
We now express the 3 eigenvalues by $a_ik + b_i  + o(1)$. The coefficients of the 
linear terms can be computed by taking $k\to\infty$ in  $M/k$, 
which gives 
\begin{align*}
\begin{pmatrix}
0&-0.25&0.25\\
0&0.125&-0.125\\
0&-0.125&0.125
\end{pmatrix},
\end{align*}
whose eigenvalues are easily shown to be $0,0,0.25$. It 
remains to compute the constant terms $b_i$. To this end, note that
\begin{align*}
b_1 + b_2  + b_3 &= \text{the constant term of } \trace\prn{M} = -0.5/\pi + 
0.75,\\
2a_1b_1  &= \text{the coefficient of } k \text{ in } \trace\prn{M^2} = 0.125,\\
a_1b_1b_2 &= \text{the coefficient of } k \text{ in } \det\prn{M} = 
0.03125/\pi + 0.015625.
\end{align*}
The system of equations yields $b_1 = 
\frac{1}{4}, b_2= 
\frac{1}{4}, b_3=\frac{1}{4}-\frac{1}{2\pi}$.

	\section{Completion of the proof of Theorem~\ref{thm: global}}\label{sec: E}

\subsection{Extension to the case $d>k$}\label{sec: dkextension}
Given $d > k$, append $d-k$ zeros to the end of each row of $\WW \in M(k,k)$ to define $\widetilde{\WW}\in M(k,d)$.  Similarly,
extend the target $\VV$ to $\widetilde{\VV}\in M(k,d)$.
Denote the associated objective function by $\widetilde{\cal{F}}$ and note that if $\WW \in M(k,k)$ is a critical point of $\mathcal{F}$, then
$\widetilde{\WW}\in M(k,d)$ is a critical point of $\widetilde{\cal{F}}$.

We make use of the following result,
adapted from Lemma 8 in \cite{safran2017spurious}
\begin{lemma}
(Notation and assumptions as above.) Assume $d > k$ and set $m = d-k$.
Let $\WW$ be a critical point of $\mathcal{F}$ which has no parallel rows. Then the Hessian $\widetilde{H}$ of
$\widetilde{\cal{F}}$ at $\widetilde{\WW}$ may, after a permutation of rows and columns, be written in block diagonal form $[\widetilde{H_{ii}}]_{i\in\ibr{m+1}}$
where $\widetilde{H_{11}}= H \in M(k^2,k^2)$ is the Hessian of $\mathcal{F}$, and for $i > 1$, the matrices $\widetilde{H_{ii}}$ are all equal to the $k\times k$-matrix $M=[m_{ij}]$ defined by
\[
m_{ij} = \begin{cases} 
&\frac{1}{2} + \frac{1}{2\pi} \sum_{\ell\in \ibr{k}}\left( \frac{\sin(\theta_{\ww_i,\ww_\ell}) \|\ww_\ell\|}{\|\ww_i\|}- \frac{\sin(\theta_{\ww_i,\vv_\ell})}{\|\ww_i\|}\right),\quad i = j\\
&\frac{1}{2\pi}(\pi - \theta_{\ww_i,\ww_j}), \quad i \ne j
\end{cases}
\]
\end{lemma}

\begin{thm}\label{thm: global2}
(Assumptions and notation of Theorem~\ref{thm: global}) Let $d > k$.
\begin{enumerate}
\item Suppose $\widetilde{\WW} = \widetilde{\VV}$. In addition to the eigenvalues described in Theorem~\ref{thm: global},
there will be an 2 additional eigenvalues: one equal to $\frac{1}{4}$, multiplicity $m(k-1)$, the other to $\frac{k+2}{4}$, multiplicity $m$.
\item Suppose $\WW$ is of type A. Then $\widetilde{\WW}$ will have an additional $2$ eigenvalues.
One equal to $\frac{1}{4} -\frac{1}{\pi\sqrt{k}} + O(k^{-1})$, multiplicity $m(k-1)$, the other
to $\frac{k+1}{4} -\frac{1}{\pi\sqrt{k}} + O(k^{-1})$, multiplicity $m$.
\item Suppose $\WW$ is of type II. Then $\widetilde{\WW}$ will have an additional $3$ eigenvalues.
One equal to $\frac{1}{4} + \frac{4}{\pi^2 k} + O(k^{-2})$ of multiplicity $m(k-2)$, and two eigenvalues of multiplicity
$m$, one equal to $\frac{k+1}{4} + O(k^{-\frac{1}{2}})$, the other to 
$\frac{1}{2} +  O(k^{-\frac{1}{2}})$.
\end{enumerate}
In particular, type A and type II spurious minima exist for all $d \ge k \ge 6$.
\end{thm}
\proof Suppose $\widetilde{\WW} = \widetilde{\VV}$. The matrix $M\in M(k,k)$ defines an $S_k$-map of $\real^k$. Computing $M$,
we find that $m_{ii} = \frac{1}{2}$ and $m_{ij} = \frac{1}{4}$, $i,j \in \ibr{k}$, $i \ne j$.
Write $(\real^k,S_k)$ uniquely as the orthogonal direct sum $(H_{k-1},S_k)\oplus (T,S_k)$ 
Since $M$ is an $S_k$-map,
$M: H_{k-1}\arr H_{k-1}$ and $M: T\arr T$. Taking $X = [1,-1,0,\cdots, 0] \in H_{k-1}$, $M(X) = (m_{11}-m_{12}) X$,
giving the eigenvalue $\frac{1}{4}$. Similarly, for the eigenvalue associated to $(T_k,S_k)$ is $\frac{k+2}{4}$. The argument for
Type A critical points is similar: both the diagonal and off-diagonal entries are easily computed given the
estimates on the critical points used in the proof of Theorem~\ref{thm: global}. 
Finally, for type II critical points, we use the $S_{k-1}$-representation
$(\real^{k},S_{k-1})$ which has isotypic decomposition $\mathfrak{s}_{k-1} + 2 \mathfrak{t}$. The eigenvalue associated to
the $\mathfrak{s}_{k-1}$ factor is found exactly as for type A critical points and only uses the $m_{11}$ and $m_{12}$ entries of $M$.
For the eigenvalues associated to the factor $2 \mathfrak{t}$,
we use the realizations spanned by the basis vector $\vv_k$ and the vector $\sum_{i\in\ibr{k-1}} \vv_i$. However,
$\sin(\theta_{\ww_k,\vv_k})$ appears in the expression for $m_{kk}$ and this leads to the presence of terms
in $k^{-\frac{1}{2}}$ since $\sin(\theta_{\ww_k,\vv_k}) = 
\frac{4}{\pi\sqrt{k}}$~\cite{ArjevaniField2020}.  \qed

\section{Empirical results}\label{sec: empirical}

\subsection{Perturbing the trained model} \label{sec: 
emp_pert}
Our analysis shows that local minima exhibit a small number of distinct 
eigenvalues, independent of the number inputs $d$ and hidden neurons $k$. However, during 
the 
training processes we expect to see a small number clusters of eigenvalues 
forming upon convergence. Below, we perturb the type II local minima of $k=20$ 
by adding an independent zero-mean Gaussian noise per entry for different 
choices of 
variance. 
\begin{figure}[H]
	\begin{minipage}{0.45\linewidth}
		\includegraphics[scale=0.4]{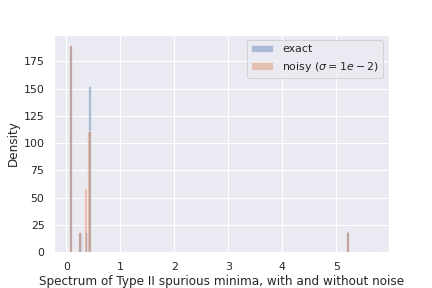}
	\end{minipage}
	\begin{minipage}{0.45\linewidth}
		\includegraphics[scale=0.4]{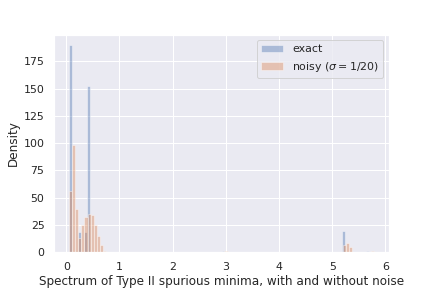}
	\end{minipage}

	\begin{minipage}{0.45\linewidth}
	\includegraphics[scale=0.4]{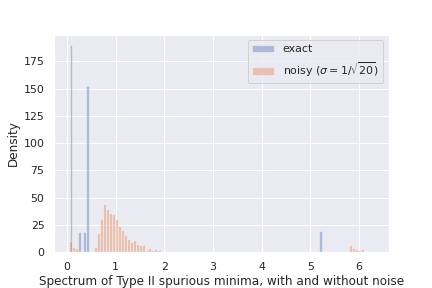}
	\end{minipage}
	\begin{minipage}{0.45\linewidth}
		\includegraphics[scale=0.4]{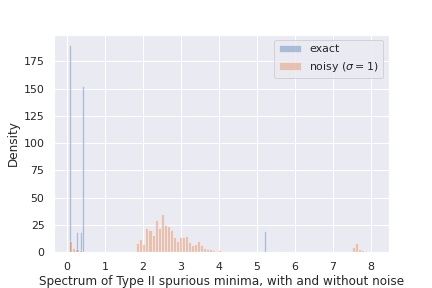}
	\end{minipage}
	\caption{The spectrum of the Hessian at type II spurious minima 		
	where the entries are perturb by adding independent zero-mean Gaussian 
	random entries with different variance values. As expected, the eigenvalues 
	accumulate in clusters around the eigenvalues of the type 
	II minima.}
	\label{fig:fig2}
\end{figure}

\subsection{Eigenvalue data for type A, I, II spurious 
minima 
}\label{sec: evs_emp}
In the sequel, we provide numerical estimates for the Hessian 
spectrum at types A, I and II minima. The Hessian
is computed using the expressions 
given in \pref{sec: C}, and evaluated using the 
estimates of the spurious minima. The spectrum is then approximated 
numerically using \text{LinAlg}, a linear algebra package of 
Python.

\begin{figure}[H]
\includegraphics[scale=0.4]{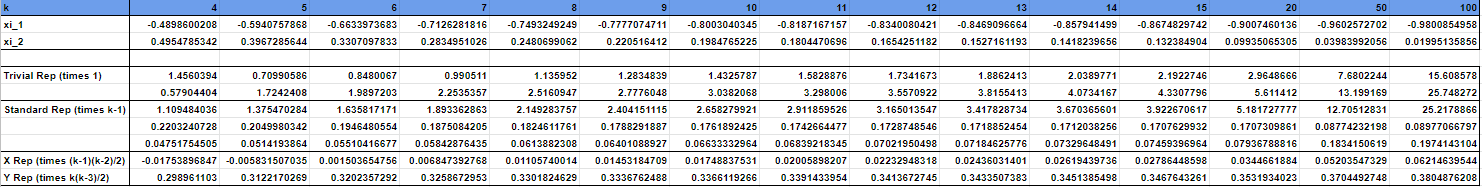}
\caption{The spectrum of type A spurious 
minima.}
\end{figure}

\begin{figure}[H]
	\includegraphics[scale=0.4]{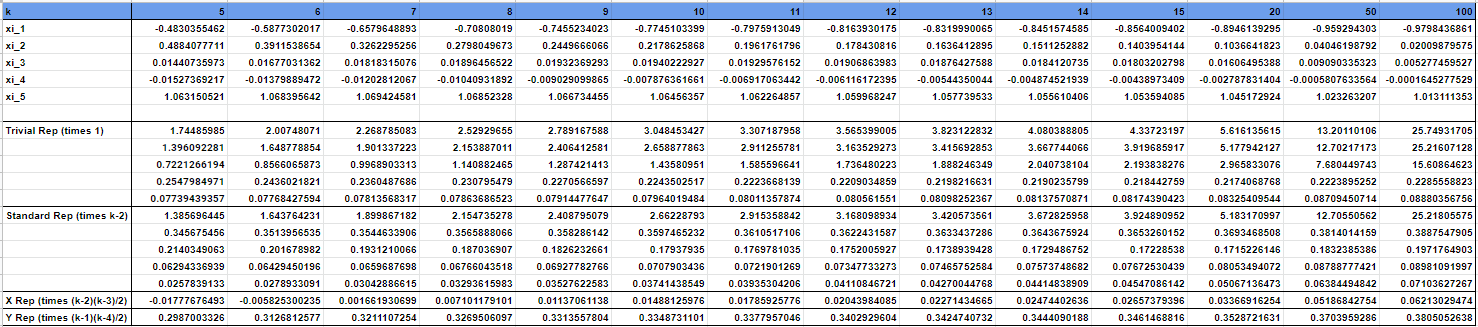}
	\caption{The spectrum type I spurious 
	minima.}
\end{figure}

\begin{figure}[H]
	\includegraphics[scale=0.4]{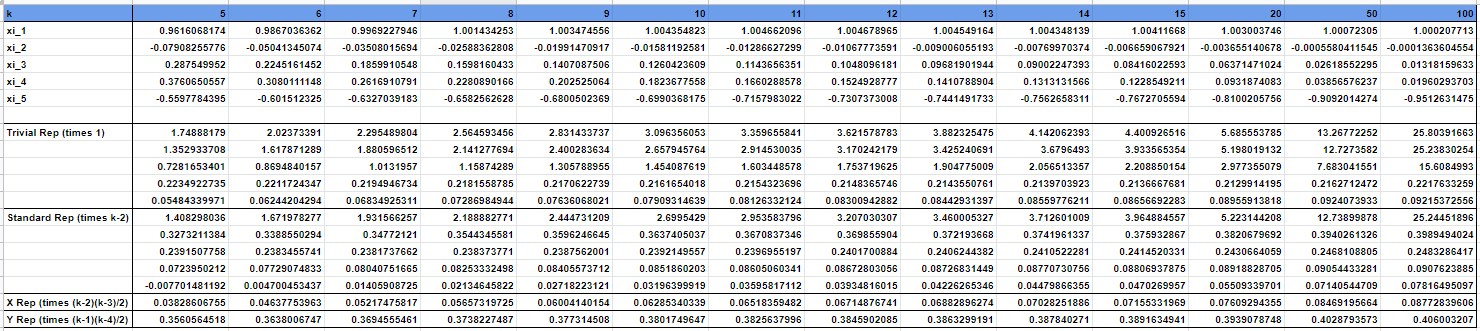}
	\caption{The spectrum type II spurious 
	minima.}
\end{figure}

%\begin{figure}[H]
%	\begin{minipage}{0.45\linewidth}
%		%\includegraphics[scale=0.49]{figs/density.eps}
%		\includegraphics[scale=0.4]{figs/noise1.png}
%	\end{minipage}
%	\begin{minipage}{0.45\linewidth}
%		%\includegraphics[scale=0.49]{figs/chart.eps}
%		\includegraphics[scale=0.4]{figs/noise2.png}
%	\end{minipage}
%	
%	\begin{minipage}{0.45\linewidth}
%		%\includegraphics[scale=0.49]{figs/density.eps}
%		\includegraphics[scale=0.4]{figs/noise3.png}
%	\end{minipage}
%	\begin{minipage}{0.45\linewidth}
%		%\includegraphics[scale=0.49]{figs/chart.eps}
%		\includegraphics[scale=0.4]{figs/noise4.png}
%	\end{minipage}
%	\caption{The spectrum of the Hessian at type II spurious minima 		
%		where the entries are perturb by adding independent zero-mean Gaussian 
%		random entries of different variance values. As expected the 
%eigenvalues 
%		accumulate in clusters around the eigenvalues of the limiting type 
%		II minima.}
%	\label{fig:fig2}
%\end{figure}

%
	
\end{document}